\documentclass{article}

\usepackage{microtype}
\usepackage{xcolor}
\usepackage{wrapfig} 
\usepackage{graphicx}
\usepackage{epstopdf}
\usepackage{caption}
\usepackage{subcaption}
\usepackage{booktabs} 
\usepackage{hyperref}
\usepackage[section]{placeins}
\usepackage{float}
\usepackage[acronym]{glossaries}
\newacronym{bmmv}{BMMV}{Block Multiple Measurement Vectors}
\newacronym{rip}{RIP}{restricted isometry property}
\newacronym{brip}{b-RIP}{block restricted isometry property}
\newacronym{mmv}{MMV}{multiple measurement vectors}
\newacronym{smv}{SMV}{single measurement vector}
\newacronym{omp}{OMP}{orthogonal matching pursuit}
\newacronym{momp}{M-OMP}{ \acrshort*{mmv} orthogonal matching pursuit}
\newacronym{msp}{M-SP}{\acrshort*{mmv} subspace pursuit}
\newacronym{bsbl}{M-BSBL}{block sparse Bayesian learning extended for \acrshort*{mmv}}
\newacronym{focuss}{FOCUSS}{ FOCal Underdetermined System Solver }
\newacronym{mfocuss}{M-FOCUSS}{ \acrshort*{mmv}-\acrlong*{focuss}}
\newacronym{ir}{IR}{implicit regularization}
\newacronym{irmmv}{IR-MMV}{implicit regularization in \acrlong{mmv}}
\newacronym{asrer}{ASRER}{average signal-to reconstruction-error-ratio}
\newacronym{smnr}{SMNR}{signal to measurement noise ratio}
\newacronym{lasso}{LASSO}{least absolute shrinkage and selection operator}
\newcommand{\wrt}{with respect to }
\newacronym{mamp}{AMP-MMV}{approximate message passing based \acrshort*{mmv}}
\newacronym{snr}{SNR}{signal to noise ratio}
\newacronym{rmse}{RMSE}{relative root mean squared error}

\usepackage[preprint]{icml2026}

\usepackage{amsmath, amsfonts}
\usepackage{amssymb}
\usepackage{mathtools}
\usepackage{amsthm}

\usepackage{array}

\usepackage{enumitem}
\usepackage{thmtools}
\usepackage{appendix}
\allowdisplaybreaks[1]
\usepackage[capitalize,noabbrev]{cleveref}

\theoremstyle{plain}
\newtheorem{theorem}{Theorem}[section]

\newtheorem{lemma}[theorem]{Lemma}
\newtheorem{corollary}[theorem]{Corollary}
\theoremstyle{definition}
\newtheorem{definition}[theorem]{Definition}

\theoremstyle{remark}
\newtheorem{remark}[theorem]{Remark}

\makeatletter
\newcommand{\BiggTw}{\bBigg@{2}}
\newcommand{\BiggTh}{\bBigg@{3}}
\newcommand{\BiggFo}{\bBigg@{4}}
\newcommand{\BiggFi}{\bBigg@{5}}
\newcommand{\ddt}{\frac{\mathrm{d}}{\mathrm{d}t}}
\makeatother

\usepackage[textsize=tiny]{todonotes}

\icmltitlerunning{Tuning-Free Structured Sparse Recovery of Multiple Measurement Vectors using Implicit Regularization}

\begin{document}

\twocolumn[
  \icmltitle{Tuning-Free Structured Sparse Recovery of Multiple Measurement Vectors using Implicit Regularization}



  \icmlsetsymbol{equal}{*}

  \begin{icmlauthorlist}
    \icmlauthor{Lakshmi Jayalal}{1}
    \icmlauthor{Sheetal Kalyani}{1}
  \end{icmlauthorlist}

  \icmlaffiliation{1}{Department of Electrical Engineering, Indian Institute of Technology, Madras, India}
  
  \icmlcorrespondingauthor{Lakshmi Jayalal}{ee19d751@smail.iitm.ac.in}

  \icmlkeywords{Machine Learning, ICML, compressed sensing, Implicit regularization, Sparse Signal Processing, Multiple Measurement Vectors, Gradient Flow}

  \vskip 0.3in
]



\printAffiliationsAndNotice{}  

\begin{abstract}
  Recovering jointly sparse signals in the \acrfull*{mmv} setting is a fundamental problem in machine learning, but traditional methods often require careful parameter tuning or prior knowledge of the sparsity of the signal and/or noise variance. We propose a tuning-free framework that leverages \acrfull*{ir} from overparameterization to overcome this limitation. Our approach reparameterizes the estimation matrix into factors that decouple the shared row-support from individual vector entries and applies gradient descent to a standard least-squares objective.
We prove that with a sufficiently small and balanced initialization, the optimization dynamics exhibit a ``momentum-like" effect where the true support grows significantly faster. %
Leveraging a Lyapunov-based analysis of the gradient flow, we further establish formal guarantees that the solution trajectory converges towards an idealized row-sparse solution. Empirical results demonstrate that our tuning-free approach achieves performance comparable to optimally tuned established methods. Furthermore, our framework significantly outperforms these baselines in scenarios where accurate priors are unavailable to the baselines.

\end{abstract}

\section{Introduction}
\label{sec:introduction}

Recovering structured signals from incomplete linear measurements is a central theme in modern signal processing, compressed sensing, and machine learning. A particularly important setting is the \acrfull*{mmv} problem corresponding to signals sharing a common underlying structure, particularly a shared sparse support. This joint sparsity structure, prevalent in applications  from neuroimaging \citep{chen2022mmv} to multiple-input multiple-output (MIMO) channel estimation \citep{yang2024mmv} allows for an improved recovery over processing single vectors individually. Classical approaches for \acrshort*{mmv} including \acrfull*{msp} \citep{liu2019mmv}, \acrfull*{momp} \citep{zhang2022some, chen2020mmv}, \acrfull*{mamp} \citep{ziniel2012efficient} and \acrfull*{mfocuss} {introduced by} \citet{cotter2005sparse}, often depend on the optimal tuning of regularization parameters or require prior knowledge of the sparsity level. While Bayesian methods like \acrfull*{bsbl} \citep{liu2013compression} can learn these parameters, they are often susceptible to noise.
These limitations motivate the search for robust, tuning-free methods for achieving row sparsity. 

An alternative paradigm based on \acrfull*{ir} through overparameterization has emerged with the works of \citep{arora2018optimization, arora2019implicit, li2018algorithmic}. Specifically, gradient descent is used on an overparameterized or factorized representation of the target signal rather than adding explicit penalty terms (like  $\ell_p^{}$ norms) to the loss function to encourage sparsity. The dynamics of the optimization process, often combined with specific initialization strategies or early stopping, implicitly guide the solution towards desirable structures like sparsity or low rank. This phenomenon has been analyzed in context of \acrfull*{smv} sparse recovery by \citet{ vaskevicius2019implicit, li2021implicit, chou2023more}. 

Unlike prior implicit regularization analyses in the \acrshort*{smv} setting, such as \citet{vaskevicius2019implicit, li2021implicit, chou2023more}, which factorize a single vector under an $\ell_2^{}$ loss, the \acrshort*{mmv} problem involves recovering a matrix with a shared row-sparse structure. This shared support couples the optimization dynamics across columns and naturally leads to a Frobenius-norm objective, motivating a parameterization tailored to \acrshort*{mmv} recovery. Building on these observations, we propose an overparameterized formulation for row-sparse \acrshort*{mmv} recovery and analyze the resulting implicit regularization induced by gradient descent.

We consider the standard least-squares loss function $\mathcal{L}(\mathbf{X}) = \Vert {\mathbf{Y} - \mathbf{A}\mathbf{X}}\Vert _F^{2}$ and optimize it using gradient descent. We propose a novel overparameterization that leverages the Hadamard product $(\odot)$ and Hadamard power $(^{\odot}_{})$, which represent element-wise multiplication and product, respectively. This reparameterization is $ \mathbf{X} = (\mathbf{g}^{\odot 2}_{} \mathbf{1}_L^{}) \odot \mathbf{V} $,
where  $\mathbf{1}_L = [1, 1, \dots, 1]_{1 \times L}^{}$ is a row vector of ones that replicates the squared elements of $\mathbf{g} \in \mathbb{R}^N_{}$ across the columns of $\mathbf{V} \in \mathbb{R}^{N \times L}_{}$. The core idea is that the gradient dynamics acting on the vector $\mathbf{g}$, which captures the shared row support, will implicitly drive most of its elements towards zero. {Due to the multiplicative structure of our parameterization, the corresponding rows of $\mathbf{V}$ are simultaneously driven to zero,} thereby inducing row sparsity in the reconstructed matrix, $\mathbf{X}$. This parametrization is designed specifically to handle multiple vectors sharing a common structure, distinguishing it from other \acrshort*{ir} methods that focus on \acrshort*{smv} by factorizing a single vector. Furthermore, it contrasts with matrix and tensor factorization methods that induce a low-rank bias rather than the desired row sparsity.

Our main contributions are: (1) we introduce a novel Hadamard parameterization for the \acrshort*{mmv} problem and derive its gradient descent update rules; (2) to the best of our knowledge, we are the first to utilize a Lyapunov-based gradient flow analysis in the context of implicit regularization. This approach departs from standard Grönwall-type arguments, yielding  initialization values which are practically feasible. We establish formal guarantees that the trajectory approaches an ideal row-sparse solution under a small and balanced initialization;  and (3) we further demonstrate the practical effectiveness of our approach through simulations, showing our approach achieves performance comparable to optimally tuned benchmarks (like \acrshort*{momp}, \acrshort*{msp}, \acrshort*{mamp}, \acrshort*{bsbl} and \acrshort*{mfocuss}) while overcoming their sensitivity to parameter misspecification.

\section{Related Works}
\label{sec:related_works}

This work builds upon research in several areas, primarily \acrshort*{mmv} sparse recovery algorithms, \acrshort*{ir} in machine learning {and the techniques used to analyze IR dynamics}.

\textbf{Sparse \acrshort*{mmv} Recovery Algorithms}: 
The challenge of recovering jointly sparse signals from \acrshort*{mmv} is 
addressed by diverse algorithms, many adapted from \acrshort*{smv} methods. Approaches range from greedy pursuit algorithms like \acrshort*{msp},
and \acrshort*{momp} to iteratively reweighted least square based \acrshort*{mfocuss} and Gaussian approximation approaches like \acrshort*{mamp}.
These traditional approaches often rely on explicit sparsity-promoting penalties (like $l_p$ norms) or greedy selection rules, sometimes requiring knowledge of the sparsity level or structure. Additionally, frameworks like \acrfull*{bsbl} \citep{liu2013compression} exploit statistical correlations to learn hyperparameters. But, when relying on estimated priors, frameworks like \acrshort*{bsbl} struggle to maintain accurate recovery in low \acrfull*{snr} regimes. Our work offers an alternative by achieving row sparsity implicitly through optimization dynamics.

\textbf{Implicit Regularization via Overparameterization: }
A significant body of recent work focuses on the phenomenon of \acrfull*{ir}, where {the dynamics of }
gradient descent { on an }
overparameterized model {find structured solutions } 
without explicit regularizers \citep{arora2018optimization, arora2019implicit}. 
{This  effect}
has been shown to implicitly favor low-rank solutions {in matrix sensing} \citep{soltanolkotabi2023implicit}
{and }
tensor sensing \citep{razin2021implicit, hariz2024implicit}. Crucially, implicit regularization has also been investigated for sparse recovery \citep{li2021implicit, zhao2022high, vaskevicius2019implicit, wu2020implicit} and structured sparse recovery \citep{li2023implicit} in the \acrshort*{smv} setting. 
The scale of initialization is a pivotal factor in these processes, often determining the nature of the learned solution. However, the factorizations used in prior work 
 are suboptimal for capturing the joint sparsity structure inherent in \acrshort*{mmv} applications. Our work addresses this gap by introducing a parameterization specifically designed for this context. In this framework, the dynamics of gradient descent on our proposed factors  implicitly promote the row-sparse structure critical for joint recovery.



\textbf{Analysis Techniques for Implicit Regularization:}
Analyzing the dynamics and convergence of gradient descent in overparameterized models requires specialized techniques. While methods based on Singular Value Decomposition (SVD) (e.g.,  \citet{arora2019implicit, soltanolkotabi2023implicit, li2018algorithmic}) { and} 
direct element-wise analysis \citep{li2021implicit, zhao2022high, vaskevicius2019implicit, li2023implicit} {are} effective for some sparse recovery problems, they become  intractable for the complex, non-linear parametrization such as the one we propose. We, therefore, adopt a dynamical systems perspective using gradient flow inspired by the works \citet{razin2021implicit} and \citet{hariz2024implicit}. 
However, unlike these prior studies which rely on Grönwall's inequality to bound solution trajectories, we employ a Lyapunov-based analysis. This enables us to avoid an exponential dependence on time for initialization and yields practically useful initialization values.

\section{Preliminaries}

We begin by establishing the basic notation followed by the formal definitions necessary for our analysis.

\textbf{Notations:}
Boldface capital letters are reserved for matrices. Boldface lowercase letters denote vectors, and lowercase letters denote scalars. $\mathbf{X}^{\top}$ is the transpose of matrix $\mathbf{X}$. $X^{}_{ij}$ is the $(i,j)$-th element of $\mathbf{X}_{}$.  $\mathbf{X}_{i:}^{}$ denotes the $i$-th row of any matrix $\mathbf{X}$.  $\Vert \cdot\Vert _F^{}, \Vert \cdot_{}\Vert _2^{}$ represent the Frobenius norm of a matrix, and the Euclidean norm of a vector. $\text{diag}(\mathbf{g})$ denotes a diagonal matrix with vector $\mathbf{g}$ as its diagonal elements. We use $\mathcal{\phi}(\mathbf{X}) = \Vert \mathbf{Y} - \mathbf{A}\mathbf{X}\Vert _F^2 $ to denote the squared Frobenius norm of the residual $\mathbf{Y} - \mathbf{A}\mathbf{X}$. The term $\mathbf{1}_{N\times L}^{}$ is a matrix of all ones of size $N\times L$.

With the notation established, we now define $\mu$-coherence to characterize the sensing matrix. We also introduce the unbalancedness and row-unbalancedness constants, which are crucial for analyzing the dynamics of our parameterization.

\begin{definition}[Mutual coherence]

Let $\mathbf{A}\in\mathbb R^{M\times N}_{}$ be a sensing matrix with $\ell_2^{}$-normalized columns, i.e., $\Vert\mathbf A_{:i}\Vert_2^{}=1$ for all $i\in[N]$. The \textbf{mutual coherence} of $\mathbf A$ is defined as
    $$\mu(\mathbf{A}) = \max_{i\neq j} {\vert \langle\mathbf{A}_{:i}^{},\mathbf{A}_{:j}^{}\rangle\vert }$$
    where $\mathbf{A}_{:i}^{}$ is the $i$-th column of sensing matrix $\mathbf{A}$ and $\langle\cdot,\cdot\rangle$ denotes the inner product.
    \label{def:IRMMV_mucoherence}
This notion of mutual coherence is standard in compressed sensing and has also been adopted in recent analyses of implicit regularization in overparameterized models, e.g., \citet{li2021implicit}. Throughout this work, we assume that the sensing matrix $\mathbf{A}$ is $\mu$-coherent with $\ell_2$-normalized columns.
\end{definition}

{\begin{definition}\label{def:globalBalancedness}
    The constant $\epsilon(t)\geq 0 $ be defined as the unbalancedness constant ({introduced by} \citet{razin2021implicit}), where $\epsilon(t) =  \left\vert \frac{1}{2}\Vert \mathbf{g}(t)\Vert ^2_2- \Vert \mathbf{V}(t)\Vert _F^2\right\vert  $ for any time $t\geq 0$. 
\end{definition}
\begin{definition}\label{def:row:balancedness}
    The constant $\epsilon_r^{}(t)\geq 0 $ be defined as the row-unbalancedness constant, where $\epsilon_r^{}(t) =  \underset{i \in [N]}{\max}\left\vert  \frac{1}{2}{g}_i^2(t) - \sum\limits_j{V}_{ij}^2(t)\right\vert  $ for any time $t\geq 0$ and $\forall i\in[N]$.
\end{definition}}
While defined at any time $t\geq 0$, we show in Lemma \ref{lemma:MMV_Balancedness} that both $\epsilon(t)$ and $\epsilon_r^{}(t)$ are conserved quantities under the gradient flow dynamics. Consequently, we treat them as time-invariant constants throughout this work, hence dropping the time index.
With these preliminaries established, we now introduce our \acrshort*{ir} framework for the \acrshort*{mmv} problem.
\section{IR-MMV}
\label{sec:IR-MMV}

Consider the task of recovering an \acrshort*{mmv} $\mathbf{X} \in \mathbb{R}^{N \times L}$, characterized by a common sparsity pattern across its rows. The measurement model is given by:
\begin{align}
    \mathbf{Y} = \mathbf{A}\mathbf{X}+\mathbf{W}, \label{eqn:measurement_model}
\end{align}
where $\mathbf{Y} \in \mathbb{R}^{M \times L}$ is the measurement matrix, $\mathbf{A} \in \mathbb{R}^{M \times N}$ is the sensing matrix (with $M \leq N$), and $\mathbf{W} \in \mathbb{R}^{M \times L}$ represents additive measurement noise. The objective is to recover $\mathbf{X}$, which is characterized by having at most $K$ non-zero rows.

\begin{algorithm}[tb]
\caption{{Implicit Regularization for MMV (\acrshort*{irmmv})}}
\label{alg:ir-mmv}
\begin{algorithmic}
\STATE \textbf{Input:} Sensing matrix $\mathbf{A} \in \mathbb{R}^{M \times N}$, measurement matrix $\mathbf{Y}\in \mathbb{R}^{M \times L}$
%
\STATE \textbf{Initialize:} $\alpha_V^{} = 5\times 10^{-4}_{},  \alpha_g^{} = \alpha_V^{}{\sqrt{2L}} $, learning rates $\eta_g = \eta_v = 10^{-4}_{}$, number of iterations $T = 5\times 10^6_{}$,  $\mathbf{g}(0) \in \mathbb{R}^{N}$ with $\mathbf{g}(0) = \alpha_g\mathbf{1}_L^{}$, $\mathbf{V}(0) \in \mathbb{R}^{N \times L}$ with $\mathbf{V}(0) = \alpha_v\mathbf{1}_{N\times L}$.

\STATE \textbf{Reconstruct initial signal:} $\mathbf{X}(0) = (\mathbf{g}(0)^{\odot 2} \mathbf{1}_L) \odot \mathbf{V}(0)$.

\FOR{$t = 0, 1, \dots, T-1$}
    \STATE $\boldsymbol{\Lambda} \gets \mathbf{A}^\top_{}(\mathbf{Y} - \mathbf{A}\mathbf{X})$
    \STATE  Update parameters: 
    \begin{align*}
        \mathbf{g}(t+1) &= \mathbf{g}(t) + 4\eta_g^{} \left(\mathbf{g}(t)\odot \left((\boldsymbol{\Lambda}\odot \mathbf{V}(t)) \mathbf{1}_L^\top\right) \right)\\
        \mathbf{V}(t+1) &= \mathbf{V}(t) + 2\eta_v \left( (\mathbf{g}_{}^{\odot 2}(t+1)\mathbf{1}_L^{}) \odot \boldsymbol{\Lambda} \right)
    \end{align*}
    \STATE Reconstruct signal: $\mathbf{X}(t+1) = (\mathbf{g}^{\odot 2}(t+1) \mathbf{1}_L) \odot \mathbf{V}(t+1)$
\ENDFOR

\STATE \textbf{Output:} Estimated signal matrix $\mathbf{X}$.
\end{algorithmic}
\end{algorithm}

Instead of directly optimizing for $\mathbf{X}$, we introduce an overparameterized representation of $\mathbf{X}$ based on a Hadamard product factorization. This reparameterization is designed such that the inherent dynamics of gradient descent implicitly promote row sparsity. Specifically, we propose to factorize $\mathbf{X}$ as:
\begin{align}
    \mathbf{X} = (\mathbf{g}^{\odot 2} \mathbf{1}_L^{}) \odot \mathbf{V}, \label{eqn:hadamard_param}
\end{align}
where $\mathbf{g} = [g_1, g_2, \dots, g_N]^\top \in \mathbb{R}^N$ acts as a row-wise scaling vector, and $\mathbf{V} \in \mathbb{R}^{N \times L}$ is a component matrix. The term $\mathbf{1}_L = [1, 1, \dots, 1]_{1 \times L}$ ensures that the squared elements of $\mathbf{g}$ are replicated across columns. The square term on $\mathbf{g}$ (i.e., $\mathbf{g}^{\odot 2}$) is introduced to keep the row activations positive. Recall that the proposed reparameterization when applied to gradient descent optimization will implicitly drive elements of $\mathbf{g}$ corresponding to inactive rows towards zero. This phenomenon will encourage row sparsity in the reconstructed $\mathbf{X}$, leveraging the dynamics of the optimization process rather than explicit regularization terms. 
We define the mean squared loss as: \begin{align}
    \mathcal{L}^{}_{}(\mathbf{g}, \mathbf{V}) =  \Vert\mathbf{Y} - \mathbf{A}\left((\mathbf{g}^{\odot 2} \mathbf{1}_L^{}) \odot \mathbf{V}\right)\Vert_F^2.
\end{align}

The optimization is performed using gradient descent on the factors $\mathbf{g}$ and $\mathbf{V}$. At each iteration, the updated factors are used to reconstruct $\mathbf{X}$. The gradient descent update rules for $\mathbf{g}$ and $\mathbf{V}$ are derived from the loss function $\mathcal{L}^{}_{}(\mathbf{g}, \mathbf{V})$. 
The gradient descent updates for $\mathbf{g}$ and $\mathbf{V}$ are:
\begin{align}
    \mathbf{g}(t+1) &= \mathbf{g}(t) - \eta_g \nabla_{\mathbf{g}} \mathcal{L}(\mathbf{g}(t), \mathbf{V}(t)) \label{eqn:IRMMV_gUpdate} \\ 
    \mathbf{V}(t+1) &= \mathbf{V}(t) - \eta_v \nabla_{\mathbf{V}} \mathcal{L}(\mathbf{g}(t), \mathbf{V}(t)) \label{eqn:IRMMV_vUpdate}
\end{align}
The reconstructed signal matrix $\mathbf{X}$ at time $t$ is then given by:
\begin{align}
    \mathbf{X}(t+1) = (\mathbf{g}^{\odot2}_{}(t+1)\mathbf{1}_L^{})\odot\mathbf{V}(t+1). \label{eqn:IRMMV_xUpdate}
\end{align}
The approach is outlined in Algorithm \ref{alg:ir-mmv}.
In line with the analysis of matrix and tensor factorization by \citet{gunasekar2017implicit, arora2018optimization, li2018algorithmic, arora2019implicit, razin2021implicit, hariz2024implicit}, we also model small learning rate for gradient descent through infinitesimal limit, i.e. through gradient flow:
\begin{align}
    \frac{\mathrm{d}}{\mathrm{d}t}{g}_l^{}(t) &= - \nabla{g_{l}^{}} \mathcal{L}(\mathbf{g}(t), \mathbf{V}(t)),\\
  \frac{\mathrm{d}}{\mathrm{d}t}V_{lm}^{}(t) &=  -\nabla{V_{lm}^{}} \mathcal{L}(\mathbf{g}(t), \mathbf{V}(t)),
\end{align}
where $g_{l}(t)$ represents the element of factor $\mathbf{g}$ corresponding to the $l$-th element at time $t$ and $V_{lm}^{}(t)$ corresponds to the $(l,m)$-th element of factor $\mathbf{V}$ at time $t$. In the following section, we will formally analyze the dynamics of gradient flow to show that they guide the solution towards a row-sparse structure.
\section{Theoretical Analysis}
\label{sec:IRMMV_TheoreticalSection}

We begin by identifying a key conservation law of the optimization dynamics for the proposed approach. The following lemma shows that the proposed overparameterization preserves a balancedness property, both globally and at the row level, throughout the optimization process.

\begin{lemma}[Balancedness]
\label{lemma:MMV_Balancedness}
For the system evolving under continuous gradient flow for the set of update equations \eqref{eqn:IRMMV_gUpdate}, \eqref{eqn:IRMMV_vUpdate} and \eqref{eqn:IRMMV_xUpdate}, derived from the loss function $\mathcal{L}(\mathbf{g}, \mathbf{V})=\Vert \mathbf{Y}-\mathbf{A}(\mathbf{g}^{\odot2}_{}\mathbf{1}_L\mathbf{V})\Vert _{F}^{2}$, the following properties hold for any time $t \ge 0$
\begin{enumerate}
    \item Global balancedness: The quantity $\frac{1}{2}\Vert \mathbf{g}(t)\Vert _{2}^{2}-\Vert \mathbf{V}(t)\Vert _{F}^{2}$ is conserved throughout the optimization process:
    $$ \frac{1}{2}\Vert \mathbf{g}(t)\Vert _{2}^{2}-\Vert \mathbf{V}(t)\Vert _{F}^{2}=\frac{1}{2}\Vert \mathbf{g}(0)\Vert _{2}^{2}-\Vert \mathbf{V}(0)\Vert _{F}^{2} $$
    \item  Row-wise balancedness: For each individual row $i \in [N]$, the quantity $\frac{1}{2}g_{i}^{2}(t)-\sum\limits_{j\in[L]}V_{ij}^{2}(t)$ is also conserved:
    $$ \frac{1}{2}g_{i}^{2}(t)-\sum_{j\in[L]}V_{ij}^{2}(t)=\frac{1}{2}g_{i}^{2}(0)-\sum_{j\in[L]}V_{ij}^{2}(0) $$
\end{enumerate}
\end{lemma}
\begin{proof}[Proof sketch (for detailed proof see \ref{lemma:MMV_BalancednessApp})]
       {The result follows by showing that under gradient flow, $
\frac{1}{2}\sum\limits_{l=1}^{N}\frac{\mathrm{d}}{\mathrm{d}t} g_l^{2}(t)  = \sum\limits_{l=1}^{N}\frac{\mathrm{d}}{\mathrm{d}t} \Vert \mathbf{V}_{l:}(t)\Vert _2^2$ for all $t\geq 0$.} 
\end{proof}   
   
Lemma \ref{lemma:MMV_Balancedness} proves that if the parameters are balanced in the beginning at time $t = 0$ (i.e., $\frac{1}{2}\Vert \mathbf{g}(0)\Vert _{2}^{2}=\Vert \mathbf{V}(0)\Vert _{F}^{2}$), then they will remain perfectly balanced over time, meaning $\frac{1}{2}\Vert \mathbf{g}(t)\Vert _{2}^{2}=\Vert \mathbf{V}(t)\Vert _{F}^{2}$ for all subsequent time steps.
It also shows that for each row $i$, the relative proportions between the scaling power of $g_i$ and the ``size" of the $i$-th row of $\mathbf{V}$ (captured by its squared Frobenius norm) are preserved during learning. Here, the proposed \acrshort*{ir} mechanism for inducing sparsity acts consistently at a row-level. If the optimization drives $g_i^{}(t)$ towards zero for a particular row $i$, then, due to the row-wise balancedness, the corresponding row $\mathbf{V}_{i:}^{}$ must also approach zero, causing the entire $i$-th row of $\mathbf{X}$ to become sparse.  
\begin{remark}\label{remark:epsilonrEpsilonRelation}
In the general case, $\epsilon$ and $\epsilon_{r}$ are independent quantities. However, under the specific initialization of Algorithm 1 (where $g(0)$ and $V(0)$ are initialized with constant values $\alpha_{g}$ and $\alpha_{V}$), it strictly holds that $\epsilon_{r} \leq \epsilon$ for all $t \geq 0$.  (See Appendix \ref{proofOfRemark} for the proof.)
\end{remark}
We now characterize how the norms of individual rows of the reconstructed signal evolve under gradient flow. The next lemma provides upper and lower bounds on the rate of change of each row norm.
\begin{theorem}[Dynamics of row norm]
\label{thm:MMV_gradientFlowCompleteInequalityRowNormMainDoc}
Consider a system evolving under continuous gradient flow with updates equations \eqref{eqn:IRMMV_gUpdate}, \eqref{eqn:IRMMV_vUpdate} and \eqref{eqn:IRMMV_xUpdate} with initialization $g(0)=\alpha_{g}^{}\mathbf{1}_{L}$ and $V(0)=\alpha_{V}^{}\mathbf{1}_{N\times L}$ where $\alpha_g^{}$ and $\alpha_V^{}$ are small positive scalars. Let $\mathbf{\Lambda}(t):=\mathbf{A}^{\top}(\mathbf{Y}-\mathbf{A}(\mathbf{g}(t)^{\odot2}\mathbf{1}_{L}\odot \mathbf{V}(t)))$ and $\boldsymbol{\lambda}_{i}(t)$ denote the $i$-th row of $\mathbf{\Lambda}(t)$, and 
\begin{align}\hat{\mathbf{x}}_{i}(t) := \begin{cases} \frac{\mathbf{X}_{i:}^{}(t)}{\Vert \mathbf{X}_{i:}^{}(t)\Vert _{2}} & \text{if } \Vert \mathbf{X}_{i:}^{}(t)\Vert _{2} \neq 0 \\ \mathbf{0} & \text{if } \Vert \mathbf{X}_{i:}^{}(t)\Vert _{2} = 0 \end{cases}.
\end{align} Then, for any time $t \ge 0$ and for all $i \in [N]$ where $g_i^2(t) > 0$ and $\sum\limits_{m\in[N]}V_{im}^2 (t)> 0$,\\
if $\left<\boldsymbol{\lambda}_{i}^{}(t),\hat{\mathbf{x}}_i^{} (t)\right> \geq 0$ then, 
\begin{align}
    \frac{\mathrm{d}}{\mathrm{d}t} \Vert \mathbf{X}_{i:}^{}(t)\Vert _2 
      & \leq  24\left<\boldsymbol{\lambda}_{i}^{}(t),\hat{\mathbf{x}}_i^{}(t) \right>\left(\epsilon + \Vert \mathbf{X}_{i:}^{}(t)\Vert _2^{2/3}\right)^2_{}\\
     \frac{\mathrm{d}}{\mathrm{d}t} \Vert \mathbf{X}_{i:}^{}(t)\Vert _2  &\geq  {6}\left<\boldsymbol{\lambda}_i^{}(t), \hat{\mathbf{x}}^{}_{i}(t)\right> \left(\frac{{\Vert \mathbf{X}_{i:}^{}(t)\Vert _2^2}}{\epsilon_{}^{} + \Vert \mathbf{X}_{i:}^{}(t)\Vert _2^{\frac{2}{3}}} \right),
\end{align}
and if $\left<\boldsymbol{\lambda}_{i}^{}(t),\hat{\mathbf{x}}_i^{} (t)\right> < 0
$ then,
\begin{align}
     \frac{\mathrm{d}}{\mathrm{d}t} \Vert \mathbf{X}_{i:}^{}(t)\Vert _2 
      & \geq  24\left<\boldsymbol{\lambda}_{i}^{}(t),\hat{\mathbf{x}}_i^{} (t)\right>\left(\epsilon + \Vert \mathbf{X}_{i:}^{}(t)\Vert _2^{2/3}\right)^2_{}\\
     \frac{\mathrm{d}}{\mathrm{d}t} \Vert \mathbf{X}_{i:}^{}(t)\Vert _2  &\leq  {6}\left<\boldsymbol{\lambda}_i^{}(t), \hat{\mathbf{x}}^{}_{i}(t)\right> \left(\frac{{\Vert \mathbf{X}_{i:}^{}(t)\Vert _2^2}}{\epsilon_{}^{} + \Vert \mathbf{X}_{i:}^{}(t)\Vert _2^{\frac{2}{3}}} \right).
\end{align}
\end{theorem}
\begin{proof}[Proof sketch (for detailed proof see Theorem \ref{thm:MMV_gradientFlowCompleteInequalityRowNorm})]
    The proof begins by expressing the time derivative of the row norm, $\frac{\mathrm{d}}{\mathrm{d}t} \Vert \mathbf{X}_{i:}^{}(t)\Vert _2$, in terms of the underlying parameters ${g}_i^{}(t)$ and $\sum\limits_j^{}{V}_{ij}^{}(t)$. The upper and lower bounds are then established by relating the parameter norms ($g_i^2$ and $\sum_j V_{ij}^2$) back to the matrix row norm ($\Vert \mathbf{X}_{i:}^{}\Vert _2$) using the row-unbalancedness constant, $\epsilon$. 
\end{proof}

To gain further insight into the row-wise dynamics, we next consider the special case of perfect row-balancedness. Under this condition, the evolution of each row norm admits a simplified and more explicit characterization.
\begin{corollary}[Dynamics of row norms under perfect balancedness]\label{corr:IRMMV_EvolutionOfComponents}
Assume that the system evolves under continuous gradient flow, as described in Lemma \ref{lemma:MMV_Balancedness}, with perfect row balancedness, i.e., $\frac{1}{2}g_{i}^{2}(t)=\sum\limits_{j\in[L]}V_{ij}^{2}(t)$ for all $i \in [N]$ and for any time $t \ge 0$. 
Then, the rate of change of the Euclidean norm of the $i$-th row of $\mathbf{X}(t)$ is characterized by:
   
    \begin{align}
        \frac{\mathrm{d}}{\mathrm{d}t} \Vert \mathbf{X}_{i:}^{}(t)\Vert _2 & =   {2^{2/3}_{}6}\left<\boldsymbol{\lambda}(t), \hat{\mathbf{x}}^{}_{i}(t)\right> \Vert \mathbf{X}_{i:}^{}(t)\Vert _2^{4/3}.
    \end{align}
\end{corollary}
\begin{proof}[Proof sketch (for detailed proof see Corollary \ref{corr:IRMMV_EvolutionOfComponentsApp})]
    {The proof begins by expressing the derivative $\frac{\mathrm{d}}{\mathrm{d}t} \Vert \mathbf{X}_{i:}^{}(t)\Vert _2 $ in terms of the underlying parameters $({\mathbf{g}}(t), {\mathbf{V}}(t))$. Then the result follows by applying row-balancedness assumption and substituting back the row norm.}
\end{proof}
Theorem \ref{thm:MMV_gradientFlowCompleteInequalityRowNormMainDoc} and Corollary \ref{corr:IRMMV_EvolutionOfComponents} show that the evolution rate of each row's norm is roughly proportional to its current norm raised to the power $4/3$.  This implies that when the unbalancedness constant is small,  this proportionality creates a ``momentum-like" effect: smaller row norms evolve more slowly, while larger row norms change more rapidly. This dynamic, in turn, fosters an incremental learning effect: a phenomenon where different parts of the signal are learned sequentially rather than all at once. This process is realized by certain rows growing significantly in magnitude while others remain small, ultimately leading to an implicit regularization that favors low-rank solutions.
The following lemma connects the behavior of the optimization in parameter space to the evolution of the reconstructed signal. It shows that the distance between two signal trajectories can be upper-bounded by the distance between their corresponding parameters.
\begin{lemma} [Distance bound in terms of parameters]
Let the estimated trajectory, $\mathbf{X}(t)$ be parametrized by $({\mathbf{g}}(t), {\mathbf{V}}(t))$ and a reference trajectory $\tilde{\mathbf{X}}(t)$, be parameterized by $(\tilde{\mathbf{g}}(t), \tilde{\mathbf{V}}(t))$ both of which evolve under the continuous gradient flow, as described in Lemma \ref{lemma:MMV_Balancedness}. 
Let $T_{}^{}>0$ and $D>0$ be scalars.  Specifically, assume that for all $t\in[0, T_{}]$ and all $n\in[N]$, the condition ${\max}\left\{\vert g_n^{}(t)\vert , \Vert V_{n:}^{}(t)\Vert _2^{}, \vert \tilde{g}_n^{}(t)\vert , \Vert \tilde{V}_{n:}^{}(t),\Vert _2^{}\right\}\leq D$ holds.
 Then at any time $t \in [0,T_{}^{}]$, the following inequality holds:
\begin{align} 
    \left\Vert \mathbf{X}(t) - \tilde{\mathbf{X}}(t)\right\Vert ^{2}_F &\leq  8D^4_{}\Big\Vert (\mathbf{g}(t), \mathbf{V}(t)) \left.- (\tilde{\mathbf{g}}(t), \tilde{\mathbf{V}}(t))\right\Vert ^2_{},
\end{align}
where, $\Big\Vert (\mathbf{g}(t), \mathbf{V}(t)) \left.- (\tilde{\mathbf{g}}(t), \tilde{\mathbf{V}}(t))\right\Vert ^2_{} = \sum\limits_{n\in[N]}\sum\limits_{l\in[L]}\left(\left({V}_{nl}^{}(t) - \tilde{{V}}_{nl}^{}(t)\right)^2 \right.\left.+\left({g}_{n}^{}(t) - \tilde{g}_n^{}(t)\right)^2_{}\right)$.\label{lemma:MMV_NormasIndividualComponents}
\end{lemma}
\begin{proof}[Proof sketch (for detailed proof see Lemma \ref{lemma:MMV_NormasIndividualComponentsApp})]
 The proof begins by substituting the parameterizations for  $\mathbf{X}(t)$ and $\tilde{\mathbf{X}}(t)$ into the squared Frobenius distance, $\Vert \mathbf{X}(t) - \tilde{\mathbf{X}}(t)\Vert _F^2$. The result then follows from an algebraic rearranging of terms and the application of standard norm inequalities, including the triangle inequality.
\end{proof} 
This provides the necessary link between the matrix trajectories and their underlying parameters, thus serving as the final component required for the convergence analysis. Specifically, the lemma bounds the distance between the reference trajectory, $\tilde{\mathbf{X}}(t)$, and the estimated trajectory, $\mathbf{X}(t)$ by the distance between their respective underlying parameters $({\mathbf{g}}(t), {\mathbf{V}}(t)), (\tilde{\mathbf{g}}(t), \tilde{\mathbf{V}}(t))$. 
Building on this analysis, the following theorem establishes our main convergence guarantee.
\begin{theorem}
\label{thm:IRMMV_mainResult}
Consider the system evolving under continuous gradient flow, where parameters are initialized such that perfect row-balancedness holds for all rows, i.e., $\frac{1}{2}g_{i}^{2}(0)=\sum\limits_{j\in[L]}V_{ij}^{2}(0)$ for all $i \in [N]$.
Let the sensing matrix $\mathbf{A}$ have $\ell_2$-normalized columns and be $\mu$-coherent.
Let $\mathbf{X}(t)\in\mathbb{R}^{N\times L}$ be the estimated trajectory obtained by the gradient flow with the proposed parameterization $\mathbf{X}(t)=(\mathbf{g}(t)^{\odot2}\mathbf{1}_{L})\odot \mathbf{V}(t)$. 
If the initial value $\alpha_V^{}, \alpha_g^{}$ (where $g(0)=\alpha_g \mathbf{1}_L$ and $V(0)=\alpha_V \mathbf{1}_{N \times L}$, and $\alpha_V = \frac{1}{\sqrt{2L}}\alpha_g$) is sufficiently small, specifically satisfying:

\begin{align}
    {\alpha_V^{}}&\leq \frac{\epsilon_{\text{app}}^{}}{{4D^2_{} }{}\sqrt{2NL(2L+1)}}\label{eqn:IRMMV_thmInitialization},
\end{align}
where  $\epsilon_{\text{app}}^{} \in (0,1)$, then for time $T_{}^{}>0$, there exists a rank-$K$ trajectory $\tilde{\mathbf{X}}(t)$ such that the estimate $\mathbf{X}(t)$ is close to $\tilde{\mathbf{X}}(t)$ until $t\geq T_{}^{}$,
i.e.,
\begin{align}
    \Vert \mathbf{X}(t)-\tilde{\mathbf{X}}(t)\Vert _{F} < \epsilon_{\text{app}}^{}
\end{align}
for all $t \in [0, T_{}^{}]$.
\end{theorem}
\begin{proof}[Proof sketch (for detailed proof see Appendix \ref{sec:IRMMV_proofofThmV1})]
We prove this by first constructing a reference rank-$K$ trajectory, $\tilde{\mathbf{X}}(t)$, and then showing that the estimated trajectory, $\mathbf{X}(t)$, remains close to the reference rank-$K$ trajectory for a sufficiently small initialization as defined in \eqref{eqn:IRMMV_thmInitialization}. The parameters of the estimated trajectory are initialized uniformly and balanced, i.e., $\mathbf{g}(0) = \alpha_g^{}\mathbf{1}_L^{}$ and $\mathbf{V}(0) = \alpha_V^{}\mathbf{1}_{N\times L}$. To construct the reference rank-$K$ trajectory, $\tilde{\mathbf{X}}(t)$, we begin by defining its own parameters $(\tilde{\mathbf{g}}(t), \tilde{\mathbf{V}}(t))$, such that $\tilde{\mathbf{X}}(t) = (\tilde{\mathbf{g}}(t)^{\odot2}\mathbf{1}_{L})\odot \tilde{\mathbf{V}}(t)$. The dynamics of these parameters are governed by the gradient flow on the loss function $\mathcal{L}(\tilde{\mathbf{g}}(t), \tilde{\mathbf{V}}(t))$. Crucially, for any support set $\mathcal{S}_K^{}$, the corresponding parameters are initialized as, for $i\in[N]$,
\begin{align}
    \tilde{g}_{i}^{}(0)& = \begin{cases}\sqrt{2}\rho^{1/3}_{}\ &\forall i \in \mathcal{S}_K^{}\\
0 &\forall i\notin \mathcal{S}_K^{}\end{cases},\label{eqn:IRMMV_ProofSketchtildaG}\\
\intertext{and}
\tilde{V}_{ij}^{}(0)& = \begin{cases}\frac{1}{\sqrt{L}}\rho^{1/3}_{}\ &\forall i\in\mathcal{S}_K^{}\\
0 &\forall i\notin \mathcal{S}_K^{}\end{cases},\ \forall \ j\in [L] ,\label{eqn:IRMMV_ProofSketchtildeV}
\end{align}
where, $\rho<L^{3/2}_{}\alpha_V^{3}$ is a small positive scalar. Because these non-support rows for reference trajectory start at zero,  
Lemma \ref{lemma:MMV_Balancedness}
and Corollary \ref{corr:IRMMV_EvolutionOfComponents}, ensure they remain zero for all time $t\geq0$. This construction guarantees the existence of a $\tilde{\mathbf{X}}(t)$, which is at most rank-$K$ matrix throughout its evolution. 
Note that this construction is purely for this theoretical proof; the proposed algorithm does not assume any knowledge of the support set or sparsity.
To bound the distance between  $\mathbf{X}(t)$ and $\tilde{\mathbf{X}}(t)$, we analyze the divergence of their corresponding parameters using a Lyapunov function, $\mathcal{E}(t) = \Big\Vert \big(\mathbf{g}(t), \mathbf{V}(t)\big) - \big(\tilde{\mathbf{g}}(t), \tilde{\mathbf{V}}(t)\big)\Big\Vert^2_{}$. This allows  practical initialization scale as is apparent in Equation \eqref{eqn:IRMMV_thmInitialization}. This stands in contrast to standard analysis techniques that rely on Grönwall's inequality \citep{razin2021implicit, gidel2019implicit}, which will lead to an initialization upper bound which will be exponentially smaller in $T$ (i.e., $\exp(-\beta T),$ where $\beta$ is the smoothness constant).

The Lyapunov function is decomposed into three components: the energy of the estimated parameters, the energy of reference parameters and the corresponding cross term. The time derivative of each term is then upper bounded with a time-integrable growth. Integrating these bounds yields a rational polynomial-in-time bound on the Lyapunov function, ensuring that the parameter distance remains below a prescribed threshold for all $t\in[0,T_{}^{}]$ provided the initialization satisfies \eqref{eqn:IRMMV_thmInitialization}.

Finally, the closeness of the parameter trajectories 
is translated to the distance between the trajectories $\mathbf{X}(t)$ and $\tilde{\mathbf{X}}(t)$ by applying the bound from Lemma \ref{lemma:MMV_NormasIndividualComponents}. This leads to the theorem's main guarantee: for a sufficiently small initialization $\alpha_V^{}, \alpha_g^{}$, the estimated solution $\mathbf{X}(t)$ is guaranteed to remain within a user-defined distance $\epsilon_{\text{app}}^{}$ of the reference row-sparse trajectory $\tilde{\mathbf{X}}(t)$ for a specified time duration.
\end{proof}
This theorem provides {the} theoretical guarantee for the proposed framework. It formalizes the implicit regularization effect by establishing that with a sufficiently small and balanced initialization, the algorithm's estimated trajectory, $\mathbf{X}(t)$ is guaranteed to stay within a predefined error $(\epsilon_{\text{app}}^{})$ of a reference rank-$K$ trajectory, $\tilde{\mathbf{X}}(t)$. By starting near the origin, the dynamics ensure that only a few rows that happen to have a slightly larger initial correlation with the true signal are selected to grow rapidly. Meanwhile balanced initialization ensures the proportional evolution of the factors required for the momentum-like dynamics to guide the algorithm's trajectory toward the row-sparse solution. 

{The following corollary extends the theoretical guarantees of Theorem \ref{thm:IRMMV_mainResult}. It establishes that if all reference trajectories converges uniformly to a global minimum of the objective function, then the estimated trajectory produced by the gradient flow of our proposed overparameterization will likewise converge to that same solution.}
\begin{corollary}\label{cor:IRMMV_MainResult}
Assume the conditions of Theorem \ref{thm:IRMMV_mainResult} and in addition, assume that all reference trajectories $\tilde{\mathbf{X}}(t)$ converge to a solution $\mathbf{X}^\star_{}\in\mathbb{R}^{N\times L}$ and this convergence is uniform in the sense that the trajectories are confined to a bounded domain, and for any $\epsilon_{\text{app}}^{} >0$ there exists a time $T_c^{}\leq T_{}^{}$ after which they are all within a distance $\epsilon_{\text{app}}^{}$ from $\mathbf{X}^\star_{}$. Then for any $\epsilon_{\text{app}}^{}>0$, if the initialization scales $\alpha_V^{}, \alpha_g^{}$ are sufficiently small, {for any time $t \in [T_c^{}, T_{}^{}]$ it holds that}
$\left\Vert \mathbf{X}(t) - \mathbf{X}^\star_{}\right\Vert ^{}_F\leq 2\epsilon_{\text{app}}^{}$.
\end{corollary}
\begin{proof}[Proof sketch (for detailed proof see Appendix \ref{sec:IRMMV_proofofCorollaryV1})]
    We assume that all reference trajectories $\tilde{\mathbf{X}}(t)$ converge to the oracle solution $\mathbf{X}^\star_{}$ within a finite time $T_{}>0$. This means that for any desired approximation error $\epsilon_{\text{app}}^{} >0$ we have $\Vert \tilde{\mathbf{X}}(t) - \mathbf{X}^\star_{}\Vert \leq \epsilon_{\text{app}}^{}$. 

    From Theorem \ref{thm:IRMMV_mainResult}, if the initialization values $\alpha_V^{}, \alpha_g^{}$ are sufficiently small, the estimated trajectory is guaranteed to be within a distance of $\epsilon_{\text{app}}^{}$ from the reference trajectory at least until time $T_{}^{}$. This is a conditional guarantee that holds as long as both trajectories remain within a bounded domain. We show that the estimated trajectory ${\mathbf{X}(t)}$ is also confined to a bounded domain. We know from our assumptions that the reference trajectory $\tilde{\mathbf{X}}(t)$ is bounded. 
    The result follows from applying the triangle inequality. 
\end{proof}

\section{Simulation}
\label{sec:IRMMV_SimulationSection}

In this section, we present simulations to reinforce our theoretical results. We also compare the performance of our proposed \acrshort*{irmmv} approach against five established MMV recovery algorithms: \acrshort*{momp}, \acrshort*{msp}, \acrshort*{mamp}, \acrshort*{bsbl} and \acrshort*{mfocuss}.
Unless otherwise specified, the default values for simulation parameters are: $M = 500$, $N = 10000$, $L = 20$, row sparsity level $K = 3$ $(K<N)$ typically used in \acrshort*{ir}/\acrshort*{mmv} literature. The sensing matrix $\mathbf{A}$ is generated by drawing its entries from an i.i.d. sub-Gaussian distribution (to satisfy $\mu$-coherence property) with $\ell_2$-normalized columns. In line with our theoretical analysis, the algorithm's hyperparameters are configured as follows: $\mathbf{g}$ is initialized with $\alpha_V^{} = 5\times 10^{-4}_{}$ to satisfy the small initialization requirement of Theorem \ref{thm:IRMMV_mainResult}, while the component matrix $\mathbf{V}$ is initialized to $\alpha_V^{} = \frac{1}{\sqrt{2L}}\alpha_g^{}$ to ensure the perfect row-wise balancedness. Finally, the maximum number of iterations is set to $T = 5\times 10^6_{}$ to provide the algorithm with sufficient time to converge from its deliberately small initial state. For our experiments, the parameters of \acrshort*{mfocuss} are set to $p=0.8$ and $\lambda=0.01$ as recommended by \citet{cotter2005sparse}. Additionally, the exact sparsity level $K$ was provided to \acrshort*{momp}, \acrshort*{msp} and \acrshort*{mamp}, with the latter also receiving the true noise variance.

\subsection{Balancedness and Incremental Learning}
\begin{figure}[t!]
    \centering \includegraphics[width=0.475\textwidth]{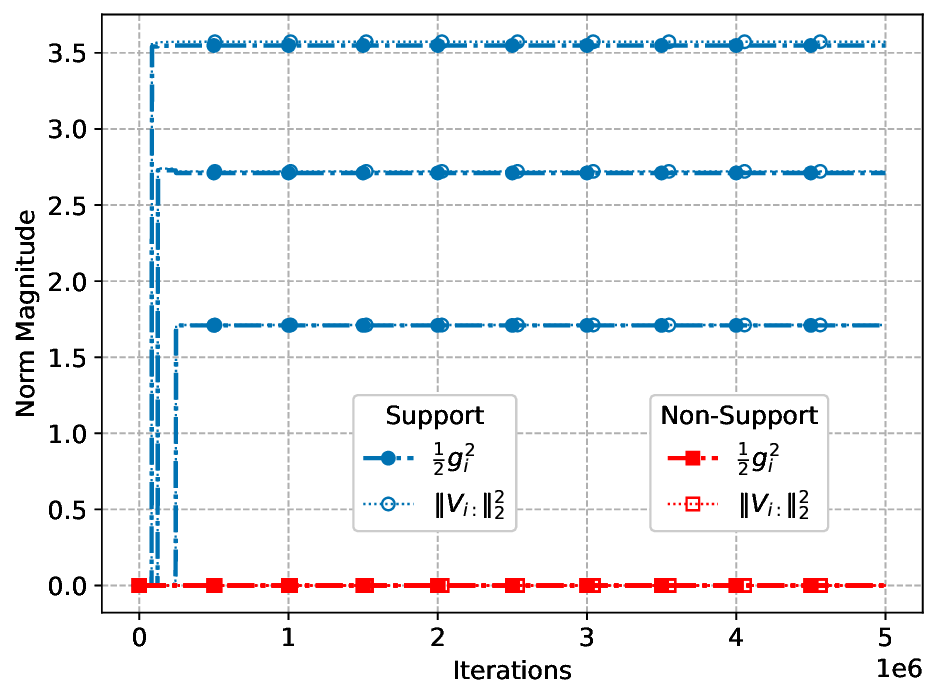}
    \caption{ Evolution of the norms of the components for rows in the support and non-support sets}
    \label{fig:coordinate-norms-iterates}
\end{figure}

For this experiment, the ground truth \acrshort*{mmv}, $\mathbf{X}$, is constructed such that its active rows are set to constant values of $\mathbf{1}_L^{}, 2\cdot\mathbf{1}_L^{} $ and $3\cdot\mathbf{1}_L^{}$. Recall that the parameters are initialized to ensure perfect row-wise balancedness. The learning rates for the gradient descent updates for $g$ (i.e.,$ \eta_g^{}$) and $V$ (i.e., $\eta_v^{}$) are both set to $ 10^{-4}_{}$, a small, constant value chosen for stable gradient descent. As illustrated in Figure \ref{fig:coordinate-norms-iterates}, the evolution of $\frac{1}{2}g_i^{2}(t)$ and $\Vert\mathbf{V}_{i:}^{}(t)\Vert_2^{2}$ is tracked throughout the optimization process. The empirical result support the theoretical findings of Lemma \ref{lemma:MMV_Balancedness}, by demonstrating that this balance is preserved over time. Note that the difference between the scalar component $g_i$ and the row norm $\Vert\mathbf{V}_{i:}\Vert_2^{}$ does not grow confirms that the selected learning rate is sufficiently small. This validates the equivalence between our gradient descent implementation and the gradient flow dynamics assumed in our analysis.

This dynamic leads to a momentum-like effect and an incremental learning phenomenon, where rows of the \acrshort*{mmv} are learned one at a time, a behavior consistent with prior findings by \citet{gissin2019implicit, razin2021implicit, hariz2024implicit} in the context of implicit regularization via overparametrization. The magnitude of the row norms directly influences the learning speed, with smaller norms evolving more slowly and larger norms changing more rapidly. This dynamic implicitly promotes structured sparsity. The time-invariant balancedness property (Lemma \ref{lemma:MMV_Balancedness}) guarantees that if an element $g_i(t)$ approaches zero, the corresponding row $\mathbf{V}_{i:}$ will also converge to zero, rendering the entire $i$-th row of $\mathbf{X}$ sparse.

\begin{figure*}[!ht] 
 \centering 
 \begin{subfigure}[b]{0.452\textwidth} \includegraphics[width=\textwidth]{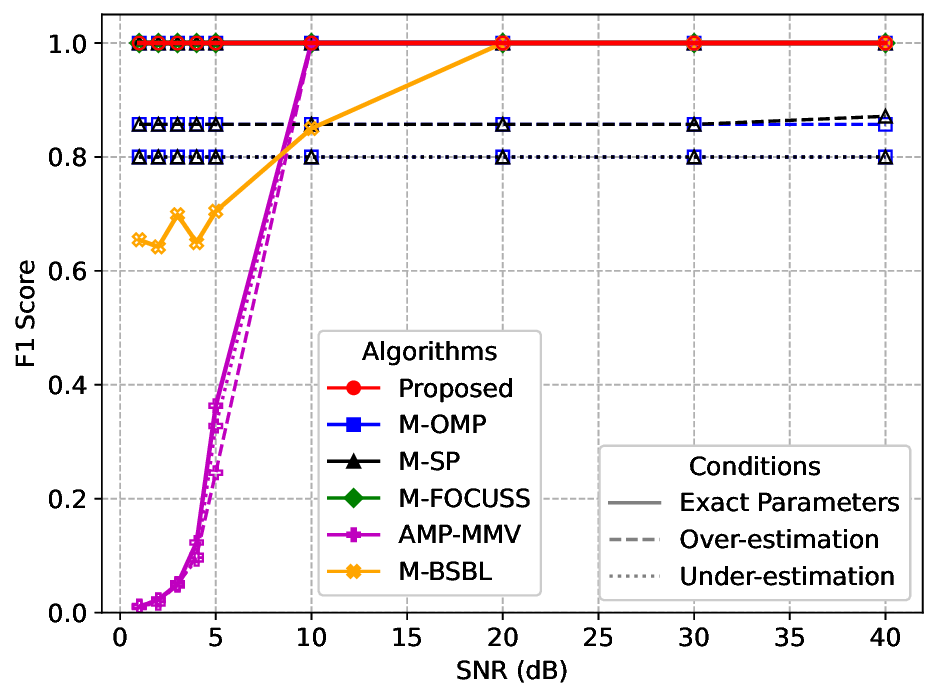}
 \caption{}
 \label{subfig:perfComparison_F1}
    \end{subfigure}
    \begin{subfigure}[b]{0.452\textwidth}
        \includegraphics[width=\textwidth]{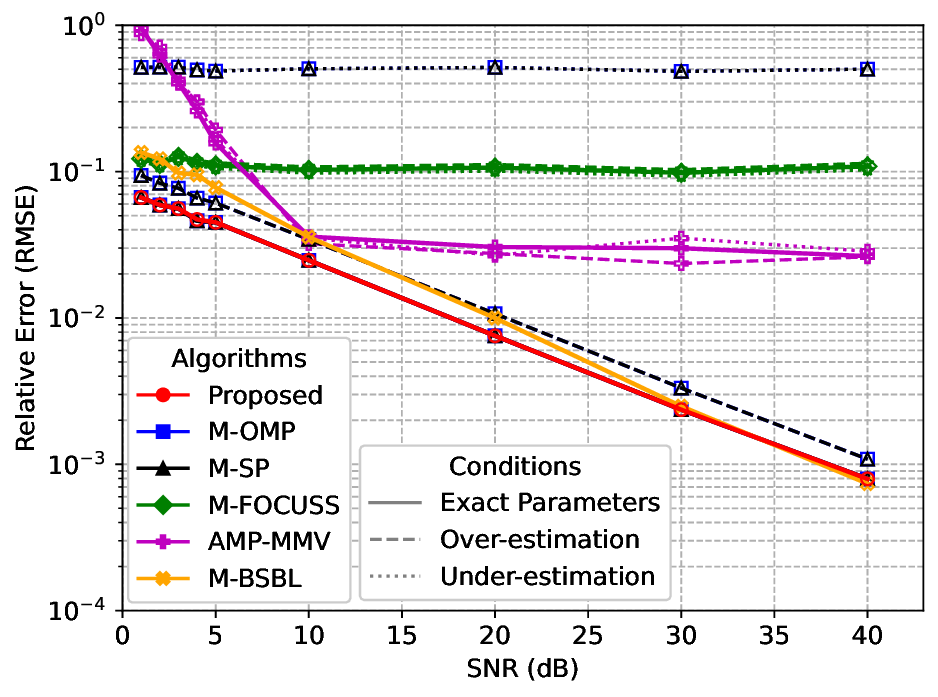}
        \caption{}
        \label{subfig:perfComparison_RMSE}
    \end{subfigure}
    \begin{subfigure}[b]{0.452\textwidth} \includegraphics[width=\textwidth]{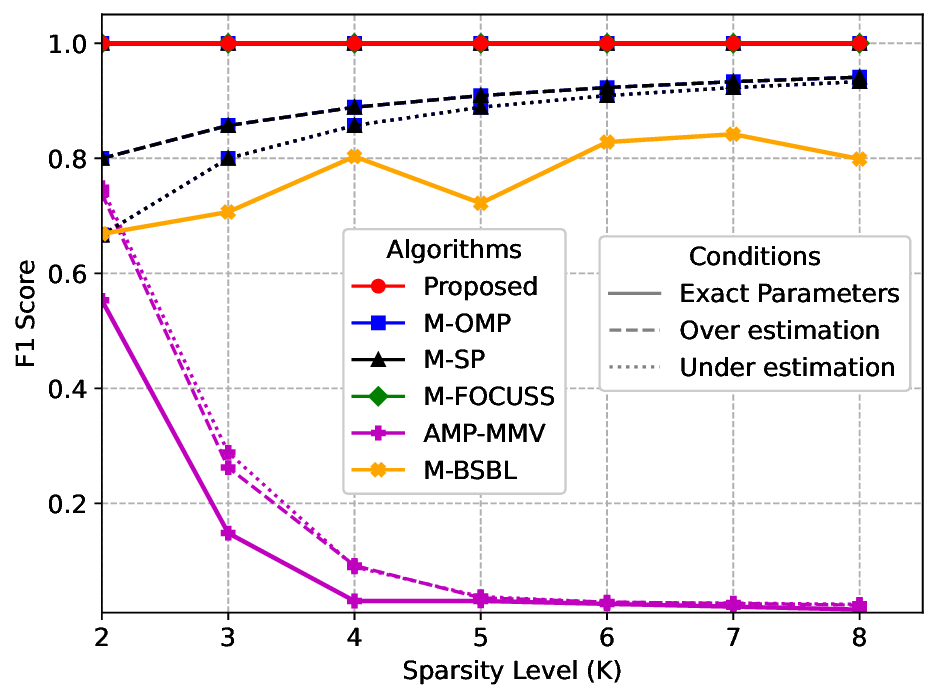}
 \caption{}
 \label{subfig:perfSparsityComparison_F1}
    \end{subfigure}
    \begin{subfigure}[b]{0.452\textwidth} \includegraphics[width=\textwidth]{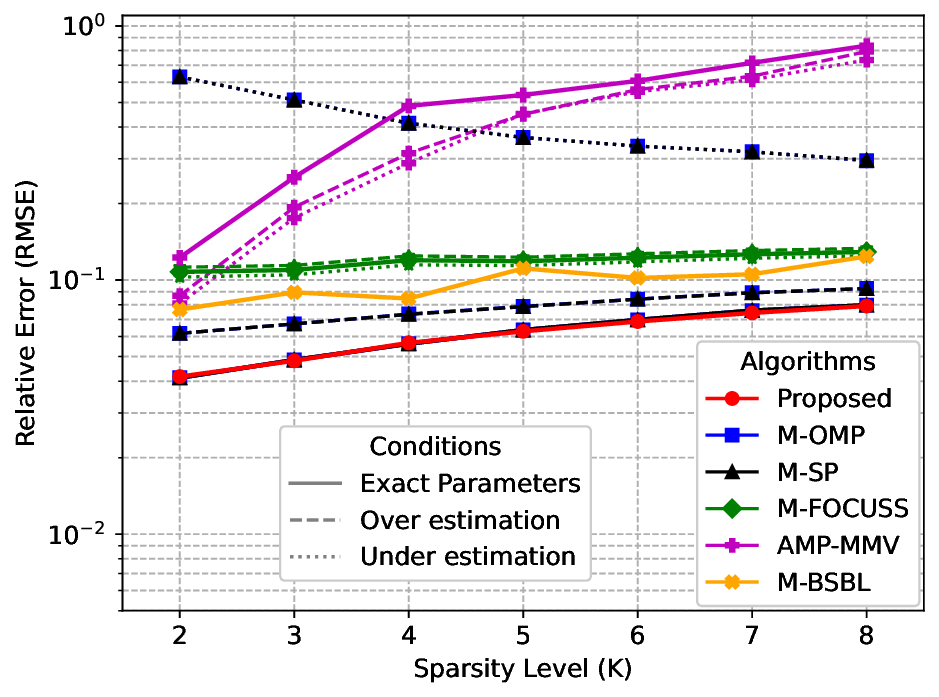}
 \caption{}
 \label{subfig:perfSparsityComparison_RMSE}
    \end{subfigure}
 \caption{Performance comparison of the proposed \acrshort*{irmmv}approach against baseline algorithms with both optimal and mismatched parameters. (a) Support Recovery (F1-Score) vs. SNR (b) Relative Error (RMSE) vs. SNR, (c) Support Recovery (F1-Score) vs. Sparsity $(K)$ and (d)  Relative Error (RMSE) vs.  Sparsity $(K)$.
 \label{fig:perfComparison}
 }
 \end{figure*}
\subsection{Performance Comparison with Other Algorithms}

The empirical performance of \acrshort*{irmmv} is compared against established \acrshort*{mmv} reconstruction algorithms, including \acrshort*{momp}, \acrshort*{msp},  \acrshort*{mamp}, \acrshort*{bsbl} and \acrshort*{mfocuss}. We evaluate recovery using two metrics: the reconstruction accuracy, measured by the \acrlong*{rmse}, $\text{RMSE} = \frac{\|\mathbf{X} - \hat{\mathbf{X}}\|_F}{\|\mathbf{X}\|_F}$, and the support recovery, quantified by the F$1$-Score F$1 = 2 \cdot \frac{\text{precision} \cdot \text{recall}}{\text{precision} + \text{recall}}$. The values are averaged over $10$ trials. In all experimental cases, the measurement matrix $\mathbf{Y}$ is generated by adding noise to the signal according to the model $\mathbf{Y} = \mathbf{A}\mathbf{X}+\mathbf{W}$. Consistent with \acrshort*{ir} literature, we construct the ground truth \acrshort*{mmv}, $\mathbf{X}$, such that its active rows are set to constant values of $\mathbf{1}_L^{}$. We vary the Signal-to-Noise Ratio (SNR) while fixing the row-sparsity at $K=3$. To further study the impact of signal complexity, we vary the sparsity level $K$ with the \acrshort*{snr} fixed at $4$dB.

As seen in Figure \ref{fig:perfComparison}, our approach consistently achieves performance comparable to the optimally tuned baselines. Our approach achieves this performance without any parameter tuning or prior knowledge of data-specific parameters. While \acrshort*{bsbl} does not require sparsity priors, they exhibit poor performance in the low \acrshort*{snr} regime $(0$dB - $10$dB$)$. This is likely because, in these high-noise settings, reliably learning the hyperparameters from the data becomes infeasible. We also observe that the \acrshort*{rmse} of \acrshort*{mfocuss} remains consistently higher than that of the other approaches, even though its F$1$ score is $1$. 

The performance of \acrshort*{momp}, \acrshort*{msp}, \acrshort*{mfocuss}, and \acrshort*{mamp} drops when we introduce parameter mismatches (e.g., $K\pm 1$, $\lambda\pm10\%$ and $\pm 10\%$ deviation in noise variance). As observed in Figures \ref{subfig:perfComparison_RMSE}, \ref{subfig:perfSparsityComparison_RMSE}, the greedy methods (\acrshort*{momp}, \acrshort*{msp}) perform very poorly when the sparsity level is underestimated ($K-1$), failing to converge even at high SNR. When the sparsity is overestimated ($K+1$), the performance deteriorates, though the degradation is less severe compared to the underestimation case. In contrast, overestimation yields better error rates but prevents these algorithms from achieving perfect support recovery, with F$1$-scores saturating below $1$ (Figures \ref{subfig:perfComparison_F1}, \ref{subfig:perfSparsityComparison_F1}). We defer the results on MNIST data to Appendix \ref{sec:MNISTResultApp}.

Summarizing, methods such as \acrshort*{momp}, \acrshort*{msp}, \acrshort*{mfocuss}, and \acrshort*{mamp} degrade in the presence of parameter misspecification while our method does not. Furthermore, we also outperform Bayesian approaches like \acrshort*{bsbl} at low \acrshort*{snr}. Hence, our approach has significant utility in practical scenarios where typically there is parameter misspecification and/or low \acrshort*{snr}.

\section{Conclusion}


We introduce a novel overparameterized factorization based on the Hadamard product to induce \acrshort*{ir} for structured sparse recovery of \acrshort*{mmv}. We show that the optimization dynamics naturally induce an implicit bias toward row-sparse solutions. Our characterization of the row norm dynamics identifies a momentum-like effect, where rows with larger norms grow significantly faster than those with smaller norms, leading to an incremental learning process that provably favors low-rank solutions. We prove that the relative scaling between the parameters is preserved both globally and for each row throughout the optimization process. Furthermore, by employing a Lyapunov-based analysis of the gradient flow, we derive a formal convergence guarantee showing that with a sufficiently small and balanced initialization, the algorithm's trajectory remains close to an idealized row-sparse solution. We demonstrate that the proposed method achieves performance comparable to algorithms like \acrshort*{momp}, \acrshort*{msp}, \acrshort*{mamp} and \acrshort*{mfocuss} when they are provided with exact parameters. Moreover, our framework significantly outperforms these baselines in scenarios where they suffer from parameter mismatch, and maintains robust recovery even in low \acrshort{snr} regimes where \acrshort{bsbl} performs poorly. Critically, the performance of the proposed framework is achieved without requiring prior knowledge of the sparsity level of the signal or its noise variance. 

\section*{Impact Statement}
This paper presents work whose goal is to advance the field of Machine
Learning. There are many potential societal consequences of our work, none
which we feel must be specifically highlighted here.

\bibliography{main}
\bibliographystyle{icml2026}

\newpage
\appendix
\onecolumn
\onecolumn
\section{Gradient Updates}
{This section details the derivation of the gradient descent update rules for the parameters $\mathbf{g}(t)$ and $\mathbf{V}(t)$.}
{We consider the loss function for the proposed parameterization, $\mathbf{X}(t) = \mathbf{g}^{\odot2}_{}(t)\mathbf{1}_L\odot\mathbf{V}(t)$, which is} reproduced here for readability:
\begin{align}
    \mathcal{L}(\mathbf{g}(t), \mathbf{V}(t)) \ = \Vert \mathbf{Y} - \mathbf{A}(\mathbf{g}^{\odot2}_{}(t)\mathbf{1}_L\odot\mathbf{V}(t))\Vert _F^2.
\end{align}
Let $\boldsymbol{\Lambda} := \mathbf{A}^{\top}\left(\mathbf{Y} - \mathbf{A}\mathbf{X}(t)\right)$.
{We begin by deriving the partial derivative of the loss function $\mathcal{L}(\mathbf{g}(t), \mathbf{V}(t))$ with respect to a single element $g_l^{}(t)$ (the $l$-th element of $\mathbf{g}(t)$):}
\begin{align}
    \nabla_{g_l^{}(t)}&\mathcal{L}(\mathbf{g}(t), \mathbf{V}(t)) = \nabla_{g_l^{}(t)}\BiggTh(\sum\limits_{i\in[M]}^{}\sum\limits_{j\in[L]}^{} \left(Y_{ij} - \sum\limits_{k\in[N]}^{}A_{ik}g_k^{2}(t)V_{kj}(t)\right)^2_{}\BiggTh)\\
    &= -4\sum\limits_{i\in[M]}^{}\sum\limits_{j\in[L]}^{} \left(Y_{ij} - \sum\limits_{k\in [N]}^{}A_{ik}g_k^{2}(t)V_{kj}(t)\right)A_{il}g_l^{}(t)V_{lj}(t)\\
     &=-4g_l^{}(t)\sum\limits_{i\in[M]}^{}\sum\limits_{j\in[L]}^{} A^{}_{il}\left(Y_{ij} - \sum\limits_{k\in[N]}^{}A_{ik}g_k^{2}(t)V_{kj}(t)\right)V_{lj} (t)\label{eqn:IRMMV_dLdgv1}\\
    &=-4g_l^{}(t)\sum\limits_{j\in[L]}^{} \mathbf{A}^\top_{}\left(\mathbf{Y}_{} - \mathbf{A}(\mathbf{g}^{\odot2}_{}(t)\mathbf{1}_L\mathbf{V}(t))\right)_{lj}V_{lj} (t)\label{eqn:IRMMV_dLdgv2} \\
    &=-4g_l^{}(t)\sum\limits_{j\in[L]}^{} \Lambda_{lj}(t)V_{lj} (t)\label{eqn:IRMMV_dLdgv3}.
\end{align}
In vector form, the gradient with respect to $\mathbf{g}$ is given by:
\begin{align}
    \nabla_{\mathbf{g}(t)}\mathcal{L}(\mathbf{g}(t), \mathbf{V}(t)) =  - 4\mathbf{g}(t)\odot\left( (\boldsymbol{\Lambda}(t)\odot\mathbf{V}(t))\mathbf{1}_L^\top\right).
\end{align} 
{Similarly, we compute the gradient of the loss function with respect to the element $V_{lm}^{}(t)$} (the $(l,m)$-th element of $\mathbf{V}(t))$:
\begin{align}
    \nabla_{V_{lm}^{}(t)}&\mathcal{L}(\mathbf{g}(t), \mathbf{V}(t)) = \nabla_{V_{lm}(t)^{}}\sum\limits_{i\in[M]}^{}\sum\limits_j^{} \left(Y_{ij} - \sum\limits_{k\in[N]}^{}A_{ik}g_k^{2}(t)V_{kj}(t)\right)^2_{}\\
    &= -2 \sum\limits_{i\in[M]}^{} \left(Y_{im} - \sum\limits_{k\in[N]}^{}A_{ik}g_k^{2}(t)V_{km}(t)\right)A_{il}g_l^2(t)\\
    &= -2 g_l^2(t)\sum\limits_{i\in[M]}^{}A^{}_{il} \left(Y_{im} - \sum\limits_{k\in[N]}^{}A_{ik}g_k^{2}(t)V_{km}(t)\right)\label{eqn:IRMMV_dLdvv0}\\
    &= -2 g_l^2(t)\left(\mathbf{A}^{\top}\left(\mathbf{Y} - \mathbf{A}(\mathbf{g}^{\odot2}(t)\mathbf{1}_L\odot\mathbf{V}(t)\right)\right)_{lm}^{} \label{eqn:IRMMV_dLdvv1}\\
    &= -2 g_l^2(t)\Lambda_{lm}^{} (t)\label{eqn:IRMMV_dLdvv2}.
\end{align}
In matrix form this can be written as:
\begin{align}
 \nabla_{\mathbf{V}(t)}\mathcal{L}(\mathbf{g}(t), \mathbf{V}(t)) = -2 \mathbf{g}^{\odot 2}_{}(t)\mathbf{1}_L^{}\odot\boldsymbol{\Lambda}(t)
\end{align}
\section{Characterization of the gradient flow}
\label{app:CharacterizationOfGradientFlow}
{This section provides the lemmas and proofs that characterize the behavior of the system under the continuous gradient flow. In the following lemma, we establish a key ``balancedness" property that is conserved by the gradient flow dynamics. Throughout this work, we denote the $i$-th row of any matrix $\mathbf{Z}$ as $\mathbf{Z}_{i:}^{}$ and let $\mathbf{1}_{N\times L}$ denote a matrix of ones of size $N \times L$.}
\begin{lemma}[Lemma \ref{lemma:MMV_Balancedness} restated]
\label{lemma:MMV_BalancednessApp}
For the system evolving under continuous gradient flow for the set of update equations \eqref{eqn:IRMMV_gUpdate}, \eqref{eqn:IRMMV_vUpdate} and \eqref{eqn:IRMMV_xUpdate}, derived from the loss function $\mathcal{L}(\mathbf{g}(t), \mathbf{V}(t))=\Vert \mathbf{Y}-\mathbf{A}(\mathbf{g}(t)^{\odot2}_{}\mathbf{1}_L\odot\mathbf{V}(t))\Vert _{F}^{2}$, the following properties hold for any time $t \ge 0$:
\begin{enumerate}
    \item Global balancedness: The quantity $\frac{1}{2}\left\Vert \mathbf{g}(t)\right\Vert_{2}^{2} -\left\Vert \mathbf{V}(t)\right\Vert_{F}^{2} $ is conserved throughout the optimization process:
    $$ \frac{1}{2}\Vert \mathbf{g}(t)\Vert _{2}^{2}-\Vert \mathbf{V}(t)\Vert _{F}^{2}=\frac{1}{2}\Vert \mathbf{g}(0)\Vert _{2}^{2}-\Vert \mathbf{V}(0)\Vert _{F}^{2} $$
    \item  Row-wise balancedness: For each individual row $i \in [N]$, the quantity $\frac{1}{2}g_{i}^{2}(t)-\sum_{j\in[L]}V_{ij}^{2}(t)$ is also conserved:
    $$ \frac{1}{2}g_{i}^{2}(t)-\sum_{j\in[L]}V_{ij}^{2}(t)=\frac{1}{2}g_{i}^{2}(0)-\sum_{j\in[L]}V_{ij}^{2}(0) $$
\end{enumerate}

\end{lemma}
\begin{proof}
Let $\boldsymbol{\Lambda}(t) := \mathbf{A}^{\top}\left(\mathbf{Y} - \mathbf{A}(\mathbf{g}^{\odot2}(t)\mathbf{1}_L\odot\mathbf{V}(t))\right)$. 
Gradient flow is related to gradient as follows:
\begin{align}
    \frac{\mathrm{d}}{\mathrm{d}t} {g}_l^{}(t) &= - \nabla_{g_{l}^{}}^{} \mathcal{L}(\mathbf{g}(t), \mathbf{V}(t)).\\
    \intertext{From \eqref{eqn:IRMMV_dLdgv3},}
    \frac{\mathrm{d}}{\mathrm{d}t} {g}_l^{}(t)&= 4g_l^{}(t)\sum\limits_{j\in[L]}^{} \Lambda_{lj}(t)V_{lj} (t)\label{eqn:MMV_gradgl}.\\
    \intertext{Similarly, from \eqref{eqn:IRMMV_dLdvv2},}
    \frac{\mathrm{d}}{\mathrm{d}t} V_{lm}^{}(t) &=  -\nabla_{V_{lm}^{}}^{} \mathcal{L}(\mathbf{g}(t), \mathbf{V}(t)),\\
    & = 2g_l^{2}(t)\Lambda_{lm}(t) \label{eqn:IRMMV_dVdtv1}.\\
    \frac{\mathrm{d}}{\mathrm{d}t} \Vert \mathbf{V}_{l:}^{}(t)\Vert _2^2
    &= \frac{\mathrm{d}}{\mathrm{d}t} \Bigg(\sum\limits_{m\in[L]}{V}_{lm}^2(t)\Bigg) =\sum\limits_{m\in[L]}\frac{\mathrm{d}}{\mathrm{d}t} {V}_{lm}^2(t)  \\
    &= 2 \sum\limits_{m\in[L]}{V}_{lm}(t)\frac{\mathrm{d}}{\mathrm{d}t} {V}_{lm}(t).\\
    \intertext{Substituting \eqref{eqn:IRMMV_dVdtv1}}
    \frac{\mathrm{d}}{\mathrm{d}t} \Vert \mathbf{V}_{l:}^{}(t)\Vert _2^2 
    &= 4 g_l^2(t) \sum\limits_{m\in[L]}{V}_{lm}(t)  \Lambda_{lm}^{}(t) .\label{eqn:IRMMV_dNormvdt_v1}
\end{align}
\begin{align}
\intertext{From \eqref{eqn:MMV_gradgl} and \eqref{eqn:IRMMV_dNormvdt_v1},}
g_l(t)\frac{\mathrm{d}}{\mathrm{d}t} g_l^{}(t) = \frac{\mathrm{d}}{\mathrm{d}t} \Vert \mathbf{V}_{l:}(t)\Vert _2^2\\
\frac{1}{2}\frac{\mathrm{d}}{\mathrm{d}t}g_l^{2}(t) = \frac{\mathrm{d}}{\mathrm{d}t} \Vert \mathbf{V}_{l:}(t)\Vert _2^2\label{eqn:MMV_unbalancednessDer_rowWise}\\
\frac{1}{2}\sum\limits_{l\in[N]}\frac{\mathrm{d}}{\mathrm{d}t} g_l^{2}(t)  = \sum\limits_{l\in[N]}\frac{\mathrm{d}}{\mathrm{d}t} \Vert \mathbf{V}_{l:}(t)\Vert _2^2\\
\frac{1}{2}\frac{\mathrm{d}}{\mathrm{d}t} \Vert \mathbf{g}(t)\Vert ^2_2 = \frac{\mathrm{d}}{\mathrm{d}t} \Vert \mathbf{V}(t)\Vert _F^2 \label{eqn:MMV_unbalancednessDerFull}.
\end{align}
Here, $\mathbf{V}_{l:}^{} (t)$ is the $l$-th row of matrix $\mathbf{V}(t)$. 
{Integrating both sides of Equation \eqref{eqn:MMV_unbalancednessDerFull} with respect to $t$ from $0$ to $t$ directly yields the relationship:} 
\begin{align}
    \frac{1}{2}\left(\Vert \mathbf{g}(t)\Vert ^2_2 - \Vert \mathbf{g}(0)\Vert ^2_2 \right) = \Vert \mathbf{V}(t)\Vert _F^2 - \Vert \mathbf{V}(0)\Vert _F^2.
\end{align}
Rearranging we get that
\begin{align}
           \frac{1}{2}\Vert \mathbf{g}(t)\Vert ^2_2- \Vert \mathbf{V}(t)\Vert _F^2 = \frac{1}{2}\Vert \mathbf{g}(0)\Vert ^2_2 - \Vert \mathbf{V}(0)\Vert _F^2 .
        \end{align}
        The second result follows from integrating both sides of \eqref{eqn:MMV_unbalancednessDer_rowWise} with respect to $t$ from $0$ to $t$.     
\end{proof}
\begin{remark}[Remark \ref{remark:epsilonrEpsilonRelation} restated]
In the general case, $\epsilon$ and $\epsilon_{r}$ are independent quantities. However, under the specific initialization of Algorithm 1 (where $g(0)$ and $V(0)$ are initialized with constant values $\alpha_{g}$ and $\alpha_{V}$), it strictly holds that $\epsilon_{r} \leq \epsilon$ for all $t \geq 0$.
\end{remark}

\begin{proof} \label{proofOfRemark}Recall that the algorithm initializes the parameters as constants across all indices: $g_{i}(0) = \alpha_{g}$ for all $i \in [N]$ and $V_{ij}(0) = \alpha_{V}$ for all $i \in [N], j \in [L]$.

Let $e_{i}(t)$ denote the unbalancedness of the $i$-th row at time $t$:
\begin{align}
    e_{i}(t) = \frac{1}{2}g_{i}^{2}(t) - \sum_{j \in [L]}V_{ij}^{2}(t).
\end{align}
At initialization ($t=0$), we have:
\begin{align}
    e_{i}(0) = \frac{1}{2}\alpha_{g}^{2} - L\alpha_{V}^{2} = c, \quad \forall i \in [N],
\end{align}
where $c$ is a constant scalar. Lemma \ref{lemma:MMV_BalancednessApp} demonstrates that this row-wise unbalancedness is conserved throughout the gradient flow. Therefore, $e_{i}(t) = e_{i}(0) = c$ for all $t \geq 0$.

Substituting this into the definition of the global unbalancedness constant $\epsilon$ (Definition \ref{def:globalBalancedness}), we obtain:
\begin{align}
    \epsilon(t) = \left\vert \sum_{i=1}^{N} e_{i}(t) \right\vert = \left\vert \sum_{i=1}^{N} c \right\vert = N\vert c\vert.
\end{align}
Similarly, for the row-unbalancedness constant $\epsilon_{r}$ (Definition \ref{def:row:balancedness}), we have:
\begin{align}
    \epsilon_{r}(t) = \max_{i \in [N]} \vert e_{i}(t)\vert  = \max_{i \in [N]} \vert c\vert  = \vert c\vert .
\end{align}
Comparing the two quantities, since $N \geq 1$:
\begin{align}
    \epsilon(t) = N \epsilon_{r}(t) \implies \epsilon_{r}(t) \leq \epsilon(t).
\end{align}
Finally, since these quantities are conserved throughout time, the time indexing $(t)$ can be dropped, yielding:
\begin{align}
    \epsilon_{r} \leq \epsilon.
\end{align}
\end{proof}
\begin{lemma}
\label{lemma:MMV_rowBalancedness}
For the system evolving under continuous gradient flow, as described in Lemma \ref{lemma:MMV_Balancedness}, with $\mathbf{\Lambda}(t):=\mathbf{A}^{\top}(\mathbf{Y}-\mathbf{A}(\mathbf{g}^{\odot2}(t)\mathbf{1}_{L}\odot \mathbf{V}(t)))$ and any row $l \in [N]$, the following relationship hold for any time $t \ge 0$: 
 \begin{align}
    \frac{1}{2} \frac{\mathrm{d}}{\mathrm{d}t} g_l^{2}(t) &= \sum\limits_{j\in[L]} \frac{\mathrm{d}}{\mathrm{d}t} V_{lj}^{2}(t) = 4\sum\limits_{j\in[L]}^{} (\mathbf{\Lambda}(t)\odot \mathbf{X}(t))_{lj} = 4 \left(\mathbf{\Lambda}(t) \mathbf{X}^\top_{}(t)\right)_{ll}^{}
\end{align}    

\end{lemma}
\begin{proof}
{
    \begin{align}
    \intertext{From \eqref{eqn:IRMMV_dNormvdt_v1} and \eqref{eqn:MMV_unbalancednessDer_rowWise},} 
    \frac{1}{2}\frac{\mathrm{d}}{\mathrm{d}t} g_l^{2}(t) = \frac{\mathrm{d}}{\mathrm{d}t} \Vert \mathbf{V}_{l:}(t)\Vert _2^2 &= 4 g_l^2(t) \sum\limits_{j\in[L]}{V}_{lj}(t)  \Lambda_{lj}^{}(t) \label{eqn:IRMMV_derivativeequalitygnVV0}\\
    &= 4 \sum\limits_{j\in[L]}{\Lambda}_{lj}(t)  X_{lj}^{}(t) \label{eqn:IRMMV_derivativeequalitygnVV1}.
\end{align}
The last line follows from rearranging and substituting for $X_{lj}^{}(t) = g_{l}^2(t)V_{lj}^{}(t)$ in \eqref{eqn:IRMMV_derivativeequalitygnVV0}.\\
In matrix form, equation \eqref{eqn:IRMMV_derivativeequalitygnVV1} can be written as $4\sum\limits_{j\in[L]}^{} (\mathbf{\Lambda}(t)\odot \mathbf{X}(t))_{lj} $ or $ 4 \left(\mathbf{\Lambda}(t) \mathbf{X}^\top_{}(t)\right)_{ll}^{}$.} 
\end{proof}
\begin{theorem}[Theorem \ref{thm:MMV_gradientFlowCompleteInequalityRowNormMainDoc} restated]
\label{thm:MMV_gradientFlowCompleteInequalityRowNorm}

For the system evolving under continuous gradient flow with updates as in \eqref{eqn:IRMMV_gUpdate}, \eqref{eqn:IRMMV_vUpdate} and \eqref{eqn:IRMMV_xUpdate} with initialization $g(0)=\alpha_{g}\mathbf{1}_{L}$ and $V(0)=\alpha_{V}\mathbf{1}_{N\times L}$ where $\alpha_g$ and $\alpha_V$ are small positive scalars. Let $\boldsymbol{\lambda}_{i}(t)$ denote the $i$-th row of $\mathbf{\Lambda}(t)$, and {
\begin{align}\hat{\mathbf{x}}_{i}(t) := \begin{cases} \frac{\mathbf{X}_{i:}^{}(t)}{\Vert \mathbf{X}_{i:}^{}(t)\Vert _{2}} & \text{if } \Vert \mathbf{X}_{i:}^{}(t)\Vert _{2} \neq 0 \\ \mathbf{0} & \text{if } \Vert \mathbf{X}_{i:}^{}(t)\Vert _{2} = 0 \end{cases}.
\end{align} Then, for any time $t \ge 0$ and for all $i \in [N]$ where $g_i^2(t) > 0$ and $\sum\limits_{m\in[N]}V_{im}^2 (t)> 0$,}\\
if $\left<\boldsymbol{\lambda}_{i}^{}(t),\hat{\mathbf{x}}_i^{} (t)\right> \geq 0$, 
\begin{align}
    \frac{\mathrm{d}}{\mathrm{d}t} \Vert \mathbf{X}_{i:}^{}(t)\Vert _2 
      & \leq  24\left<\boldsymbol{\lambda}_{i}^{}(t),\hat{\mathbf{x}}_i^{}(t) \right>\left(\epsilon + \Vert \mathbf{X}_{i:}^{}(t)\Vert _2^{2/3}\right)^2_{}\\
     \frac{\mathrm{d}}{\mathrm{d}t} \Vert \mathbf{X}_{i:}^{}(t)\Vert _2  &\geq  {6}\left<\boldsymbol{\lambda}_i^{}(t) \hat{\mathbf{x}}^{}_{i}(t)\right> \left(\frac{{\Vert \mathbf{X}_{i:}^{}(t)\Vert _2^2}}{\epsilon_{}^{} + \Vert \mathbf{X}_{i:}^{}(t)\Vert _2^{2/3}} \right),
\end{align}
if $\left<\boldsymbol{\lambda}_{i}^{}(t),\hat{\mathbf{x}}_i^{} (t)\right> < 0
$, 
\begin{align}
     \frac{\mathrm{d}}{\mathrm{d}t} \Vert \mathbf{X}_{i:}^{}(t)\Vert _2 
      & \geq  24\left<\boldsymbol{\lambda}_{i}^{}(t),\hat{\mathbf{x}}_i^{} (t)\right>\left(\epsilon + \Vert \mathbf{X}_{i:}^{}(t)\Vert _2^{2/3}\right)^2_{}\\
     \frac{\mathrm{d}}{\mathrm{d}t} \Vert \mathbf{X}_{i:}^{}(t)\Vert _2  &\leq  {6}\left<\boldsymbol{\lambda}_{i}^{}(t), \hat{\mathbf{x}}^{}_{i}(t)\right> \left(\frac{{\Vert \mathbf{X}_{i:}^{}(t)\Vert _2^2}}{\epsilon_{}^{} + \Vert \mathbf{X}_{i:}^{}(t)\Vert _2^{2/3}} \right).
\end{align}

\end{theorem}
\begin{proof}
{Note that 
\begin{align}
     \left<\boldsymbol{\lambda}_i(t), \hat{\mathbf{x}}_i^{}(t)\right>  =  \left(\boldsymbol{\Lambda}(t) \frac{\mathbf{X}^\top_{}(t)}{\Vert \mathbf{X}_{i:}^{}(t)\Vert _2}\right)_{ii}^{}. \label{eqn:IRMMV_innerProductLambdahatx}
\end{align}
With this identity established, we now derive the dynamics of the row norm $\Vert \mathbf{X}_{i:}^{}(t)\Vert _2$:}
    \begin{align}
    \frac{\mathrm{d}}{\mathrm{d}t} \Vert \mathbf{X}_{i:}^{}(t)\Vert _2 &= \frac{1}{2\Vert \mathbf{X}_{i:}^{}(t)\Vert _2}\frac{\mathrm{d}}{\mathrm{d}t} \left(\sum\limits_j^{} X_{ij}^2(t)\right).\label{eqn:IRMMV_LemmaB3V1}
    \intertext{{Recall that $X_{ij}^{}(t) = g_i^2(t)V_{ij}^{}(t)$. Using Lemma \ref{lemma:MMV_rowBalancedness} and equation \eqref{eqn:MMV_gradgl}, we have}}
   \frac{\mathrm{d}}{\mathrm{d}t} \sum\limits_jX_{ij}^{2}(t) & = 4g_i^4 (t)(\mathbf{\Lambda}(t) \mathbf{X}^\top_{}(t))_{ii}^{} +  16\sum\limits_{j\in[L]}V_{ij}^{2}(t)g_i^2(t)\sum\limits_k\Lambda_{ik}^{}(t)X_{ik}^{}(t)\\ 
    & = 4\left(g_i^4(t) (\mathbf{\Lambda}(t) \mathbf{X}^\top_{}(t))_{ii}^{} +  4\sum\limits_{j\in[L]}V_{ij}^{2}(t)g_i^2(t)(\mathbf{\Lambda} (t)\mathbf{X}^\top_{}(t))_{ii}^{}\right)\\ 
    & = 4(\mathbf{\Lambda}(t) \mathbf{X}^\top_{}(t))_{ii}^{} \left(g_i^4(t)  +  4\sum\limits_{j\in[L]}V_{ij}^{2}(t)g_i^2(t)\right)\label{eqn:MMV_growthflowEqualityrow}
\intertext{Substituting equation \eqref{eqn:MMV_growthflowEqualityrow} in \eqref{eqn:IRMMV_LemmaB3V1}, }
    \frac{\mathrm{d}}{\mathrm{d}t} \Vert \mathbf{X}_{i:}^{}(t)\Vert _2
    &= \frac{2}{\Vert \mathbf{X}_{i:}^{}(t)\Vert _2}(\mathbf{\Lambda}(t) \mathbf{X}^\top_{}(t))_{ii}^{} \left(g_i^4(t)  +  4\sum\limits_{j\in[L]}V_{ij}^{2}(t)g_i^2(t)\right)\label{eqn:IRMMV_DerivativeRowNormV2}\\
    & \overset{a}{\leq}  24\sum\limits_{j\in[L]}\Lambda_{ij}^{}(t)\frac{{X}_{ij}^{}(t)}{\Vert \mathbf{X}_{i:}^{}(t)\Vert _2} \left(\epsilon + \Vert \mathbf{X}_{i:}^{}(t)\Vert _2^{2/3}\right)^2_{}\label{eqn:IRMMV_DerivativeRowNormUpperV1}\\
      & \leq  24\left<\boldsymbol{\lambda}_{i}^{}(t),\hat{\mathbf{x}}_i^{}(t) \right>\left(\epsilon + \Vert \mathbf{X}_{i:}^{}(t)\Vert _2^{2/3}\right)^2_{}.
\intertext{{Note that in this case, $\left<\boldsymbol{\lambda}_i(t), \hat{\mathbf{x}}(t)_i^{}\right>$ is non-negative. 
From the definition of unbalancedness, we have $\left\vert \frac{1}{2}{g_i^2}(t)  - \sum\limits_{m\in[L]}V_{im}^2(t) \right\vert  \leq \epsilon_r^{}$.} The inequality in step $(a)$ 
follows by substituting  \eqref{eqn:IRMMV_innerProductLambdahatx}, \eqref{eqn:IRMMV_SumV2Inequality} and \eqref{eqn:IRMMV_g2Inequality} in \eqref{eqn:IRMMV_DerivativeRowNormV2}.} 
\sum\limits_{m\in[L]}^{}V_{im}^2(t)&\leq \epsilon_r^{} + \text{min}\left\{\sum\limits_{m\in[L]}^{}V_{im}^2(t), \frac{1}{2}g_i^2(t), \frac{1}{2}g_i^2(t)\right\}\\
&\leq \epsilon_r^{} + \left(\frac{1}{4}g_i^4(t)\sum\limits_{m\in[L]}^{}V_{im}^2(t)\right)^{\frac{1}{3}}_{}.\\ 
\intertext{Recall that $X_{im}^{}(t) = g_{i}^2(t)V_{im}^{}(t)$:}
\sum\limits_{m\in[L]}^{}V_{im}^2(t)&\leq \epsilon + \Vert \mathbf{X}_{i:}^{}(t)\Vert _2^{2/3}\label{eqn:IRMMV_SumV2Inequality}.\\
\intertext{Similarly, we have:}
\frac{1}{2}g_i^2(t)&\leq \epsilon + \Vert \mathbf{X}_{i:}^{}(t)\Vert _2^{2/3} \label{eqn:IRMMV_g2Inequality}.
\end{align}
For the lower bound, we start from \eqref{eqn:IRMMV_DerivativeRowNormV2} and rewrite the expression as follows:
\begin{align}
    \frac{\mathrm{d}}{\mathrm{d}t} \Vert \mathbf{X}_{i:}^{}(t)\Vert _2 & =   \frac{2}{\Vert \mathbf{X}_{i:}^{}(t)\Vert _2}(\mathbf{\Lambda}(t) \mathbf{X}^\top_{}(t))_{ii}^{} \left(g_i^4(t)  +  4\sum\limits_{j\in[L]}V_{ij}^{2}(t)g_i^2(t)\right)\label{eqn:IRMMV_dNormXdtV1}\\
     & =  \frac{2}{\Vert \mathbf{X}_{i:}^{}(t)\Vert _2}(\mathbf{\Lambda}(t) \mathbf{X}^\top_{}(t))_{ii}^{} \BiggFo(\frac{g_i^4(t)\sum\limits_{j\in[L]}V_{ij}^{2}(t)}{\sum\limits_{j\in[L]}V_{ij}^{2}(t)}+  4\frac{\sum\limits_{j\in[L]}V_{ij}^{2}(t)g_i^4(t)}{g_i^2(t)}\BiggFo)\\
     & =  \frac{2}{\Vert \mathbf{X}_{i:}^{}(t)\Vert _2}(\mathbf{\Lambda}(t) \mathbf{X}^\top_{}(t))_{ii}^{} \sum\limits_{j\in[L]} X_{ij}^2(t)\BiggFo(\frac{1}{\sum\limits_{j\in[L]}V_{ij}^{2}(t)}+ 2\frac{1}{\frac{1}{2}g_i^2(t)}\BiggFo).\\
     \intertext{{Substituting the inequalities  \eqref{eqn:IRMMV_SumV2Inequality} and \eqref{eqn:IRMMV_g2Inequality}, }}
     \frac{\mathrm{d}}{\mathrm{d}t} \Vert \mathbf{X}_{i:}^{}(t)\Vert _2 & \geq  {6}\left<\boldsymbol{\lambda}_i(t), \hat{\mathbf{x}}^{}_{i}(t)\right> \left(\frac{{\Vert \mathbf{X}_{i:}^{}(t)\Vert _2^2}}{\epsilon_{}^{} + \Vert \mathbf{X}_{i:}^{}(t)\Vert _2^{2/3}} \right).\label{eqn:IRMMV_DerivativeRowNormLowerV1}
\end{align}
When $\left<\boldsymbol{\lambda}_i, \hat{\mathbf{x}}_i^{}\right> < 0 $, the direction of inequality in \eqref{eqn:IRMMV_DerivativeRowNormUpperV1} and \eqref{eqn:IRMMV_DerivativeRowNormLowerV1} reverses resulting in the second part of the lemma. 
\end{proof}

Now, in the following lemma we look at a special case when $\epsilon = 0$
\begin{corollary}[Corollary \ref{corr:IRMMV_EvolutionOfComponents} restated]
\label{corr:IRMMV_EvolutionOfComponentsApp}
Assume that the system evolves under continuous gradient flow, as described in Lemma \ref{thm:MMV_gradientFlowCompleteInequalityRowNorm}, with perfect row balancedness (i.e., $\frac{1}{2}g_{i}^{2}(t)=\sum_{j}V_{ij}^{2}(t)$ $\forall i \in [N]$ and time $t \ge 0$). The rate of change of the Euclidean norm of the $i$-th row of $\mathbf{X}(t)$ is characterized by:
    \begin{align}
        \frac{\mathrm{d}}{\mathrm{d}t} \Vert \mathbf{X}_{i:}^{}(t)\Vert _2 & =   {2^{2/3}_{}\cdot 6}\left<\boldsymbol{\lambda}_i(t), \hat{\mathbf{x}}^{}_{i}(t)\right> \Vert \mathbf{X}_{i:}^{}(t)\Vert _2^{4/3}
    \end{align}
\end{corollary}
\begin{proof}

\begin{align}
\intertext{Rewriting \eqref{eqn:IRMMV_DerivativeRowNormV2},}
     \frac{\mathrm{d}}{\mathrm{d}t} \Vert \mathbf{X}_{i:}^{}(t)\Vert _2 & =   \frac{2}{\Vert \mathbf{X}_{i:}^{}(t)\Vert _2}(\mathbf{\Lambda}(t) \mathbf{X}^\top_{}(t))_{ii}^{} \left(g_i^4(t)  +  4\sum\limits_{j\in [L]}V_{ij}^{2}(t)g_i^2(t)\right).\\
\intertext{Substituting perfect-row balancedness condition $\frac{1}{2}g_i^2(t) = \sum\limits_{j\in [L]}V_{ij}^2(t)$:}
     \frac{\mathrm{d}}{\mathrm{d}t} \Vert \mathbf{X}_{i:}^{}(t)\Vert _2 & =   \frac{6}{\Vert \mathbf{X}_{i:}^{}(t)\Vert _2}(\mathbf{\Lambda} (t)\mathbf{X}^\top_{}(t))_{ii}^{}\,g_i^4(t) \overset{a}{=}   {2^{2/3}_{}6}\left<\boldsymbol{\lambda}_i(t), \hat{\mathbf{x}}^{}_{i}(t)\right> \Vert \mathbf{X}_{i:}^{}(t)\Vert _2^{4/3},
\end{align}

where equality (a) follows by substituting $g_i^4(t) = 2^{2/3}\Vert \mathbf{X}_{i:}^{}(t)\Vert _2^{4/3}$ since $\Vert \mathbf{X}_{i:}^{}(t)\Vert _2^2 = \frac{1}{2}g_i^6(t)$, and recalling $\langle \boldsymbol{\lambda}_i(t), \hat{\mathbf{x}}_i(t) \rangle = \frac{(\mathbf{\Lambda} (t)\mathbf{X}^\top(t))_{ii}}{\Vert \mathbf{X}_{i:}^{}(t)\Vert _2}$.


\end{proof}
\begin{lemma}\label{lemma:MMV_LipchitsAndContinousofLossFunction}
 For the system evolving under continuous gradient flow, the loss function $\mathcal{L}(\mathbf{g}(t),\mathbf{V}(t)) = \Vert \mathbf{Y}-\mathbf{A}(\mathbf{g}(t)^{\odot2}\mathbf{1}_{L}\odot \mathbf{V}(t))\Vert _{F}^{2}$ is locally Lipschitz continuous with respect to each component $g_{n}(t)$ (for all $n \in [N]$) and $V_{nl}(t)$ (for all $n \in [N], l \in [L]$) over a domain of interest $\mathcal{D}:=\{ (g_n^{}(t), V_{nl}^{}(t)) \in \mathbb{R}^{N} \times \mathbb{R}^{N \times L} : \vert g_{n}^{}(t)\vert  \le B_{g}, \vert V_{nl}^{}(t)\vert  \le B_{V} \forall\; n \in [N], l \in [L] \}$ for some constants $B_{g}>0$ and $B_{V}>0$. Furthermore, $\mathcal{L}(g(t),V(t))$ is uniformly continuous with respect to time $t$.

\end{lemma}
\begin{proof}
    We begin by proving the local Lipschitz continuity of the loss function. This requires bounding the term $\bigg\vert \mathcal{L}(\mathbf{g}(t), \mathbf{V}(t)) - \mathcal{L}\left(\tilde{\mathbf{g}}(t), \tilde{\mathbf{V}}(t)\right)\bigg\vert $ by the distance between the parameters $( \tilde{\mathbf{g}}(t),\tilde{\mathbf{V}}(t))$ and $\left( \mathbf{g}(t), {\mathbf{V}}(t)\right)$. Let the distance between the parameters at time $t$ be defined as

    \begin{align}
        \left\Vert  ( \tilde{\mathbf{g}}(t),\tilde{\mathbf{V}}(t)) \ -\left( \mathbf{g}(t), {\mathbf{V}}(t)\right)\right\Vert ^2 
        =
        \sum\limits_{j\in[L]}^{}\sum\limits_{k\in[N]}^{}\left(\left( \tilde{V}_{kj}^{}(t) -  {V}_{kj}^{}(t)\right)^2_{} +\left(\tilde{g}_k^{}(t)- g_k^{}(t)\right)^2_{}\right). \label{eqn:IRMMV_square_dist_appendix}
    \end{align}

   For brevity, we omit the explicit time dependency $(t)$ for variables $\mathbf{g}$, $\mathbf{V}$, $\mathbf{X}$, and $\mathbf{\Lambda}$ unless otherwise specified. 

    We now expand term $\bigg\vert \mathcal{L}(\mathbf{g}, \mathbf{V}) - \mathcal{L}\left(\tilde{\mathbf{g}}, \tilde{\mathbf{V}}\right)\bigg\vert $:
    \begin{align}
        \bigg\vert \mathcal{L}(\mathbf{g}, \mathbf{V}) - \mathcal{L}\left(\tilde{\mathbf{g}}, \tilde{\mathbf{V}}\right)\bigg\vert 
        &=  \BiggTh\vert \sum_{i\in[M]}\sum_{j\in[L]}\left(Y_{ij}^{}-\sum_{k\in[N]}^{}A_{ik}^{}g_k^2V_{kj}^{}\right)^2- \sum_{i\in[M]}\sum_{j\in[L]}\left(Y_{ij}^{}-\sum_{k\in[N]}^{}A_{ik}^{}\tilde{g}_k^2\tilde{V}_{kj}^{}\right)^2\BiggTh\vert \\
        &\leq  \sum_{i\in[M]}\sum_{j\in[L]}\BiggTh\vert \left(\sum_{k\in[N]}^{}A_{ik}^{}\Big(\tilde{g}_k^2\tilde{V}_{kj}^{} - g_k^2V_{kj}^{}\Big)\right) \left(2Y_{ij}^{}-\sum_{k\in[N]}^{}A_{ik}^{}\Big(g_k^2V_{kj}^{}+\tilde{g}_k^2\tilde{V}_{kj}^{}\Big)\right)\BiggTh\vert \\
        &\leq  2C_1^{}\sum_{i\in[M]}\sum_{j\in[L]}\BiggTh\vert \sum_{k\in[N]}^{}\Big(\tilde{g}_k^2\tilde{V}_{kj}^{} - g_k^2V_{kj}^{}\Big) \BiggTh\vert \label{eqn:IRMM_DiffLossgVtildegtildeVV1},
    \end{align}
    where $C_1^{}  = B_Y^{}+N\mu B_g^2B_V^{}$. Last line follows by noting that $g_i^{}, V_{ij}^{}\in\mathcal{D}$, and that $\mathbf{A}$ is $\mu$-coherent with $\ell_2$-normalized columns i.e., $\left\vert \sum\limits_{b\in[M]}^{}{A}^{}_{bn}{A}_{bc}^{}\right\vert  \leq \mu\leq 1$. {Let $B_Y^{} = \underset{i,j}{\text{max}}\left\{Y_{ij}^{}\right\}$}:
    \begin{align}
        \left\vert \tilde{g}_k^2\tilde{V}_{kj}^{} - g_k^2V_{kj}^{}\right\vert  &= \left\vert \tilde{g}_k^2\tilde{V}_{kj}^{} -  \tilde{g}_k^2{V}_{kj}^{} + \tilde{g}_k^2{V}_{kj}^{}- g_k^2V_{kj}^{}\right\vert \\
        &= \left\vert  \tilde{g}_k^2\left(\tilde{V}_{kj}^{} -  {V}_{kj}^{}\right) + {V}_{kj}^{}\left(\tilde{g}_k^2- g_k^{2}\right)\right\vert .\\
        \intertext{Substituting the parameter bounds from the domain $\mathcal{D}$}
        &\leq B_g^2\left\vert  \tilde{V}_{kj}^{} -  {V}_{kj}^{}\right\vert  + 2B_V^{}B_g^{}\left\vert \tilde{g}_k^{}- g_k^{}\right\vert   \\      
         &\leq C_2^{}\left(\left\vert  \tilde{V}_{kj}^{} -  {V}_{kj}^{}\right\vert  +\left\vert \tilde{g}_k^{}- g_k^{}\right\vert \right). \label{eqn:IRMMV_gk2Vkj}
    \end{align}
    where $C_2^{} = \text{max}\left\{1,B_g^2, 2B_V^{}B_g^{}\right\}$. Substituting \eqref{eqn:IRMMV_gk2Vkj} back into \eqref{eqn:IRMM_DiffLossgVtildegtildeVV1}, we can finally write: 
    \begin{align}
        \bigg\vert \mathcal{L}(\mathbf{g}, \mathbf{V}) - \mathcal{L}\left(\tilde{\mathbf{g}}, \tilde{\mathbf{V}}\right)\bigg\vert  
        &\leq 2C_1^{}C_2^{}M    \sum_{j\in[L]}\sum_{k\in[N]}^{}\left(\left\vert  \tilde{V}_{kj}^{} -  {V}_{kj}^{}\right\vert  +\left\vert \tilde{g}_k^{}- g_k^{}\right\vert \right).\\
        \intertext{Using the norm equivalence $\sum\limits_{i=1}^M\sum\limits_{j=1}^{N}\vert a_{ij}\vert \leq \sqrt{MN\sum\limits_{i=1}^M\sum\limits_{j=1}^{N}\vert a_{ij}\vert ^2}$:}
        \bigg\vert \mathcal{L}(\mathbf{g}, \mathbf{V}) - \mathcal{L}\left(\tilde{\mathbf{g}}, \tilde{\mathbf{V}}\right)\bigg\vert  
        &\leq 2C_1^{}C_2^{}M \sqrt{NL} \sqrt{  \sum_{j\in[L]}\sum_{k\in[N]}^{}\left(\left( \tilde{V}_{kj}^{} -  {V}_{kj}^{}\right)^2 +\left(\tilde{g}_k^{}- g_k^{}\right)^2\right)}\\
        &\leq 2C_1^{}C_2^{}M\sqrt{NL}    \left\vert  ( \tilde{\mathbf{g}},\tilde{\mathbf{V}}) \ -\left( \mathbf{g}, {\mathbf{V}}\right)\right\vert .
    \end{align}
    Thus, $\mathcal{L}(\mathbf{g}, \mathbf{V})$ is locally Lipchitz continuous \wrt $g_l^{}\ \forall \ l\in[N]$ and ${V}_{lm}^{} \ \forall\ l\in[N], \ m \in [L]$. 
    Now, we proceed to prove the uniform continuity of $\mathcal{L}(\mathbf{g}(t), \mathbf{V}(t))$ \wrt $t$. 
    \begin{align}
    \frac{\mathrm{d}}{\mathrm{d}t} \mathcal{L}(\mathbf{g}(t), \mathbf{V}(t)) &= \sum\limits_{n\in[N]}\nabla_{g_{n}^{}}\mathcal{L}(\mathbf{g}(t), \mathbf{V}(t)) \frac{\mathrm{d}}{\mathrm{d}t} g_{n}^{}(t) + \sum\limits_{n\in[N]}\sum\limits_{l\in[L]}\nabla_{V_{nl}^{}}\mathcal{L}(\mathbf{g}(t), \mathbf{V}(t)) \frac{\mathrm{d}}{\mathrm{d}t} V_{nl}^{}(t).\\
    \intertext{By substituting the gradient flow relations $\frac{\mathrm{d}}{\mathrm{d}t} g_i^{}(t) = - \nabla_{g_i^{}(t)}\mathcal{L}(\mathbf{g}(t), \mathbf{V}(t)) $ and $\frac{\mathrm{d}}{\mathrm{d}t}  V_{il}^{}(t) = - \nabla_{V_{il}^{}(t)}\mathcal{L}(\mathbf{g}(t), \mathbf{V}(t))$, this becomes:}
    \frac{\mathrm{d}}{\mathrm{d}t} \mathcal{L}(\mathbf{g}(t), \mathbf{V}(t))&= -\sum\limits_{n\in[N]}\left(\nabla_{g_{n}^{}}\mathcal{L}(\mathbf{g}(t), \mathbf{V}(t))\right)^2_{}-\sum\limits_{n\in[N]}\sum\limits_{l\in[L]}\left(\nabla_{V_{nl}^{}}\mathcal{L}(\mathbf{g}(t), \mathbf{V}(t))\right)^2_{}.\\
    \intertext{ Substituting equation \eqref{eqn:IRMMV_dLdgv3} and \eqref{eqn:IRMMV_dLdvv2} in the above equation, we have:}
    \frac{\mathrm{d}}{\mathrm{d}t} \mathcal{L}(\mathbf{g}(t), \mathbf{V}(t))    &= \sum\limits_{n\in[N]} \left( \left(4g_n^{}(t)\sum\limits_{l\in[L]}^{}\Lambda_{nl}(t)V_{nl}^{}(t)\right)^2 +\sum\limits_{l\in[L]}^{}4 g_n^4(t)\Lambda_{nl}^{2}(t)\right).
    \label{eqn:IRMMV_derivativewrttV1}
    \end{align}
     To complete the proof, we must show that the magnitude of this time derivative is bounded by a constant. First, we establish an upper bound for $\vert {\Lambda}_{nl}^{}(t)\vert ^{}$:
      \begin{align}
   \vert {\Lambda}_{nl}^{}(t)\vert ^{} &= \left\vert \left(\mathbf{A}^{\top}\left(\mathbf{Y} - \mathbf{A}{\mathbf{X}(t)}\right)\right)_{nl}^{}\right\vert \\
   &= \left\vert \sum\limits_{b\in[M]}^{}\left({A}^{}_{bn}\left({Y}_{bl}^{} - \sum\limits_{c\in[N]}^{}{A}_{bc}^{}{X}_{cl}^{}(t)\right)\right)\right\vert \\
   &\leq \left\vert \sum\limits_{b\in[M]}^{}{A}^{}_{bn}{Y}_{bl}^{}\right\vert  + \left\vert \sum\limits_{b\in[M]}^{}{A}^{}_{bn}\sum\limits_{c\in[N]}^{}{A}_{bc}^{}{X}_{cl}^{}(t)\right\vert .\\
   \intertext{Using $\mathbf{A}$ is $\ell_2$-normalized columns and ${X}_{cl}^{}(t) = {g}_c^{2}(t){V}_{cl}(t)$, we have:}
   \vert {\Lambda}_{nl}^{}\vert ^{}&\leq \left\vert \sum\limits_{b\in[M]}^{}{A}^{}_{bn}{Y}_{bl}^{}\right\vert  + \sum\limits_{c\in[N]}^{}\left\vert \sum\limits_{b\in[M]}^{}{A}^{}_{bn}{A}_{bc}^{}\right\vert \bigg\vert {g}_c^2(t){V}_{cl}^{}(t)\bigg\vert .\\
   \intertext{Substituting the parameter bounds from the domain of interest $\mathcal{D}$ in the above equation, we have:}
  \vert {\Lambda}_{nl}^{}\vert ^{}
  &\leq M B_Y^{} + N \mu   B_g^2B_V^{} \\
  &\leq M\left( B_Y^{} + N \mu   B_g^2B_V^{}\right).\label{eqn:IRMMV_B1NormLambda}
\end{align}
Substituting this bound along with the parameter bounds form $\mathcal{D}$ in \eqref{eqn:IRMMV_derivativewrttV1}, we obtain:
\begin{align}
          \left\vert \frac{\mathrm{d}}{\mathrm{d}t} \mathcal{L}(\mathbf{g}(t), \mathbf{V}(t))\right\vert &\leq \sum\limits_n\left( 16L^2_{}B_g^{2}B_V^{2} M^2_{}C_1^{2} + 4LB_g^4M^2_{}C_1^2\right)\\
          &\leq 4NM^2_{}L_{}C_2^{2}C_1^{2}\Bigg(L + \frac{1}{2}\Bigg)\label{eqn:IRMMV_AbsgradientFlowLoss}.
     \end{align}
    Since $M, N, L, B_g^{}, B_V^{}, C_1^{}, C_2^{}$ are all finite constants, the entire expression is bounded by a constant. Thus, $\mathcal{L}(\cdot,\cdot)$ is uniform \wrt $t$.
\end{proof}
\begin{lemma}
\label{lemma:IRMMV_ParametersStays0ifInitializedat0}
For the system evolving under continuous gradient flow, as described in Corollary \ref{corr:IRMMV_EvolutionOfComponentsApp}:\begin{enumerate}[label=\textbf{Case \Roman*:}, align=left]
    \item If $\frac{1}{\sqrt{2}}\vert g_i^{}(0)\vert  = \Vert V_{i:}^{}(0)\Vert  = 0
    $, then $\frac{1}{\sqrt{2}}\Vert g_i^{}(t)\Vert  = \Vert V_{i:}(t)\Vert  = 0$ for all $t>0$.
    \item If $\frac{1}{\sqrt{2}}\vert g_i^{}(0)\vert  = \Vert V_{i:}^{}(0)\Vert  > 0
    $, then $\frac{1}{\sqrt{2}}\vert g_i^{}(t)\vert  = \Vert V_{i:}(t)\Vert  > 0$ for all $t>0$.
\end{enumerate}  
    
\end{lemma}
\begin{proof}
    
    
    \textbf{Case I:} $\frac{1}{\sqrt{2}}\vert g_i^{}(0)\vert  = \Vert V_{i:}^{}(t)(0)\Vert  = 0
    $; We also have $V_{il}^{}(0) = 0\ \forall \ l\in[L]$. \\
    {Recall that gradient flow is related to gradient as:}
    \begin{align}\frac{\mathrm{d}}{\mathrm{d}t} g_i^{}(t) = - \nabla_{g_i^{}(t)}\mathcal{L}(\mathbf{g}(t), \mathbf{V}(t)) &= {4g_i^{}\sum\limits_j^{} \Lambda_{ij}V_{ij}}, \nonumber\\
    \intertext{and}
    \frac{\mathrm{d}}{\mathrm{d}t}  V_{il}^{}(t) = - \nabla_{V_{il}^{}(t)}\mathcal{L}(\mathbf{g}(t), \mathbf{V}(t))&={2 g_i^2\Lambda_{il}^{}}.\end{align} This system forms an initial value problem in both $g_i^{}(t)$ and $V_{il}^{}(t)$. {Because the loss function gradient is locally Lipschitz continuous \wrt the parameters $(\mathbf{g}(t), \mathbf{V}(t))$ and uniformly continuous \wrt time $t$}, Lemma \ref{lemma:MMV_LipchitsAndContinousofLossFunction} and \citet[Theorem 2.2]{teschl2012ordinary}\footnote{\citep[Theorem 2.2]{teschl2012ordinary} (Picard-Lindelöf): Suppose $f \in C\left(U, \mathbb{R}^n\right)$, where $U$ is an open subset of $\mathbb{R}^{n+1}$, and $\left(t_0, x_0\right) \in U$. If $f$ is locally Lipschitz continuous in the second argument, uniformly with respect to the first, then there exists a unique local solution $\bar{x}(t) \in C^1(I)$ of the initial value problem, where I is some interval around $t_0$.}, {guarantees} a unique solution for $g_i^{}(t)$ and $V_{il}^{}(t)$.
    At $t=0$, 
    \begin{align}
    \frac{\mathrm{d}}{\mathrm{d}t} g_i^{}(t)\BiggTh\vert _{t=0} = 0 ;\;\;
    \frac{\mathrm{d}}{\mathrm{d}t}  V_{il}^{}(t)\BiggTh\vert _{t=0} =  0.\end{align}
    {The constant functions,  $g_i^{}(t) = 0, V_{il}(t) = 0$ form a valid solution to the initial value problem. Since the solution is unique, this must be the only solution.}

    {In our setting $g_i^{}(t)$ and $V_{i:}^{}(t)$ are continuous, $\frac{1}{\sqrt{2}}\vert g_i^{}(0)\vert  = \Vert V_{i:}^{}(0)\Vert  > 0$, and $\left<\boldsymbol{\lambda}_{i}(t), \hat{\mathbf{x}}_{i}(t)\right>$, is bounded because parameters lie in bounded domain.  To complete the proof, we now show that the differential equations governing these norms follow the format required by the cited lemma: }
    
    \begin{align}
        \frac{\mathrm{d}}{\mathrm{d}t} \vert g_i^{}(t)\vert  &= \frac{1}{2\vert g_i^{}(t)\vert } \frac{\mathrm{d}}{\mathrm{d}t} g_i^{2}(t); \;\frac{\mathrm{d}}{\mathrm{d}t} \Vert V_{i:}^{}(t)\Vert  = \frac{1}{2\Vert V_{i:}\Vert }\frac{\mathrm{d}}{\mathrm{d}t} \Vert V_{i:}^{}(t)\Vert ^2.\\
        \intertext{ Recalling the result from Lemma \ref{lemma:MMV_rowBalancedness}, {we have the dynamics of the norms in terms of $ \left(\mathbf{\Lambda} \mathbf{X}^\top_{}\right)_{ii}^{}$. By applying the innerproduct identity (equation \eqref{eqn:IRMMV_innerProductLambdahatx}), we have:}}
         \frac{\mathrm{d}}{\mathrm{d}t} \vert g_i^{}(t)\vert 
        &=\frac{4g_i^{2}(t)\Vert V_{i:}^{}(t)\Vert }{\vert g_i^{}(t)\vert }\left<\boldsymbol{\lambda}_{i}^{}(t), \hat{\mathbf{x}}_{i}^{}(t)\right>.\label{eqn:IRMM_dnormgdtv1}
\intertext{Similarly, we have}
        \frac{\mathrm{d}}{\mathrm{d}t} \Vert V_{i:}^{}(t)\Vert 
        &= 2\frac{g_i^{2}(t)\Vert V_{i:}^{}(t)\Vert }{\Vert V_{i:}^{}(t)\Vert }\left<\boldsymbol{\lambda}_{i}^{}(t), \hat{\mathbf{x}}_{i}^{}(t)\right>. \label{eqn:IRMM_dnormVdtv1}
    \end{align}
      Applying the perfect row-balancedness condition  (namely, $\frac{1}{2}g_i^{2}(t) = \sum\limits_jV_{ij}^2 = \Vert \mathbf{V}_{i:}^{}(t)\Vert ^2$) to equations \eqref{eqn:IRMM_dnormgdtv1} and \eqref{eqn:IRMM_dnormVdtv1} simplify to:
    \begin{align}
        \frac{\mathrm{d}}{\mathrm{d}t} \vert g_i^{}(t)\vert  &={2\sqrt{2} g_i^2(t)}{} \left<\boldsymbol{\lambda}_{i}^{}(t), \hat{\mathbf{x}}_{i}^{}(t)\right>,\\
        \intertext{and}
        \frac{\mathrm{d}}{\mathrm{d}t} \Vert V_{i:}^{}(t)\Vert  &=4{\Vert V_{i:}^{}(t)\Vert ^2_{}}{}\left<\boldsymbol{\lambda}_{i}^{}(t), \hat{\mathbf{x}}_{i}^{}(t)\right>. 
    \end{align}
    
    As all the necessary conditions of \citet[Lemma 5]{razin2021implicit} are met, we have $g_i^{}(t)>0$, $\Vert V_{i:}^{}(t)\Vert >0$.
\end{proof}



Now that we have characterized the fundamental dynamics of the gradient flow, in the next section, we develop the essential theoretical tools needed to prove our main result.
\section{Supporting Lemmas and the Proof of the Theorem \ref{thm:IRMMV_mainResult}}
\label{app:IRMMV_ProofSection}
We begin by defining the initialization schemes for both the estimated trajectory $\mathbf{X}\left(t\right)$, and some reference rank-$K$ trajectory $\tilde{\mathbf{X}}\left(t\right)$.
\subsubsection*{Initialization of estimated trajectory }
At time $t = 0$, the parameters are initialized uniformly irrespective of the sparsity support $\mathcal{S}_K^{}$, i.e., 
\begin{align}
    g_i(0) = \alpha_g^{}; \; V_{ij}(0)^{} = \alpha_V^{} \ \forall \ i\in [N] ,  \ j \in [L].
\end{align}
To ensure perfect row-balancedness (i.e.,$\frac{1}{2}g_i^{2}(t) = \sum\limits_jV_{ij}^2(t)$), at $t=0$, we keep $\frac{1}{2}\alpha_g^2 = {L}\alpha_V^{2}$.
This initialization results in the following results, at $t = 0$
\begin{align}
X_{ij}^{}(0) &= g_i^2(0)V_{ij}^{}(0)= \alpha_g^2\alpha_V^{} = 2L\alpha_V^3 = \frac{1}{\sqrt{2L}}\alpha_g^3
\end{align}
As a result the Euclidean norm of $i$-th row of the estimated trajectory at time $t = 0$ is:
\begin{align}
    \Vert \mathbf{X}_{i:}^{}(0)\Vert _2^{}& = g_i^2(0)\Vert \mathbf{V}_{i:}(0)\Vert _2^{}= \alpha_g^2\sqrt{L}\alpha_v^{} = \frac{1}{\sqrt{2}}\alpha_g^3= 2L\sqrt{L}\alpha_V^3.\label{eqn:IRMMV_X_0rowi}
\end{align}
\subsubsection*{Initialization of the reference trajectory}
Next, we construct a reference rank-$K$ trajectory, denoted as $\tilde{\mathbf{X}}\left(t\right)$. For a given support set $\mathcal{S}_K^{}$, we initialize its corresponding parameters
so that the rows outside the support are zero:
\begin{align}
\tilde{g}_{i}^{}(0) = \begin{cases}\sqrt{2}\rho^{1/3}_{}\ &\forall i \in \mathcal{S}_K^{}\\
0 &\forall i\notin \mathcal{S}_K^{}\end{cases}\quad; \;\;\quad\tilde{V}_{i,j}^{}(0) = \begin{cases}\frac{1}{\sqrt{L}}\rho^{1/3}_{}\ &\forall i\in\mathcal{S}_K^{}\\
0 &\forall i\notin \mathcal{S}_K^{}\end{cases},\ \forall \ j\in [N].\label{eqn:IRMMV_VreferenceInit}
\end{align}
Consequently, the initial row norms are given by

\begin{align}
\Vert \tilde{\mathbf{V}}_{i:}(0)\Vert _2 = \begin{cases}\rho^{1/3}_{}\ &\forall i \in \mathcal{S}_K^{}\\
0 &\forall i\notin \mathcal{S}_K^{}\end{cases} .\label{eqn:IRMMV_NormVreferenceInit}
\end{align}
Here $\rho >0$ is a very small positive constant. This construction ensures that the reference trajectory is perfectly row-balanced (i.e., $\frac{1}{2}g_i^{2}(t) = \sum\limits_jV_{ij}^2$), at initialization and rank-$K$ at initialization. Note that since $X_{ij}^{}(t) = g_i^2(t)V_{ij}^{}(t)$ the above initializations result in the following expression: 
\begin{align}\Vert \tilde{\mathbf{X}}_{i:}^{}(0)\Vert _2^{} = \begin{cases}
    2\rho\ &\forall \ i \in \mathcal{S}_K^{} ,\\
    0 \ &\forall \ i \notin \mathcal{S}_K^{}\label{eqn:IRMMV_NormXreferenceInit}
\end{cases}\end{align}
{Note that this construction is purely for this theoretical proof; the proposed algorithm does not assume any knowledge of the support set or sparsity.}




With the initializations established, the following lemma formalizes the dynamics of the constructed reference rank-$K$ trajectory. 
\begin{lemma}\label{lemma:MMV_rowGradientflowEvenWhenExtendedtoZeros}
Let a $K$-component system be defined by parameters $\bar{\mathbf{g}}(t) \in \mathbb{R}^{K}_{}$ and $\bar{\mathbf{V}}(t)\in\mathbb{R}^{K \times L}_{}$. Let $\mathcal{S}_K^{} \subseteq [N]$ be a support set of size $K = \vert \mathcal{S}_{K}^{}\vert  \leq N$. Let us construct an $N$-component trajectory $\tilde{\mathbf{g}}(t) \in \mathbb{R}^N_{}$ and $\tilde{\mathbf{V}}(t) \in \mathbb{R}^{N \times L}$ by embedding the $K$ - component system into a larger $N$-dimensional space and padding the remaining components with zeros. 
To formally relate the $K$-component system to its $N$-component embedding, we create an ordered correspondence between the indices of the small system, $\{1, ..., K\}$, and the ordered indices of the support set, $\mathcal{S}_K = \{s_1, s_2, ..., s_K\}$. The embedding is then defined as:
\begin{align}
    \tilde{g}_{i}(t) &:= \begin{cases} \bar{g}_{k}(t) & \text{if } i = s_k \text{ for } k \in [K] \\ 0 & \text{if } i \notin \mathcal{S}_{K} \end{cases} \label{eqn:IRMMV_tildegbargRelation}\\
    \tilde{V}_{il}(t) &:= \begin{cases} \bar{V}_{kl}(t) & \text{if } i = s_k \text{ for } k \in [K] \\ 0 & \text{if } i \notin \mathcal{S}_{K} \end{cases}\label{eqn:IRMMV_tildegbarVRelation}
\end{align}
Let $\bar{\mathbf{A}}$ be the sub-matrix of $\mathbf{A}$ such that $\bar{\mathbf{A}}$ contains only columns of $\mathbf{A}$ that correspond to the support indices $\mathcal{S}_K^{}$.
Let the dynamics of $\bar{\mathbf{g}}(t)$ and $\bar{\mathbf{V}}(t)$ be derived from the loss function $\mathcal{L}_K^{}(\bar{\mathbf{g}}(t), \bar{\mathbf{V}}(t))=\left\Vert\mathbf{Y}-\bar{\mathbf{A}}(\bar{\mathbf{g}}(t)^{\odot2}\mathbf{1}_{L}\odot \bar{\mathbf{V}}(t))\right\Vert_{F}^{2}$ where $\bar{\mathbf{X}}(t) = (\bar{\mathbf{g}}(t)^{\odot2}\mathbf{1}_{L})\odot \bar{\mathbf{V}}(t)$ follow gradient flow path. Then, the dynamics of $\tilde{\mathbf{g}}(t)$ and $\tilde{\mathbf{V}}(t)$, derived from the loss function $\mathcal{L}_N^{}(\tilde{\mathbf{g}}(t), \tilde{\mathbf{V}}(t))=\Vert\mathbf{Y}-{\mathbf{A}}(\tilde{\mathbf{g}}(t)^{\odot2}\mathbf{1}_{L}\odot \tilde{\mathbf{V}}(t))\Vert_{F}^{2}$ where $\tilde{\mathbf{X}}(t) = (\tilde{\mathbf{g}}(t)^{\odot2}\mathbf{1}_{L})\odot \tilde{\mathbf{V}}(t)$, also follow the gradient flow path. That is:
\begin{align}
\frac{\mathrm{d}}{\mathrm{d}t} \tilde{g}_{i}(t) &= -\nabla_{\tilde{g}_{i}^{}}\mathcal{L}(\tilde{\mathbf{g}}(t),\tilde{\mathbf{V}}(t)) ,\\
\quad \quad
 \frac{\mathrm{d}}{\mathrm{d}t} \tilde{V}_{il}(t) &= -\nabla_{\tilde{V}_{il}}\mathcal{L}(\tilde{\mathbf{g}}(t),\tilde{\mathbf{V}}(t)) .
 \end{align}
\end{lemma}
\begin{proof}

Let the system parameters $(\tilde{\mathbf{g}}(t), \tilde{\mathbf{V}}{}(t))$ be derived from the loss function $\mathcal{L}_N(\tilde{\mathbf{g}}(t), \tilde{\mathbf{V}}(t)):= \Vert\mathbf{Y} - {\mathbf{A}}(\tilde{\mathbf{g}}^{\odot2}(t)\mathbf{1}_L\odot \tilde{\mathbf{V}}(t))\Vert$. 

The objective of this proof is to demonstrate that the $N$-component trajectory $(\tilde{g}(t), \tilde{V}(t))$ (which is constructed by embedding a $K$-component system $(\bar{\mathbf{g}}(t), \bar{\mathbf{V}}(t))$ and padding it with zeros) follows the gradient flow dynamics derived from the $N$-component loss function, $\mathcal{L}_N(\tilde{\mathbf{g}}(t), \tilde{\mathbf{V}}(t))$.\\

The proof is divided into two parts, considering the components within the support set  ($i \in \mathcal{S}_K$) and and those outside of it ($i \notin \mathcal{S}_K$).\\

For the components within the support set ($i \in \mathcal{S}_K$): Let $k$ be the index such that $i$ is the $k$-th element in the ordered support set $\mathcal{S}_K$. Recall that the gradient flow for $\bar{g}_k^{}(t)$ is related to the gradient as:

\begin{align}
    \frac{\mathrm{d}}{\mathrm{d}t} \bar{g}_k^{}(t) &= -\nabla_{\bar{g}_k^{}(t)}\mathcal{L}_K^{}(\bar{\mathbf{g}}(t), \bar{\mathbf{V}}(t)) .
\end{align}
\begin{align}
    \intertext{Substituting the definition of $\mathcal{L}_{K}^{}(\cdot,\cdot)$:}
    \frac{\mathrm{d}}{\mathrm{d}t} \bar{g}_k^{}(t)
    &=-\nabla_{\bar{g}_k^{}(t)}\left(\sum\limits_m\sum\limits_l\left(({Y}_{ml}^{}-\sum\limits_{j\in[K]}\bar{{A}}_{mj}^{}\bar{{g}}_j^{2}(t)\bar{V}_{jl}(t)^{}\right)^2_{}\right) .\\
    \intertext{Taking the derivative:}
    \frac{\mathrm{d}}{\mathrm{d}t} \bar{g}_k^{}(t)
    &= 4 \bar
    {g}_k^{}(t)\sum\limits_m\sum\limits_l\bar{A}_{mk}^{}\left({Y}_{ml}^{}-\sum\limits_{j\in[K]}\bar{{A}}_{mj}^{}\bar{{g}}_j^{2}(t)\bar{V}_{jl}^{}(t)\right)\bar{V}_{kl}^{}(t),\\
    \intertext{By the definition of parameters, we can replace the $K$-component parameters $\bar{g}_k(t), \bar{V}_{kl}^{}(t)$ and the columns of $\bar{\mathbf{A}}$ by their corresponding $N$-component parameters $\tilde{g}_i(t), \tilde{V}_{il}^{}(t)$ and columns of $\mathbf{A}$.}
     \frac{\mathrm{d}}{\mathrm{d}t} \bar{g}_k^{}(t) &=4 \tilde
    {g}_i^{}(t)\sum\limits_m\sum\limits_l{A}_{mi}^{}\left({Y}_{ml}^{}-\sum\limits_{j\in\mathcal{S}_K^{}}{A}_{mj}^{}\tilde{{g}}_j^{2}(t)\tilde{V}_{jl}^{}(t)\right)\tilde{V}_{il}^{}(t) .\\
    \intertext{For any $j \notin \mathcal{S}_K^{}$, we have $\tilde{g}_j^{}(t) = 0, \tilde{V}_{jl}^{}(t) = 0$, hence we can also write it as}
     \frac{\mathrm{d}}{\mathrm{d}t} \bar{g}_k^{}(t)  &= 4 \tilde
    {g}_i^{}(t)\sum\limits_m\sum\limits_l{A}_{mi}^{}\left({Y}_{ml}^{}-\sum\limits_{j\in[N]}{A}_{mj}^{}\tilde{{g}}_j^{2}(t)\tilde{V}_{jl}^{}(t)\right)\tilde{V}_{il}^{}(t),\\
    \intertext{This expression is the negative partial derivative of the loss function ($\mathcal{L}_N^{}(\tilde{\mathbf{g}}(t), \tilde{\mathbf{V}}(t))$ \wrt $\tilde{g}_i^{}(t)$ i.e.,}
     \frac{\mathrm{d}}{\mathrm{d}t} \bar{g}_k^{}(t)   &= - \nabla_{\tilde{g}_i^{}(t)}\sum\limits_m\sum\limits_l\left({Y}_{ml}^{}-\sum\limits_{j\in[N]}{A}_{mj}^{}\tilde{{g}}_j^{2}(t)\tilde{V}_{jl}^{}(t)\right)^2_{},\\
    &= - \nabla_{\tilde{g}_i^{}(t)}\mathcal{L}_N^{}(\tilde{\mathbf{g}}(t), \tilde{\mathbf{V}}(t)). \label{eqn:IRMMV_GradientFlowActualLossreferenceRelation}\\
    \intertext{
By the definition \eqref{eqn:IRMMV_tildegbargRelation}, we have:}
    \frac{\mathrm{d}}{\mathrm{d}t} \tilde{g}_i^{}(t) &= \frac{\mathrm{d}}{\mathrm{d}t} \bar{g}_k(t) .\label{eqn:IRMMV_actualandreferenceparametersrelation}\\ 
\intertext{Thus combining the results from \eqref{eqn:IRMMV_GradientFlowActualLossreferenceRelation} and \eqref{eqn:IRMMV_actualandreferenceparametersrelation}, we have:} 
    \frac{\mathrm{d}}{\mathrm{d}t} \tilde{g}_i^{}(t) = - \nabla_{\tilde{g}_i^{}}\mathcal{L}_N^{}(\tilde{\mathbf{g}}(t), \tilde{\mathbf{V}}(t)).\label{eqn:IRMMV_GradientFlowNg1}
    \end{align}

Following the same line of reasoning, we begin with the gradient flow for $\bar{V}_{kl}^{}(t)$
\begin{align}
    \frac{\mathrm{d}}{\mathrm{d}t} \bar{V}_{kl}^{}(t)
    &= -\nabla_{\bar{V}_{kl}^{}(t)}\mathcal{L}_K^{}(\bar{\mathbf{g}}(t), \bar{\mathbf{V}}(t)),\\
    &=-\nabla_{\bar{V}_{kl}^{}}\bigg(\sum\limits_m\sum\limits_l\Big({Y}_{ml}^{}-\sum\limits_{j\in[K]}\bar{{A}}_{mj}^{}\bar{{g}}_j^{2}(t)\bar{V}_{jl}^{}(t)\Big)^2_{}\bigg) \\
    &=2 \bar
    {g}_k^{2}(t)\sum\limits_m\bar{A}_{mk}^{}\Big({Y}_{ml}^{}-\sum\limits_{j\in[K]}\bar{{A}}_{mj}^{}\bar{{g}}_j^{2}(t)\bar{V}_{jl}^{}(t)\Big) \\
    &=2 \tilde
    {g}_i^{2}(t)\sum\limits_m{A}_{mi}^{}\Big({Y}_{ml}^{}-\sum\limits_{j\in\mathcal{S}_K^{}}{A}_{mj}^{}\tilde{{g}}_j^{2}(t)\tilde{V}_{jl}^{}(t)\Big) .
    \end{align}
    The last line is obtained by replacing the $K$-component parameters by their $N$-component counterparts.
    \begin{align}
    \intertext{For any $j \notin \mathcal{S}_K^{}$, we have $\tilde{g}_j^{}(t) = 0,\tilde{V}_{jl}^{}(t) = 0$, hence we can also write the above as}
    \frac{\mathrm{d}}{\mathrm{d}t} \bar{V}_{kl}^{}(t)&= 2 \tilde
    {g}_i^{2}(t)\sum\limits_m{A}_{mi}^{}\Big({Y}_{ml}^{}-\sum\limits_{j\in[N]}{A}_{mj}^{}\tilde{{g}}_j^{2}(t)\tilde{V}_{jl}^{}(t)\Big),\\
    \intertext{This expression is the negative partial derivative of the loss function ($\mathcal{L}_N^{}(\tilde{\mathbf{g}}(t), \tilde{\mathbf{V}}(t))$ \wrt $\tilde{V}_{il}^{}(t)$ i.e.,}
    \frac{\mathrm{d}}{\mathrm{d}t} \bar{V}_{kl}^{}(t)
    &= - \nabla_{\tilde{V}_{il}^{}(t)}\sum\limits_m\sum\limits_l\Big({Y}_{ml}^{}-\sum\limits_{j\in[N]}{A}_{mj}^{}\tilde{{g}}_j^{2}(t)\tilde{V}_{jl}^{}(t)\Big)^2_{},\\
    &= - \nabla_{\tilde{V}_{il}^{}}\mathcal{L}_N^{}(\tilde{\mathbf{g}}(t), \tilde{\mathbf{V}}(t)).\label{eqn:IRMMV_GradientFlowActualLossreferenceRelationV}
\\
\intertext{By the definition \eqref{eqn:IRMMV_tildegbargRelation}, we have }
    \frac{\mathrm{d}}{\mathrm{d}t} \tilde{V}_{il}^{}(t) = \frac{\mathrm{d}}{\mathrm{d}t} \bar{V}_{kl}(t) \label{eqn:IRMMV_actualandreferenceparametersrelationV} .\\
\intertext{Combining the results from \eqref{eqn:IRMMV_GradientFlowActualLossreferenceRelationV} and \eqref{eqn:IRMMV_actualandreferenceparametersrelationV} we have:} 
    \frac{\mathrm{d}}{\mathrm{d}t} \tilde{V}_{kl}^{}(t) &= - \nabla_{\tilde{V}_{il}^{}(t)}\mathcal{L}_N^{}(\tilde{\mathbf{g}}, \tilde{\mathbf{V}}). \label{eqn:IRMMV_GradientFlowNV1}
\end{align}

For the components outside the support set ($i \notin \mathcal{S}_K$):\\

By definitions \eqref{eqn:IRMMV_tildegbargRelation} and \eqref{eqn:IRMMV_tildegbarVRelation}, $\tilde{g}_i^{}(t) = 0 $ and $\tilde{V}_{il}^{}(t) = 0 $ for all $t\geq 0$. 
Also, note that $\nabla_{\tilde{V}_{il}^{}(t)}\mathcal{L}_N^{}(\tilde{\mathbf{g}}(t), \tilde{\mathbf{V}}(t)) \propto \tilde{g}_i^2(t)$ and $\nabla_{\tilde{g}_i^{}(t)}\mathcal{L}_N^{}(\tilde{\mathbf{g}}(t), \tilde{\mathbf{V}}(t)) \propto \tilde{g}_i^{}(t)$. Combining we arrive at:

\begin{align}
    \frac{\mathrm{d}}{\mathrm{d}t} \tilde{g}_i^{}(t) = 0 = -\nabla_{\tilde{g}_i^{}(t)}\mathcal{L}_N^{}(\tilde{\mathbf{g}}(t), \tilde{\mathbf{V}}(t))\label{eqn:IRMMV_GradientFlowNg2},\\
    \intertext{and}
    \frac{\mathrm{d}}{\mathrm{d}t} \tilde{V}_{il}^{}(t) = 0 = -\nabla_{\tilde{V}_{il}^{}(t)}\mathcal{L}_N^{}(\tilde{\mathbf{g}}(t), \tilde{\mathbf{V}}(t))\label{eqn:IRMMV_GradientFlowNV2}.
\end{align}
    Thus, from equations \eqref{eqn:IRMMV_GradientFlowNg1}, \eqref{eqn:IRMMV_GradientFlowNV1}, \eqref{eqn:IRMMV_GradientFlowNg2}, and \eqref{eqn:IRMMV_GradientFlowNV2} we have that $\tilde{\mathbf{g}}(t)$ and $\tilde{\mathbf{V}}(t)$ both follow gradient flow path on $N$ component.
\end{proof}

\begin{lemma}[Lemma \ref{lemma:MMV_NormasIndividualComponents} restated] \label{lemma:MMV_NormasIndividualComponentsApp}
Let the estimated trajectory, $\mathbf{X}(t)$ be parametrized by $({\mathbf{g}}(t), {\mathbf{V}}(t))$ and a reference trajectory $\tilde{\mathbf{X}}(t)$, be parameterized by $(\tilde{\mathbf{g}}(t), \tilde{\mathbf{V}}(t))$ both of which evolve under the continuous gradient flow, as described in Lemma \ref{lemma:MMV_Balancedness}. 
Let $T_{}^{}>0$ and $D>0$ scalars.  Specifically, assume that for all $t\in[0, T_{}]$ and all $n\in[N]$, the condition ${\max}\left\{\vert g_n^{}(t)\vert , \Vert V_{n:}^{}(t)\Vert _2^{}, \vert \tilde{g}_n^{}(t)\vert , \Vert \tilde{V}_{n:}^{}(t),\Vert _2^{}\right\}\leq D$ holds.

 At any time $t \in [0,T_{}^{}]$, the following inequality holds:
\begin{align} 
    \left\Vert \mathbf{X}(t) - \tilde{\mathbf{X}}(t)\right\Vert ^{2}_F &\leq  8D^4_{}\Big\Vert (\mathbf{g}(t), \mathbf{V}(t)) \left.- (\tilde{\mathbf{g}}(t), \tilde{\mathbf{V}}(t))\right\Vert ^2_{},
\end{align}
where, $\Big\Vert (\mathbf{g}(t), \mathbf{V}(t)) \left.- (\tilde{\mathbf{g}}(t), \tilde{\mathbf{V}}(t))\right\Vert ^2_{} = \sum\limits_{n\in[N]}\sum\limits_{l\in[L]}\left(\left({V}_{nl}^{}(t) - \tilde{{V}}_{nl}^{}(t)\right)^2 \right.\left.+\left({g}_{n}^{}(t) - \tilde{g}_n^{}(t)\right)^2_{}\right)$.
\end{lemma}
\begin{proof}
    \begin{align}
        \left\Vert\mathbf{X}(t) - \tilde{\mathbf{X}}(t)\right\Vert^{2}_F 
        & = \sum\limits_{n\in[N]}\left\Vert{g}_{n}^{2}(t)\mathbf{V}_{n:}^{}(t) - \tilde{g}_n^{2}(t)\tilde{\mathbf{V}}_{n:}^{}(t)\right\Vert^{2}_2\\
        & = \sum\limits_{n\in[N]}\left\Vert{g}_{n}^{2}(t)\mathbf{V}_{n:}^{}(t) +{g}_{n}^{2}(t)\tilde{\mathbf{V}}_{n:}^{}(t)\right.\left.-{g}_{n}^{2}(t)\tilde{\mathbf{V}}_{n:}^{}(t) - \tilde{g}_n^{2}(t)\tilde{\mathbf{V}}_{n:}^{}(t)\right\Vert^{2}_2\\
        & = \sum\limits_{n\in[N]}\left\Vert{g}_{n}^{2}(t)\left(\mathbf{V}_{n:}^{}(t) - \tilde{\mathbf{V}}_{n:}^{}(t)\right) \right.\left.+\tilde{\mathbf{V}}_{n:}^{}(t)\left({g}_{n}^{2}(t) - \tilde{g}_n^{2}(t)\right)\right\Vert^{2}_2\\
        & \leq \sum\limits_{n\in[N]}\left({g}_{n}^{2}(t)\left\Vert\mathbf{V}_{n:}^{}(t) - \tilde{\mathbf{V}}_{n:}^{}(t)\right\Vert _2^{} \right.\left.+\left\Vert \tilde{\mathbf{V}}_{n:}^{}(t)\right\Vert _2^{}\left\vert \left({g}_{n}^{2}(t) - \tilde{g}_n^{2}(t)\right)\vert\right\vert \right)^{2}_{},\\
        \intertext{using $(a+b)^2_{}\leq 2a^2_{}+2b^2_{}$, }
        \left\Vert \mathbf{X}(t) - \tilde{\mathbf{X}}(t)\right\Vert ^{2}_F 
        & \leq 2\sum\limits_{n\in[N]}\left(\left({g}_{n}^{2}(t)\right)^2\left\Vert \mathbf{V}_{n:}^{}(t) - \tilde{\mathbf{V}}_{n:}^{}(t)\right\Vert _2^2 \right.\left.+\left\Vert \tilde{\mathbf{V}}_{n:}^{}(t)^2_{}\right\Vert ^2_{2}\left({g}_{n}^{2}(t) - \tilde{g}_n^{2}(t)\right)^2_{}\right)\\
        & \leq 2\sum\limits_{n\in[N]}\left({g}_{n}^{4}(t)\sum\limits_{l\in[L]}\left({V}_{nl}^{}(t) - \tilde{{V}}_{nl}^{}(t)\right)^2 \right.\left.+\sum\limits_{l\in[L]}\tilde{{V}}_{nl}^{}(t)^2_{}\left(\left({g}_{n}^{}(t) - \tilde{g}_n^{}(t)\right)\left({g}_{n}^{}(t) + \tilde{g}_n^{}(t)\right)^{}_{}\right)^2_{}\right)\label{eqn:IRMMV_NormOfDifferenceBetweenReferenceAndActualTrajectory}\\
        \intertext{Using the inequalities $\left\{\vert g_n^{}(t)\vert , \Vert V_{n:}^{}(t)\Vert _2^{}\right\}
         \leq D$ and $\left\{\Vert \tilde{g}_n^{}(t)\vert , \Vert \tilde{V}_{n:}^{}(t)\Vert _2^{}\right\}
         \leq D$, we can write the above equation as:
        }
        \left\Vert \mathbf{X}(t)\right.\left. - \tilde{\mathbf{X}}(t)\right\Vert ^{2}_F
         & \leq 2\sum\limits_{n\in[N]}\sum\limits_{l\in[L]}\left(D^4\left({V}_{nl}^{}(t) - \tilde{{V}}_{nl}^{}(t)\right)^2 \right.\left.+4D^4\left({g}_{n}^{}(t) - \tilde{g}_n^{}(t)\right)^2_{}\right)\\
         & \leq 8D^4_{}\sum\limits_{n\in[N]}\sum\limits_{l\in[L]}\left(\left({V}_{nl}^{}(t) - \tilde{{V}}_{nl}^{}(t)\right)^2 \right.\left.+\left({g}_{n}^{}(t) - \tilde{g}_n^{}(t)\right)^2_{}\right)\\
         \intertext{Using the squared parameter distance defined in equation \eqref{eqn:IRMMV_square_dist_appendix}:}
         & \leq 8D^4_{}\Big\Vert (\mathbf{g}(t), \mathbf{V}(t)) \left.- (\tilde{\mathbf{g}}(t), \tilde{\mathbf{V}}(t))\right\Vert ^2_{}.
    \end{align}
\end{proof}

\subsection{Proof of Theorem \ref{thm:IRMMV_mainResult}}
\label{sec:IRMMV_proofofThmV1}

To prove the theorem, our strategy is to first construct an idealized, reference low-rank trajectory, denoted as $\tilde{\mathbf{X}}(t)$, which will serve as a theoretical benchmark. This trajectory is designed to be perfectly rank-$K$ by initializing it such that only the rows within a specific support set, $\mathcal{S}_K^{}$, are non-zero. The initial magnitude of these non-zero rows is controlled by a small positive scalar, $\rho$. 

The core of our proof will now be to show that the estimated trajectory found by the algorithm, namely, $\mathbf{X}(t)$, remains close to this reference trajectory, namely, $\tilde{\mathbf{X}}(t)$, provided the initialization is sufficiently small. It is crucial to note that estimated trajectory, $\mathbf{X}(t)$, does not assume any knowledge of the support set $\mathcal{S}_K^{}$ or sparsity level $K$.  

A standard approach to bounding the divergence between the two trajectories is to employ Grönwall's inequality \cite{razin2021implicit}. However, this technique yields an error bound that grows exponentially with time. This forces the initializations to be set far below the machine precision, rendering the theoretical guarantees practically unrealizable.

To address this, we construct a Lyapunov function, $\mathcal{E}(t)$. This approach allows us to track the divergence more tightly, leading to a bound that linearly $t$ rather than exponentially. Consequently, we derive convergence guarantees that hold for initializations that are numerically implementable.

We begin the proof using the idealized rank-$K$ trajectory, $\tilde{\mathbf{X}}(t)$ as in \eqref{eqn:IRMMV_NormXreferenceInit}. This initialization along with Corollary \ref{corr:IRMMV_EvolutionOfComponents} and Lemma \ref{lemma:IRMMV_ParametersStays0ifInitializedat0} ensures that 
the rows outside the support $\mathcal{S}_K$ are initialized at zero and remain zero under the gradient flow dynamics. Thus, $\tilde{\mathbf{X}}(t)$ is guaranteed to be an at-most rank-$K$ matrix for all $t\geq0$.

To quantify the proximity between the estimated parameters $(\mathbf{g}(t), \mathbf{V}(t))$ and reference parameters $(\tilde{\mathbf{g}}(t), \tilde{\mathbf{V}}(t))$,  we use the squared distance,  $\Big\Vert (\mathbf{g}(t), \mathbf{V}(t))\left. - (\tilde{\mathbf{g}}(t), \tilde{\mathbf{V}}(t))\right\Vert ^2$.

We formally define the Lyapunov function $\mathcal{E}(t)$ as: 
\begin{align}
\mathcal{E}(t) &= \Big\Vert (\mathbf{g}(t), \mathbf{V}(t))\left. - (\tilde{\mathbf{g}}(t), \tilde{\mathbf{V}}(t))\right\Vert ^2 = \sum\limits_{i \in [N]}\sum\limits_{l\in [L]}\Big(\big\vert g_i^{}(t) - \tilde{g}_i^{}(t)\big\vert ^2 + \big\vert V_{il}^{}(t) - \tilde{V}_{il}^{}(t)\big\vert ^2\Big)\\
\intertext{We decompose this into summation of independent row-wise components, $\mathcal{E}_{i^{}_{}}^{}(t)$, corresponding to each row index $i \in [N]$:}
\mathcal{E}(t) &= \sum\limits_{i \in [N]}\mathcal{E}_{i^{}}^{}(t)\\
 \text{where }\mathcal{E}_{i^{}}^{}(t)&=\sum\limits_{l\in [L]}\Big(\big\vert g_i^{}(t) - \tilde{g}_i^{}(t)\big\vert ^2 + \big\vert V_{il}^{}(t) - \tilde{V}_{il}^{}(t)\big\vert ^2\Big),\quad \forall i\in[N]
\end{align}
\subsubsection{Bounding the error on non-support indices $(i\notin\mathcal{S}_K^{})$}
\label{subsection:BoundingErroronNonSupport}
We begin by analyzing the error growth on the non-support indices/rows $\big(i\notin\mathcal{S}_K^{}\big)$. For these rows, the reference trajectory is zero (by construction $\tilde{g}_i^{}(0) = 0, \tilde{V}_{il}^{}(0) = 0$ and its subsequent updates $\tilde{g}_i^{}(t) = 0, \tilde{V}_{il}^{}(t) = 0$ as a consequence of Lemma \ref{lemma:MMV_rowGradientflowEvenWhenExtendedtoZeros}). Consequently, the row-wise error term simplifies to the energy of estimated trajectory alone. 
\begin{align}
    \mathcal{E}_{i^{}}^{}(t)= \mathcal{E}_{e_i^{}}(t)&=\sum\limits_{l\in [L]}\Big(g_i^{}(t)^2 + V_{il}^{}(t)^2\Big)\\
    \intertext{From perfect row-balancedness property, i.e., $\Big(\frac{1}{2}g_{i}^{2}(t)=\sum_{l\in[L]}V_{il}^{2}(t)\Big)$, error in terms of $\vert g_i^{}(t)\vert$ is expressed as}
     \mathcal{E}_{e_i^{}}(t)&=\Bigg(L + \frac{1}{2}\Bigg)\vert g_i^{}(t)\vert ^2_{} \label{eqn:IRMMV_LyapunovFunctionNotinSupport}\\
     \intertext{At initialization $(t= 0)$, the error is given by:}
     \mathcal{E}_{{e}_i^{}}(0)&=\Bigg(L + \frac{1}{2}\Bigg)\alpha_g^2 = 2L\Bigg(L + \frac{1}{2}\Bigg)\alpha_V^2\label{eqn:IRMMV_LyapunovFunctionNotinSupportInititalization}
    \intertext{Next, to bound the growth of error, we differentiate \wrt time on both sides:}
   \ddt \mathcal{E}_{e_i^{}}(t) & = \sum\limits_{l\in [L]}\bigg(\ddt g_i^{}(t)^2 + \ddt V_{il}^{}(t)^2\bigg)\\
   \ddt \mathcal{E}_{e_i^{}}(t) & = \bigg(L\ddt g_i^{}(t)^2 + \ddt \sum\limits_{l\in [L]}V_{il}^{}(t)^2\bigg)\\
    \intertext{From equation \eqref{eqn:MMV_unbalancednessDer_rowWise}, (i.e. $\Big(\frac{1}{2}\frac{\mathrm{d}}{\mathrm{d}t}g_i^{2}(t) = \frac{\mathrm{d}}{\mathrm{d}t} \Vert \mathbf{V}_{i:}(t)\Vert _2^2 = \ddt \sum\limits_{l\in[L]}^{} V_{il}^{2}\Big)$}
    \ddt \mathcal{E}_{e_i^{}}(t) & = \Bigg(L + \frac{1}{2}\Bigg)\bigg(\ddt g_i^{}(t)^2\bigg)\\
    \ddt \mathcal{E}_{e_i^{}}(t) & = 2\Bigg(L + \frac{1}{2}\Bigg)\bigg(g_i^{}(t)\ddt g_i^{}(t)\bigg)\\
    \intertext{Substituting the gradient flow dynamics for $g_i^{}(t)$ from equation \eqref{eqn:MMV_gradgl} $\Big(\frac{\mathrm{d}}{\mathrm{d}t} {g}_i^{}(t)= 4g_i^{}(t)\sum\limits_{j\in[L]}^{} \Lambda_{ij}(t)V_{lj} (t)\Big)$:}
    & = \bigg(L + \frac{1}{2}\bigg)8g_i^{2}(t)\sum\limits_{l\in[L]}^{} \Lambda_{il}(t)V_{il} (t)\\
    \intertext{Applying Cauchy-Schwarz inequality yields the following upper-bound:}
    & \leq \Bigg(L + \frac{1}{2}\Bigg)8g_i^{2}(t)\big\Vert\mathbf{\Lambda}_{i:}^{}(t)\big\Vert_2^{}\Vert\mathbf{V}_{i:}(t)\big\Vert_2^{}\label{eqn:IRMMV_LyapunovFunctionUpperBoundinTermsofLambdaVTerm1}
\end{align}
To establish an  upper bound on the residual term $ \left\Vert\mathbf{\Lambda}_{i:}^{}(t)\right\Vert_2^{}$, we leverage the monotonic decrease of the loss function \citep{ji2019gradient}. 
\begin{align}
    \Vert\mathbf{\Lambda}_{i:}^{}(t)\Vert_2 & = \Vert\mathbf{A}^\top_{i:}\left(\mathbf{Y} - \mathbf{A}\mathbf{X}(t)\right)\Vert_2^{} \overset{(a)}{\leq} \Vert\mathbf{A}^\top_{i:}\Vert_2\Vert\left(\mathbf{Y} - \mathbf{A}\mathbf{X}(t)\right)\Vert_F  \overset{(b)}{\leq}\Vert\left(\mathbf{Y} - \mathbf{A}\mathbf{X}(0)\right)\Vert_F \\
    &{\leq}\Vert\mathbf{Y}\Vert_F^{} + \Vert\mathbf{A}\Vert_F^{}\Vert\mathbf{X}(0)\Vert_F \overset{(c)}{\leq} \Vert\mathbf{Y}\Vert_F +  N \sqrt{L}\alpha_g^2\alpha_V = \Vert\mathbf{Y}\Vert_F + 2 L^{3/2}N \alpha_V^3 := C_\Lambda
\end{align}
Where $\overset{(a)}{}$ follows from sub-multiplicativity \citep{belitskii2013matrix} of the Frobenius norm (which is a direct consequence of the Cauchy-Schwarz inequality applied column-wise),  $\overset{(b)}{}$ from the $\ell_2^{}$ normalization of columns of $\mathbf{A}$ and the monotonic decrease of loss and $\overset{(c)}{}$ from equation \eqref{eqn:IRMMV_X_0rowi}
Recall that the gradient flow preserves row-wise balancedness, $\Big(\frac{1}{2}g_i^{}(t)^2_{} = \sum\limits_{j\in[L]}V_{ij}(t)^2\Big)$. Substituting the above derived bound for $ \Vert\mathbf{\Lambda}_{i:}^{}(t)\Vert_2$ and balancedness into equation \eqref{eqn:IRMMV_LyapunovFunctionUpperBoundinTermsofLambdaVTerm1}, we obtain
\begin{align}
    \ddt \mathcal{E}_{e_i^{}}(t) & \leq 4\sqrt{2}\Bigg(L + \frac{1}{2}\Bigg) C_\Lambda\vert g_i^{}(t)\vert^3\\
    \intertext{From equation \eqref{eqn:IRMMV_LyapunovFunctionNotinSupportInititalization}, we can substitute for $\vert g_i^{}(t)\vert = \big(L+\frac{1}{2}\big)^{-1/2}\mathcal{E}_{e_i^{}}(t)^{1/2}$. Then the bound becomes}
    \ddt \mathcal{E}_{e_i^{}}(t) & \leq 4\sqrt{2}\Bigg(L + \frac{1}{2}\Bigg)^{-\frac{1}{2}} C_\Lambda \mathcal{E}_{e_i^{}}^{3/2}(t) = C_\Lambda^\prime\mathcal{E}_{e_i^{}}^{3/2}(t)\label{eqn:IRMMV_LyapunovInequalitOfDerivativeNotinSupport}
\end{align}
where, $ C_\Lambda^\prime := 4\sqrt{2}\Bigg(L + \frac{1}{2}\Bigg)^{-\frac{1}{2}} C_\Lambda $. 
Rearranging the terms to separate the variables $\mathcal{E}_{e_i}$ and $t$ in equation \eqref{eqn:IRMMV_LyapunovInequalitOfDerivativeNotinSupport}, we obtain:
\begin{align}
    \mathcal{E}_{e_i}^{-3/2}(t) \frac{\mathrm{d}\mathcal{E}_{e_i}(t)}{\mathrm{d}t} & \leq C_\Lambda^\prime\\
    \int_{0}^{t} \mathcal{E}_{e_i}^{-3/2}(\tau) \frac{\mathrm{d}\mathcal{E}_{e_i}(\tau)}{\mathrm{d}\tau} \mathrm{d}\tau & \leq \int_{0}^{t} C_\Lambda^\prime \mathrm{d}\tau \nonumber \\
    \int_{\mathcal{E}_{e_i}(0)}^{\mathcal{E}_{e_i}(t)} u^{-3/2} \mathrm{d}u & \leq C_\Lambda^\prime t \label{eqn:integral_step} \\
    \left[ -2 u^{-1/2} \right]_{\mathcal{E}_{e_i}(0)}^{\mathcal{E}_{e_i}(t)} & \leq C_\Lambda^\prime t \nonumber \\
    -2 \mathcal{E}_{e_i^{}}^{-1/2}(t) + 2  \mathcal{E}_{e_i^{}}^{-1/2}(0) &\leq C_\Lambda^\prime t\\
    \intertext{Isolating the term involving $\mathcal{E}_{e_i^{}}(t)$:}
    \mathcal{E}_{e_i^{}}^{-1/2}(t) &\geq \mathcal{E}_{e_i}^{-1/2}(0)- \frac{1}{2}C_\Lambda^\prime t\\
    \intertext{Taking the reciprocal and squaring both sides yields the upper bound:}
    \Aboxed{\mathcal{E}_{e_i^{}}(t) &\leq \frac{1}{\Big(\mathcal{E}_{e_i^{}}^{-1/2}(0)- \frac{1}{2} C_\Lambda^\prime t\Big)^2}}\label{eqn:IRMMV_LyapunovTermIndividualRowNonSupport}
\end{align}

\subsubsection{Bounding the error on support indices $(i\in\mathcal{S}_K^{})$}

We now extend our analysis to the rows in the support set $\mathcal{S}_K^{}$. Unlike the non-support indices where the reference trajectory remains zero, here the reference trajectory $\tilde{\mathbf{X}}(t)$ evolves non-trivially according to the gradient flow dynamics. To bound the divergence for these indices, we decompose the total error $\mathcal{E}_{i}(t)$ into three distinct components: the energy of the estimated trajectory $\mathcal{E}_{e_i^{}}^{}(t)$, the energy of the reference trajectory $\mathcal{E}_{r_i^{}}^{}(t)$, and a cross-term $\mathcal{E}_{x_i^{}}^{}(t)$. For each row,  $i\in\mathcal{S}_k^{}$:
\begin{align}
    \mathcal{E}_{i^{}}(t)&=\sum\limits_{l\in [L]}\Big(\big(g_i^{}(t) - \tilde{g}_i^{}(t)\big)^2 + \big(V_{il}^{}(t) - \tilde{V}_{il}^{}(t)\big)^2\Big)\label{eqn:IRMMV_LyapunovFunctionSupportV1}\\ 
    \intertext{Expanding the squared terms and regrouping the components, we obtain:}
    &=\sum\limits_{l\in [L]}\Big(g_i^{2}(t) + \tilde{g}_i^{2}(t) - 2g_i^{}(t)\tilde{g}_i^{}(t) + V_{il}^{2}(t) + \tilde{V}_{il}^{2}(t) - 2V_{il}^{}(t)\tilde{V}_{il}^{}(t)\Big)\\
    &= \mathcal{E}_{e_i^{}}^{}(t)+\mathcal{E}_{r_i^{}}^{}(t) -2\mathcal{E}_{x_i^{}}^{}(t)\label{eqn:IRMMV_LyapunovFunctionSupportV2}
    \intertext{where the reference energy $\mathcal{E}_{r_i}(t)$ and the cross-term energy $\mathcal{E}_{x_i}(t)$ are defined as:}
    \mathcal{E}_{r_i^{}}^{}(t) &= \sum\limits_{l\in [L]} \Big(\tilde{g}_i^{}(t)^2 + \tilde{V}_{il}^{}(t)^2\Big), \\
    \mathcal{E}_{x_i^{}}^{}(t) &=\sum\limits_{l\in [L]} \Big(g_i^{}(t)\tilde{g}_i^{}(t) + V_{il}^{}(t)\tilde{V}_{il}^{}(t)\Big).
    \end{align}
    First we bound the energy of the reference trajectory, $\mathcal{E}_{r_i^{}}^{}(t)$. 
    
    Similar to the approach for the estimated trajectory, we utilize the  perfect row-balancedness to express the energy in terms of the scaling factor $\tilde{g}_i(t) $: 
    \begin{align}
    \mathcal{E}_{r_i^{}}^{}(t)&=\sum\limits_{l\in [L]}\Big(\tilde{g}_i^{}(t)^2 + \tilde{V}_{il}^{}(t)^2\Big)\\
     \mathcal{E}_{r_i^{}}^{}(t)&=\Bigg(L + \frac{1}{2}\Bigg)\tilde{g}_i^{}(t)^2 \label{eqn:IRMMV_LyapunovFunctionSupportReferenceTrajectory}\\
     \intertext{At time $t= 0 $, the error is}
     \mathcal{E}_{r_i^{}}^{}(0)&=\left(2L + 1\right)\rho^{2/3}\label{eqn:IRMMV_LyapunovFunctionSupportInititalization}
    \intertext{To bound the growth of energy, we differentiate with respect to time. Substituting the gradient flow dynamics (Lemma \ref{lemma:MMV_rowGradientflowEvenWhenExtendedtoZeros}) and applying the Cauchy-Schwarz inequality yields the following upper bound:}
   \ddt \mathcal{E}_{r_i^{}}^{}(t) & = \Bigg(L + \frac{1}{2}\Bigg)\bigg(\ddt \tilde{g}_i^{}(t)^2\bigg)\\
    & \leq \Bigg(L + \frac{1}{2}\Bigg)8\tilde{g}_i^{2}(t)\big\Vert\mathbf{\tilde{\Lambda}}_{i:}^{}(t)\big\Vert_2^{}\Vert\tilde{\mathbf{V}}_{i:}(t)\big\Vert_2^{}\label{eqn:IRMMV_LyapunovFunctionReferenceUpperBoundinTermsofLambdaVTerm1} 
\end{align}
where we have utilized $\Big(\frac{\mathrm{d}}{\mathrm{d}t} \tilde{g}_i^{}(t)= 4\tilde{g}_i^{}(t)\sum\limits_{j\in[L]}^{} \tilde{\Lambda}_{ij}(t)\tilde{V}_{lj} (t)\Big)$.

Now, we establish an upper bound on the residual term $\Vert\tilde{\mathbf{\Lambda}}_{i:}^{}(t)\Vert_2$. Proceeding similarly to the derivation of $C_\Lambda$, we leverage the monotonic decrease of the loss function to bound the residual by its initialization values:
\begin{align}
    \Vert\tilde{\mathbf{\Lambda}}_{i:}^{}(t)\Vert_2 
    &
    \leq \Vert\mathbf{Y}\Vert_F^{}+ \Vert\mathbf{A}\Vert_F^{}\Vert\tilde{\mathbf{X}}(0)\Vert_F^{}\leq \Vert\mathbf{Y}\Vert_F + \sqrt{K}N^{} \rho := \tilde{C}_\Lambda
\end{align}

Substituting the above bound and perfect row-balancedness, we arrive at the differential inequality:
\begin{align}
    \ddt \mathcal{E}_{r_i^{}}^{}(t) & \leq 4\sqrt{2}\Bigg(L + \frac{1}{2}\Bigg) C_\Lambda \vert\tilde{g}_i^{}(t)\vert^{3}\\
    \intertext{Using equation \eqref{eqn:IRMMV_LyapunovFunctionSupportReferenceTrajectory} we can substitute  $\vert \tilde{g}_i^{}(t)\vert = \big(L+\frac{1}{2}\big)^{-1/2}\mathcal{E}_{r_i^{}}(t)^{1/2}$.}
    \ddt \mathcal{E}_{r_i^{}}^{}(t) & \leq 4\sqrt{2}\Bigg(L + \frac{1}{2}\Bigg)^{-\frac{1}{2}} \tilde{C}_\Lambda \mathcal{E}_{r_i^{}}^{3/2}(t) = \tilde{C}_\Lambda^\prime\mathcal{E}_{r_i^{}}^{3/2}(t)\label{eqn:IRMMV_LyapunovInequalitOfDerivativeNotinSupportReferenceTrajectory}
\end{align}
where, $ \tilde{C}_\Lambda^\prime := 4\sqrt{2}\Bigg(L + \frac{1}{2}\Bigg)^{-\frac{1}{2}} \tilde{C}_\Lambda $

The above differential inequality is structurally identical to the one for the estimated trajectory solved in the sub-section \ref{subsection:BoundingErroronNonSupport}. Integrating it over the interval $[0, t]$ immediately yields the analogous upper bound:
\begin{align}
    \Aboxed{\mathcal{E}_{r_i^{}}(t) \leq \frac{1}{\Big(\mathcal{E}_{r_i^{}}^{-1/2}(0)- \frac{1}{2} \tilde{C}_\Lambda^\prime t\Big)^2}}.\label{eqn:IRMMV_LyapunovTermIndividualRowSupport}
\end{align}

\paragraph{Bounding the Cross-Term $\left(\mathcal{E}_{x_i^{}}\right)$:\\}

Having established upper bounds for the energy of both the estimated and reference trajectories, we now bound the cross-term $\mathcal{E}_{x_i}(t)$, which represents the alignment between the estimated and reference trajectories. 

Recall the decomposition of the total error $\mathcal{E}_i(t)$, which we restate here for brevity:
\begin{equation}
    \mathcal{E}_i(t) = \mathcal{E}_{e_i}(t) + \mathcal{E}_{r_i}(t) - 2\mathcal{E}_{x_i}(t).
    \tag{\ref{eqn:IRMMV_LyapunovFunctionSupportV2}}
\end{equation}
From this relation, to establish an upper bound on the total error $\mathcal{E}_i^{}(t)$, we require a lower bound on the cross-term $\mathcal{E}_{x_i^{}}^{}(t)$. 

\begin{align}
    \mathcal{E}_{x_i^{}}(t) & = \sum\limits_{l\in[L]}\left(g_i^{}(t)\tilde{g}_i^{}(t) + V_{il}^{}(t)\tilde{V}_{il}^{}(t)\right)\\
    \intertext{Recall that at $t = 0$, the initializations are $V_{ij}^{}(0) = \alpha_V^{}, g_i^{}(0) = \alpha_g^{} = \sqrt{2L}\alpha_V^{}, \tilde{g}_i^{}(0) = \sqrt{2}\rho^{1/3} \text{ and } \tilde{V}_{ij}^{}(0) =  \frac{\rho^{1/3}}{\sqrt{L}}$. The initial cross-term energy is}    
    \mathcal{E}_{x_i^{}}(0) & = \sqrt{L}(2L+1)\alpha_V^{}\rho^{1/3}_{}\\
    \intertext{Differentiating with respect to time and substituting the gradient flow dynamics as before:}
    \ddt \mathcal{E}_{x_i^{}}(t) & = \sum\limits_{l\in[L]}g_i^{}(t)\left(\ddt \tilde{g}_i^{}(t)\right) + \left(\ddt g_i^{}(t)\right)\tilde{g}_i^{}(t) + V_{il}^{}(t)\left(\ddt \tilde{V}_{il}^{}(t)\right) + \left(\ddt V_{il}^{}(t)\right)\tilde{V}_{il}^{}(t)\\
     & = \sum\limits_{l\in[L]}\left(4g_i^{}(t)\tilde{g}_i^{}(t)\sum\limits_{j\in[L]}^{} \tilde{\Lambda}_{ij}(t)\tilde{V}_{ij} (t) + 4g_i^{}(t)\tilde{g}_i^{}(t)\sum\limits_{j\in[L]}^{} \Lambda_{ij}(t)V_{ij} (t) + 2\tilde{g}_i^{2}(t)V_{il}^{}(t) \tilde{\Lambda}_{il}(t) + 2g_i^{2}(t)\Lambda_{il}(t)\tilde{V}_{il}^{}(t)\right)\\
     \intertext{Summing over $l$ allows us to group terms as}
     & = 4Lg_i^{}(t)\tilde{g}_i^{}(t)\left(\sum\limits_{j\in[L]}^{} \tilde{\Lambda}_{ij}(t)\tilde{V}_{ij} (t) + \sum\limits_{j\in[L]}^{} \Lambda_{ij}(t)V_{ij} (t)\right) + 2\vert \tilde{g}_i(t)\vert ^{2}_{}\sum\limits_{l\in[L]}V_{il}^{}(t) \tilde{\Lambda}_{il}(t) + 2\vert g_i(t)\vert ^{2}_{}\sum\limits_{l\in[L]}\Lambda_{il}(t)\tilde{V}_{il}^{}(t)\\
     \intertext{Rewriting in terms of inner products:}
     & = 4Lg_i^{}(t)\tilde{g}_i^{}(t)\left(\left<\tilde{\mathbf{\Lambda}}_{i:}^{}(t)\tilde{\mathbf{V}}_{i:} ^{}(t)\right> +\Big<\mathbf{\Lambda}_{i:}^{}(t)\mathbf{V}_{i:}^{} (t)\Big>\right) + 2\tilde{g}_i^{2}(t) \left<\mathbf{V}_{i:}^{}(t) \tilde{\Lambda}_{i:}(t)\right> + 2g_i^{2}(t)\left<\Lambda_{i:}^{}(t)\tilde{V}_{i:}^{}(t)\right>\\
     \intertext{Applying the Cauchy-Schwarz-Bunjakowski inequality \cite{teschl2012ordinary} $\left (\left\vert\langle A, B \rangle\right\vert \leq \Vert A\Vert \Vert B\Vert \implies  - \Vert A\Vert \Vert B\Vert\leq  \langle A, B \rangle \right)$ :}
     & \geq -4L\vert g_i^{}(t)\vert \vert \tilde{g}_i^{}(t)\vert \left(\Vert \tilde{\mathbf{\Lambda}}_{i:}^{}(t)\Vert \Vert \tilde{\mathbf{V}}_{i:} ^{}(t)\Vert + \Vert \mathbf{\Lambda}_{i:}^{}(t)\Vert \Vert \mathbf{V}_{i:}^{} (t)\Vert\right) - 2 \tilde{g}_i^{2}(t) \Vert \mathbf{V}_{i:}^{}(t) \Vert \Vert \tilde{\Lambda}_{i:}(t)\Vert - 2g_i^{2}(t) \Vert \Lambda_{i:}^{}(t)\Vert \Vert \tilde{V}_{i:}^{}(t)\Vert \\
     \intertext{Substituting the bounds $\|\mathbf{\Lambda}_{i:}\| \leq C_\Lambda$ and $\|\tilde{\mathbf{\Lambda}}_{i:}\| \leq \tilde{C}_\Lambda$}
     & \geq - 4L\vert g_i^{}(t)\vert \vert \tilde{g}_i^{}(t)\vert \left(\tilde{C}_{\Lambda}^{}(t) \Vert \tilde{\mathbf{V}}_{i:} ^{}(t)\Vert + C_{\Lambda}^{} \Vert \mathbf{V}_{i:}^{}(t)\Vert\right) - 2  \tilde{C}_{\Lambda} \tilde{g}_i^{2}(t) \Vert \mathbf{V}_{i:}^{}(t) \Vert  - 2 C_\Lambda^{}(t) g_i^{2}(t) \Vert \tilde{V}_{i:}^{}(t)\Vert
     \intertext{Applying row-balancedness (i.e., $\sqrt{2}\|\mathbf{V}_{i:}\| = |g_i|$)}
     & \geq -2\sqrt{2}L\vert g_i^{}(t)\vert \vert \tilde{g}_i^{}(t)\vert^2\tilde{C}_{\Lambda}^{} -  2\sqrt{2}L\vert g_i^{}(t)\vert ^2\vert \tilde{g}_i^{}(t)\vert C_{\Lambda}^{} -\sqrt{2}\tilde{C}_{\Lambda} \tilde{g}_i^{2}(t) \vert g_{i}^{}(t) \vert  - \sqrt{2} C_\Lambda^{}(t) g_i^{2}(t) \vert \tilde{g}_{i}^{}(t)\vert\\
     \intertext{Rearranging the terms and factoring out the common scalar components allows us to group the expression as}
     & \geq -\sqrt{2}\vert g_i^{}(t)\vert \vert \tilde{g}_i^{}(t)\vert^2\tilde{C}_{\Lambda}^{}(t)(2L+1) -  \sqrt{2}\vert g_i^{}(t)\vert^2 \vert \tilde{g}_i^{}(t)\vert C_{\Lambda}^{}(2L+1)\\    
    \intertext{Substituting $ \bigg(\vert g_i^{}(t)\vert = \mathcal{E}_{e_i^{}}(t)^{1/2}_{}\Big(L + \frac{1}{2}\Big)^{-1/2}\bigg)$ from equation \eqref{eqn:IRMMV_LyapunovFunctionNotinSupport}  and $\bigg(\vert \tilde{g}_i^{}(t)\vert =  \mathcal{E}_{r_i^{}}^{}(t)^{1/2}\Big(L + \frac{1}{2}\Big)^{-1/2} \bigg)$ equation \eqref{eqn:IRMMV_LyapunovFunctionSupportReferenceTrajectory}  }
    & \geq -\frac{{4}}{\sqrt{2L+1}}\vert \mathcal{E}_{e_i^{}}(t)^{1/2}_{}\vert \vert \mathcal{E}_{r_i^{}}^{}(t)\vert\tilde{C}_{\Lambda}^{}(t) -  \frac{{4}}{\sqrt{2L+1}}\vert \mathcal{E}_{e_i^{}}(t)\vert\vert \mathcal{E}_{r_i^{}}^{}(t)^{1/2}\vert C_{\Lambda}^{}\\  
    \intertext{Furthermore, by substituting the time-dependent upper bounds derived for $\mathcal{E}_{e_i^{}}^{}(t)$ and $\mathcal{E}_{r_i^{}}^{}(t)$ (specifically from equations \eqref{eqn:IRMMV_LyapunovTermIndividualRowSupport} and \eqref{eqn:IRMMV_LyapunovTermIndividualRowNonSupport}), we obtain the lower bound in terms of $t$:}
    & \geq -\frac{4}{\sqrt{2L+1}}\BiggFo(\frac{1}{\Big|\mathcal{E}_{e_i^{}}^{-1/2}(0)- \frac{1}{2} C_\Lambda^\prime t\Big|}\BiggFo) \BiggFo( \frac{1}{\Big(\mathcal{E}_{r_i^{}}^{-1/2}(0)- \frac{1}{2} \tilde{C}_\Lambda^\prime t\Big)^2}\BiggFo)\tilde{C}_{\Lambda}^{} \nonumber\\
    &\hspace{2em}-  \frac{4}{\sqrt{2L+1}}\BiggFo(\frac{1}{\Big(\mathcal{E}_{e_i^{}}^{-1/2}(0)- \frac{1}{2} C_\Lambda^\prime t\Big)^2}\BiggFo) \BiggFo( \frac{1}{\Big|\mathcal{E}_{r_i^{}}^{-1/2}(0)- \frac{1}{2} \tilde{C}_\Lambda^\prime t\Big|} \BiggFo) C_{\Lambda}^{} \label{eqn:IRMMV_LyapunovFunctionDerivativeCrossTermV2}\\
\end{align}
We observe that the lower bound depends on time-varying denominator terms $\mathcal{E}_{r_i^{}}^{-1/2}(0)- \frac{1}{2} \tilde{C}_\Lambda^\prime t$ and $\mathcal{E}_{e_i^{}}^{-1/2}(0)- \frac{1}{2} C_\Lambda^\prime t$. As system evolves from $t = 0$, the bound remain valid as long as denominator avoids singularity. This results in an upper bound on time, $t$, for which the analysis holds:





\begin{align}
    \mathcal{E}_{r_i^{}}^{-1/2}(0)- \frac{1}{2} \tilde{C}_\Lambda^\prime t & > 0 \implies \quad
    t< \frac{2}{\tilde{C}_\Lambda^\prime \sqrt{\mathcal{E}_{r_i^{}}(0)}} \\
    \mathcal{E}_{e_i^{}}^{-1/2}(0)- \frac{1}{2} C_\Lambda^\prime t &> 0 \implies \quad
     t< \frac{2}{C_\Lambda^\prime\sqrt{ \mathcal{E}_{e_i^{}}(0)}}
\end{align}

Thus the analysis holds for $t\in[0,T_{}^{})$ where,  $T_{}^{} = \min\left\{ \frac{2}{C_\Lambda^\prime\sqrt{ \mathcal{E}_{e_i^{}}(0)}}, \frac{2}{\tilde{C}_\Lambda^\prime \sqrt{\mathcal{E}_{r_i^{}}(0)}} \right\}$.

Provided $t\in[0, T_{}^{})$, we can rewrite the lower bound \eqref{eqn:IRMMV_LyapunovFunctionDerivativeCrossTermV2} as
\begin{align}
    \ddt \mathcal{E}_{x_i^{}}(t)  & \geq -\frac{4}{\sqrt{2L+1}}\BiggFo(\frac{1}{\mathcal{E}_{e_i^{}}^{-1/2}(0)- \frac{1}{2} C_\Lambda^\prime t}\BiggFo) \BiggFo( \frac{1}{\Big(\mathcal{E}_{r_i^{}}^{-1/2}(0)- \frac{1}{2} \tilde{C}_\Lambda^\prime t\Big)^2}\BiggFo)\tilde{C}_{\Lambda}^{} \nonumber\\
    &\hspace{2em}-  \frac{4}{\sqrt{2L+1}}\BiggFo(\frac{1}{\Big(\mathcal{E}_{e_i^{}}^{-1/2}(0)- \frac{1}{2} C_\Lambda^\prime t\Big)^2}\BiggFo) \BiggFo( \frac{1}{ \mathcal{E}_{r_i^{}}^{-1/2}(0)- \frac{1}{2} \tilde{C}_\Lambda^\prime t } \BiggFo) C_{\Lambda}^{} \label{eqn:IRMMV_LyapunovFunctionDerivativeCrossTermV3}
\end{align}
We now simplify the lower bound derived in \eqref{eqn:IRMMV_LyapunovFunctionDerivativeCrossTermV3}. A closer inspection of the terms reveals a specific algebraic structure: the expression resembles the derivative of an inverse product, i.e., $\ddt \frac{-1}{h(t)^{}_{} f^{}(t)}$. To exploit this, let us define two auxiliary functions $h(t) = \mathcal{E}_{e_i^{}}^{-1/2}(0)- \frac{1}{2} C_\Lambda^\prime t$ and $f(t) =\mathcal{E}_{r_i^{}}^{-1/2}(0)- \frac{1}{2} \tilde{C}_\Lambda^\prime t $. 

Then $\ddt f(t) = -\frac{1}{2}\tilde{C}_\Lambda^\prime$ and $\ddt h(t) = -\frac{1}{2} C_\Lambda^\prime$. Recall that $\tilde{C}_\Lambda^\prime := 4\sqrt{2}\Bigg(L + \frac{1}{2}\Bigg)^{-\frac{1}{2}} \tilde{C}_\Lambda$ and $ C_\Lambda^\prime := 4\sqrt{2}\Bigg(L + \frac{1}{2}\Bigg)^{-\frac{1}{2}} C_\Lambda$. Substituting these equations back into equation \eqref{eqn:IRMMV_LyapunovFunctionDerivativeCrossTermV3} and simplifying we get:
\begin{align}
    \ddt \mathcal{E}_{x_i^{}}(t)  & \geq \frac{h(t)\ddt f(t)+ f(t)\ddt h(t)}{h^2_{}(t) f^2(t)}\\
    \intertext{Rewriting the term as the derivative of an inverse product:}
    \ddt \mathcal{E}_{x_i^{}}(t)  & \geq \ddt \frac{-1}{h(t)^{}_{} f^{}(t)}\\
    \intertext{Integrating both sides with respect to time from $0$ to $t$}
    \int\limits_{ 0}^{t} \frac{\mathrm{d}}{\mathrm{d}\tau} \mathcal{E}_{x_i^{}}(\tau)\mathrm{d}\tau  & \geq \int\limits_{0}^{t} \frac{\mathrm{d}}{\mathrm{d}\tau} \frac{-1}{h(\tau)^{}_{} f^{}(\tau)}\mathrm{d}\tau\\
    \mathcal{E}_{x_i^{}}(t) - \mathcal{E}_{x_i^{}}(0)  & \geq  \frac{1}{h(0)^{}_{} f^{}(0)} - \frac{1}{h(t)^{}_{} f^{}(t)}
    \intertext{Recall that at $\mathcal{E}_{x_i^{}}(0) = \sqrt{L}(2L+1)\alpha_V^{}\rho^{1/3}_{}$, $h(0)= \mathcal{E}_{e_i^{}}^{-1/2}(0) = \left(\alpha_V^{} \sqrt{L(2L+1)}\right)^{-1}$ and $f(0) = \mathcal{E}_{r_i^{}}^{-1/2}(0) =\left(\sqrt{2L+1} \rho^{1/3}\right)^{-1}_{}$.}
    \Aboxed{\mathcal{E}_{x_i^{}}(t) &\geq 2\sqrt{L}(2L+1)\alpha_V^{}\rho^{1/3}_{} - \frac{1}{\left(\mathcal{E}_{e_i^{}}^{-1/2}(0)- \frac{1}{2} C_\Lambda^\prime t\right) \left(\mathcal{E}_{r_i^{}}^{-1/2}(0)- \frac{1}{2} \tilde{C}_\Lambda^\prime t\right)}}\label{eqn:IRMMV_LyapunovTermIndividualRowCrossTerms}
    \end{align}
We now combine the bounds \eqref{eqn:IRMMV_LyapunovTermIndividualRowNonSupport}, \eqref{eqn:IRMMV_LyapunovTermIndividualRowSupport} and \eqref{eqn:IRMMV_LyapunovTermIndividualRowCrossTerms} to obtain an upper bound on the error $\mathcal{E}_{i}^{}(t)$ for all $i \in \mathcal{S}_K^{}$. Recall from equation \eqref{eqn:IRMMV_LyapunovFunctionSupportV1} that the total error decomposes as
\begin{align}
    \mathcal{E}_{i^{}}(t)&= \mathcal{E}_{e_i^{}}^{}(t)+\mathcal{E}_{r_i^{}}^{}(t) -2\mathcal{E}_{x_i^{}}^{}(t) \tag{\ref{eqn:IRMMV_LyapunovFunctionSupportV1}}\\
    \intertext{Substituting the bounds derived from equations \eqref{eqn:IRMMV_LyapunovTermIndividualRowNonSupport}, \eqref{eqn:IRMMV_LyapunovTermIndividualRowSupport} and \eqref{eqn:IRMMV_LyapunovTermIndividualRowCrossTerms}:}
    & \leq \frac{1}{\Big(\mathcal{E}_{e_i^{}}^{-1/2}(0)- \frac{1}{2} C_\Lambda^\prime t\Big)^2} + \frac{1}{\Big(\mathcal{E}_{r_i^{}}^{-1/2}(0)- \frac{1}{2} \tilde{C}_\Lambda^\prime t\Big)^2} +  \frac{2}{\left(\mathcal{E}_{e_i^{}}^{-1/2}(0)- \frac{1}{2} C_\Lambda^\prime t\right) \left(\mathcal{E}_{r_i^{}}^{-1/2}(0)- \frac{1}{2} \tilde{C}_\Lambda^\prime t\right)}  - 4\sqrt{L}(2L+1)\alpha_V^{}\rho^{1/3}_{} \\
    \intertext{Observe that the first three terms form perfect square: }
    &\leq  \left(\frac{1}{\mathcal{E}_{e_i^{}}^{-1/2}(0)- \frac{1}{2} C_\Lambda^\prime t} + \frac{1}{\mathcal{E}_{r_i^{}}^{-1/2}(0)- \frac{1}{2} \tilde{C}_\Lambda^\prime t} \right)^2_{}  - 4\sqrt{L}(2L+1)\alpha_V^{}\rho^{1/3}_{}
    \end{align}
   Since the initialization parameters $\alpha_V^{}$ and $\rho$ are positive, we can omit this negative term to obtain a simplified upper bound:
   \begin{align}
 \mathcal{E}_{i^{}}(t)=\sum\limits_{l\in [L]}\Big(\big(g_i^{}(t) - \tilde{g}_i^{}(t)\big)^2 + \big(V_{il}^{}(t) - \tilde{V}_{il}^{}(t)\big)^2\Big)    &\leq  \left(\frac{1}{\mathcal{E}_{e_i^{}}^{-1/2}(0)- \frac{1}{2} C_\Lambda^\prime t} + \frac{1}{\mathcal{E}_{r_i^{}}^{-1/2}(0)- \frac{1}{2} \tilde{C}_\Lambda^\prime t} \right)^2_{}
\end{align}
\subsubsection{Bounding the total error in the parameter-space}
Summing over all rows (both support and non-support) allows us to bound the total parameter distance squared, $\Vert (\mathbf{g}(t), \mathbf{V}(t)) - (\tilde{\mathbf{g}}(t), \tilde{\mathbf{V}}(t))\Vert ^2$:
\begin{align}
     \Big\Vert (\mathbf{g}(t), \mathbf{V}(t)) - (\tilde{\mathbf{g}}(t), &\tilde{\mathbf{V}}(t))\Big\Vert ^2\nonumber\\
     & =  \sum\limits_{i \in [N]}\sum\limits_{l\in [L]}\Big(\big\vert g_i^{}(t) - \tilde{g}_i^{}(t)\big\vert ^2 + \big\vert V_{il}^{}(t) - \tilde{V}_{il}^{}(t)\big\vert ^2\Big)\nonumber\\
     & =  \sum\limits_{i \in [N]-\mathcal{S}_K^{}}\sum\limits_{l\in [L]}\Big(\big\vert g_i^{}(t) - \tilde{g}_i^{}(t)\big\vert ^2 + \big\vert V_{il}^{}(t) - \tilde{V}_{il}^{}(t)\big\vert ^2\Big) + \sum\limits_{i \in \mathcal{S}_K^{}}\sum\limits_{l\in [L]}\Big(\big\vert g_i^{}(t) - \tilde{g}_i^{}(t)\big\vert ^2 + \big\vert V_{il}^{}(t) - \tilde{V}_{il}^{}(t)\big\vert ^2\Big)\nonumber\\
     &= (N-K)\mathcal{E}_{e_i^{}}^{}(t) + K\mathcal{E}_i^{}(t)\\
     &\leq  \frac{N-K}{\Big(\mathcal{E}_{e_i^{}}^{-1/2}(0)- \frac{1}{2} C_\Lambda^\prime t\Big)^2}+ K  \left(\frac{1}{\mathcal{E}_{e_i^{}}^{-1/2}(0)- \frac{1}{2} C_\Lambda^\prime t} + \frac{1}{\mathcal{E}_{r_i^{}}^{-1/2}(0)- \frac{1}{2} \tilde{C}_\Lambda^\prime t} \right)^2_{}\\
     \intertext{Note that $\frac{1}{\left(\mathcal{E}_{e_i}^{-1/2}(0)- \frac{1}{2} C_\Lambda^\prime t\right)^2} < \left(\frac{1}{\mathcal{E}_{e_i}^{-1/2}(0)- \frac{1}{2} C_\Lambda^\prime t} + \frac{1}{\mathcal{E}_{r_i}^{-1/2}(0)- \frac{1}{2} \tilde{C}_\Lambda^\prime t} \right)^2$ since all terms are strictly positive. Using this inequality, we can upper bound the error contribution from the non-support rows ($N-K$ terms) by the strictly larger squared sum expression used for the support rows. Furthermore, since $N-K < N$, we can combine the coefficients to simplify the total bound:}
     &\leq  N \left(\frac{1}{\mathcal{E}_{e_i^{}}^{-1/2}(0)- \frac{1}{2} C_\Lambda^\prime t} + \frac{1}{\mathcal{E}_{r_i^{}}^{-1/2}(0)- \frac{1}{2} \tilde{C}_\Lambda^\prime t} \right)^2_{}
     \end{align}
     
\subsection{From Parameter Error to Signal Reconstruction Error}
Finally, to guarantee the recovery of the signal matrix $\mathbf{X}$, we translate the error in parameter space into the Frobenius norm of the difference of the estimated trajectory and reference trajectory $\left(\Big\Vert \mathbf{X}(t)\left. - \tilde{\mathbf{X}}(t)\right\Vert_F^2 \right)$. Utilizing the result from Lemma \ref{lemma:MMV_NormasIndividualComponentsApp}, which relates parameter distance to matrix distance for bounded trajectories, we obtain:
\begin{align}
     \Big\Vert \mathbf{X}(t)\left. - \tilde{\mathbf{X}}(t)\right\Vert_F^2 &\leq 8D^4\Big\Vert (\mathbf{g}(t), \mathbf{V}(t)) \left.- (\tilde{\mathbf{g}}(t), \tilde{\mathbf{V}}(t))\right\Vert ^2\\ 
     &\leq 8D^4  N  \left(\frac{1}{\mathcal{E}_{e_i^{}}^{-1/2}(0)- \frac{1}{2} C_\Lambda^\prime t} + \frac{1}{\mathcal{E}_{r_i^{}}^{-1/2}(0)- \frac{1}{2} \tilde{C}_\Lambda^\prime t} \right)^2_{} \label{eqn:IRMMV_DistanceReferenceTrajectoryEstimatedTrajectoryV1}\\
     \Big\Vert \mathbf{X}(t)\left. - \tilde{\mathbf{X}}(t)\right\Vert^{}_F&\leq 2\sqrt{2N}D^2    \left(\frac{1}{\mathcal{E}_{e_i^{}}^{-1/2}(0)- \frac{1}{2} C_\Lambda^\prime t} + \frac{1}{\mathcal{E}_{r_i^{}}^{-1/2}(0)- \frac{1}{2} \tilde{C}_\Lambda^\prime t} \right)\label{eqn:IRMMV_DistanceReferenceTrajectoryEstimatedTrajectoryV2}
\end{align}
Note that the right hand side of the equation \eqref{eqn:IRMMV_DistanceReferenceTrajectoryEstimatedTrajectoryV2} is monotonically increasing with time $t$. To ensure that the error $ \Big\Vert \mathbf{X}(t)\left. - \tilde{\mathbf{X}}(t)\right\Vert^{}_F$ remains bounded by $\epsilon_{\text{app}}^{}$ until $t = T_{}^{}$, we restrict the analysis to a time interval $t \in [0,T)$ where the bound $ \Big\Vert \mathbf{X}(t)\left. - \tilde{\mathbf{X}}(t)\right\Vert^{}_F \leq \epsilon_{\text{app}^{}}^{}$ holds.  

\begin{align}
   2\sqrt{2N}D^2 \BiggFo(  \frac{1}{\mathcal{E}_{e_i^{}}^{-1/2}(0)- \frac{1}{2} C_\Lambda^\prime T_{}^{}} + \frac{1}{\mathcal{E}_{r_i^{}}^{-1/2}(0)- \frac{1}{2} \tilde{C}_\Lambda^\prime T_{}^{}}\BiggFo) \leq \epsilon_{\text{app}}^{}\label{eqn:IRMMV_DistanceReferenceTrajectoryEstimatedTrajectoryV3}
\end{align}
 Let us define the worst case constants as $d(0):= \min\big\{\mathcal{E}_{e_i^{}}^{}(0)^{-1/2}, \mathcal{E}_{r_i^{}}^{}(0)^{-1/2}\big\} $ and $C_{\max}^{} = \max\big\{C_{\Lambda}^\prime, \tilde{C}_\Lambda^\prime\big\}$. 
%
Then, at any $t< T_{}$, where $T_{}^{} =\min\left\{ \frac{2}{C_\Lambda^\prime\sqrt{ \mathcal{E}_{e_i^{}}(0)}}, \frac{2}{\tilde{C}_\Lambda^\prime \sqrt{\mathcal{E}_{r_i^{}}(0)}} \right\}$, the bound \eqref{eqn:IRMMV_DistanceReferenceTrajectoryEstimatedTrajectoryV3} simplifies to
 \begin{align}
   \frac{4\sqrt{2N}D^2 }{d(0)- \frac{1}{2} C_{\max}^{} T}\leq \epsilon_{\text{app}}^{}  \label{eqn:ErrorLBV1}
 \end{align}

 Let $\rho$ be set such that the initial energy of estimated trajectory is at-least that of the reference trajectory:
 \begin{align}
      \mathcal{E}_{e_i^{}}^{}(0)&\geq \mathcal{E}_{r_i^{}}^{}(0)\\
      {L(2L+1)}\alpha_V^{2}&\geq(2L+1)\rho^{2/3}\\
      L^{3/2}\alpha_V^3\geq\rho \label{eqn:IRMMV_rhoLimit}
 \end{align}
Under this setting, $d(0) = \mathcal{E}_{e_i^{}}^{}(0)^{-1/2} =\left({\alpha_V^{}\sqrt{L(2L+1)}}\right)^{-1}_{}$. \\
Note that
\begin{align}
    C_{\max}^{} &= \max\big\{C_{\Lambda}^\prime, \tilde{C}_\Lambda^\prime\big\}\\
    \intertext{Substituting the constants $C_{\Lambda}^\prime =  \frac{8}{\sqrt{2L + 1}}\left(\Vert\mathbf{Y}\Vert_F + 2 L^{3/2}N \alpha_V^3\right),\quad \tilde{C}_\Lambda^\prime= \frac{8}{\sqrt{2L + 1}}\left(\Vert\mathbf{Y}\Vert_F + \sqrt{K}N^{} \rho\right)$}
    C_{\max}^{} &= \frac{8}{\sqrt{2L + 1}}\max\bigg\{\Vert\mathbf{Y}\Vert_F + 2 L^{3/2}N \alpha_V^3, \Vert\mathbf{Y}\Vert_F + \sqrt{K}N^{} \rho\bigg\}\\
    &= \frac{8}{\sqrt{2L + 1}}\cdot\left(\Vert\mathbf{Y}\Vert_F + N \max\bigg\{ 2 L^{3/2} \alpha_V^3, \sqrt{K} \rho\bigg\}\right)
    \intertext{Substituting the bound \eqref{eqn:IRMMV_rhoLimit}}
    C_{\max}^{}&\leq \frac{8}{\sqrt{2L + 1}} \left(\Vert\mathbf{Y}\Vert_F + N L^{3/2} \alpha_V^3\max\bigg\{ 2, \sqrt{K} \bigg\}\right)\label{eqn:IRMMV_C_maxLimit}
\end{align}

To ensure that the approximation error $\epsilon_{\text{app}}^{}$ remains small enough i.e., $\epsilon_{\text{app}}^{} \in (0,1)$, time $T$ should satisfy equation \eqref{eqn:ErrorLBV1}.

Rewriting equation \eqref{eqn:ErrorLBV1}, and substituting the constants,  
\begin{align}
    \frac{4\sqrt{2N}D^2 }{d(0)- \frac{1}{2} C_{\max}^{} T}&< \epsilon_{\text{app}}^{} \\
    \intertext{Rearranging the terms:}
    \frac{4\sqrt{2N}D^2 }{\epsilon_{\text{app}}}  &< d(0)- \frac{1}{2} C_{\max}^{} T\\
    \intertext{Grouping the constants on the right hand side and time dependent term on the other:}
    \frac{1}{2} C_{\max}^{} T &<  d(0)-\frac{4\sqrt{2N}D^2 }{\epsilon_{\text{app}}} \\
    \intertext{Solving for $T$:}
    T &<  2\frac{d(0)-\frac{4\sqrt{2N}D^2 }{\epsilon_{\text{app}}}  }{C_{\max}} \\
    \intertext{Simplifying the numerator and denominator:}
    T &<  2\frac{d(0)\epsilon_{\text{app}}-{4\sqrt{2N}D^2 }  }{\epsilon_{\text{app}}C_{\max}} \\
    \intertext{Substituting constants $d(0) = \left({\alpha_V^{}\sqrt{L(2L+1)}}\right)^{-1}_{}, C_{\max}^{} = \frac{8}{\sqrt{2L + 1}} \left(\Vert\mathbf{Y}\Vert_F + N L^{3/2} \alpha_V^3\max\bigg\{ 2, \sqrt{K} \bigg\}\right)$}
    T &<  2\frac{\frac{\epsilon_{\text{app}}^{}}{\alpha_V^{}\sqrt{L(2L+1)}}-{4\sqrt{2N}D^2 }}{\frac{8}{\sqrt{2L + 1}} \epsilon_{\text{app}}^{}\left(\Vert\mathbf{Y}\Vert_F + N L^{3/2} \alpha_V^3\max\bigg\{ 2, \sqrt{K} \bigg\}\right)} \\
    \intertext{Simplifying, we have}
    T &<  \frac{\epsilon_{\text{app}}^{}-\alpha_V^{}{4\sqrt{2NL(2L+1)}D^2 }}{4\epsilon_{\text{app}}^{}\sqrt{L}\alpha_V^{} \left(\Vert\mathbf{Y}\Vert_F + N L^{3/2} \alpha_V^3\max\bigg\{ 2, \sqrt{K} \bigg\}\right)} 
\end{align}

To ensure that the upper bound on $T$ is positive, the numerator should be positive. This limits $\alpha_V^{}$ as follows:

\begin{align}
    0&< \epsilon_{\text{app}}^{}-\alpha_V^{}4\sqrt{2NL(2L+1)}D^2\\
    \alpha_V^{}4\sqrt{2NL(2L+1)}D^2 &<\epsilon_{\text{app}}^{}\\
    \alpha_V^{} &<\frac{\epsilon_{\text{app}}^{}}{4\sqrt{2NL(2L+1)}D^2}
\end{align}

This initialization ensures that the error $\Vert\mathbf{X}(t) - \tilde{\mathbf{X}}(t)\Vert_F^{}$ remains below $\epsilon_{app}^{}$ for any time $t\in[0,T_{}^{})$.

\subsection{Proof of Corollary 1}
\label{sec:IRMMV_proofofCorollaryV1}
\begin{corollary}[Corollary \ref{cor:IRMMV_MainResult} restated]
Assume the conditions of Theorem \ref{thm:IRMMV_mainResult}, and in addition assume that all reference trajectories $\tilde{\mathbf{X}}(t)$ converge to a solution $\mathbf{X}^\star_{}\in\mathbb{R}^{N\times L}$. This convergence is uniform in the sense that the trajectories are confined to a bounded domain, and for any $\epsilon_{\text{app}}^{} >0$ there exists a time $T_c^{}\leq T$ after which they are all within a distance $\epsilon_{\text{app}}^{}$ from $\mathbf{X}^\star_{}$. Then for any $\epsilon_{\text{app}}^{}>0$, if the initialization scales $\alpha_V^{}, \alpha_g^{}$ are sufficiently small, for any time $t \in [T_c^{}, T]$ it holds that $\left\Vert \mathbf{X}(t) - \mathbf{X}^\star_{}\right\Vert ^{}_F\leq 2\epsilon_{\text{app}}^{}$
\end{corollary}
\begin{proof}
    For $\epsilon_{\text{app}}^{}>0$, there exists a time $T_c^{} >0$ after which all the reference trajectories are within a distance of $\epsilon_{\text{app}}^{}$ from the solution $\mathbf{X}^\star_{}$. i.e.,
    \begin{align}
        \Vert \tilde{\mathbf{X}}(t) - \mathbf{X}^\star\Vert \leq \epsilon_{\text{app}}^{}. 
    \end{align}
    where $t\geq T_c^{}$. 
    These reference trajectories are also confined within a ball of radius ${D}$. i.e., $\Vert \tilde{X}(t)\Vert \leq D$ for all $t\geq 0$. 
    From Theorem \ref{thm:IRMMV_mainResult}, if initialization scales $\alpha_V^{}, \; \alpha_g^{}$ are sufficiently small, then \begin{align}
        \Vert \mathbf{X}(t) - \tilde{\mathbf{X}}(t)\Vert \leq \epsilon_{\text{app}}^{} <  1. 
    \end{align}
    This bound holds atleast until $t\geq T$ or $\Vert \mathbf{X}(t)\Vert \geq D+1$. In line with the analysis done by \citet{razin2021implicit}, by the way of contradiction, we show that the second condition cannot hold, namely $\Vert \mathbf{X}(t)\Vert \leq D+1$ cannot be true for any $t\in[0,T]$. Let $t^\prime_{}\in[0, T]$ be an initial time at which $\Vert \mathbf{X}(t^{\prime}_{})\Vert \geq D+1$. Since $t^\prime_{}\leq T$, the guarantee from Theorem \ref{thm:IRMMV_mainResult} is still valid, meaning, $\Vert \mathbf{X}(t^\prime_{}) - \tilde{\mathbf{X}}(t^\prime_{})\Vert  < 1$, using the triangle inequality we have: 
    \begin{align}
        \Vert \tilde{\mathbf{X}}(t^\prime_{})\Vert &\geq \Vert {\mathbf{X}}(t^\prime_{})\Vert  - \Vert {\mathbf{X}}(t^\prime_{})-\tilde{\mathbf{X}}(t^\prime_{})\Vert > (D+1) - 1 \\
        &> D
    \end{align}
    This is a contradiction to the assumption that $\tilde{\mathbf{X}}(t)$ is confined to a ball of radius $D$ for $t = t^\prime_{}$. Thus, for all $t\in[0,T]$, the guarantee $ \Vert \mathbf{X}(t) - \tilde{\mathbf{X}}(t)\Vert \leq \epsilon_{\text{app}}^{}$ holds. 

    For any time $t\in[T_c^{}, T]$ we can apply the triangle inequality: 
    \begin{align}
        \Vert \mathbf{X}(t) - \mathbf{X}^\star\Vert \leq \Vert  
        \Vert \mathbf{X}(t) - \tilde{\mathbf{X}}(t)\Vert  + 
        \Vert  \tilde{\mathbf{X}}(t) - \mathbf{X}^\star_{}\Vert  \leq 2 \epsilon_{\text{app}}^{}
    \end{align}
    This completes the proof. 
\end{proof}

\section{Result on MNIST}
\label{sec:MNISTResultApp}

To demonstrate the efficacy of the proposed algorithm on real-world data, we conduct a recovery experiment on a set of images from the MNIST dataset. The experiment uses the first $100$ images from the MNIST dataset, each reshaped into a vector of $784$ pixels. The images are normalized to have pixel values between $0$ and $1$. A random Gaussian sensing matrix $\mathbf{A}$ of size $700 \times 784$ with $\ell_2$-normalized columns is used to generate the measurements. The measurement matrix $\mathbf{Y}$ is then computed as $\mathbf{Y} = \mathbf{A}\mathbf{X}$, where $\mathbf{X}$ represents the matrix of vectorized MNIST images. The proposed \acrshort*{irmmv} algorithm is used to recover the images from the measurements $\mathbf{Y}$ and the sensing matrix $\mathbf{A}$.
For comparison, we also show the results from the \acrshort*{momp}, \acrshort*{msp}, \acrshort*{mfocuss}, \acrshort*{mamp}, and \acrshort*{bsbl} algorithms. For \acrshort*{momp}, \acrshort*{msp}, and \acrshort*{mamp} which require prior knowledge of the row sparsity level, it is set to $K=300$. 
The simulations were performed on the GPU of a desktop computer equipped with an Intel Core i$5$-$7600K$ processor, $16$GB of DDR$4$ RAM and GeForce GTX $1070$ with $8$GB RAM. The MNIST dataset was obtained under the Creative Commons Attribution-Share Alike 3.0 license. 

\begin{figure*}[t]
    \centering
    \newcommand{\imgwidth}{0.13\linewidth} 
    \setlength{\tabcolsep}{1pt} 
    
    \begin{tabular}{lccccc} 
        \toprule
        \textbf{Method} & \textbf{Digit 5} & \textbf{Digit 0} & \textbf{Digit 4} & \textbf{Digit 1} & \textbf{Digit 9} \\ 
        \midrule
        
        \textbf{Original} & 
        \includegraphics[width=\imgwidth]{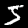} & 
        \includegraphics[width=\imgwidth]{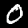} & 
        \includegraphics[width=\imgwidth]{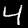} & 
        \includegraphics[width=\imgwidth]{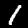} & 
        \includegraphics[width=\imgwidth]{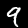} \\ \addlinespace[2pt]
        
        \textbf{Proposed} & 
        \includegraphics[width=\imgwidth]{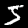} & 
        \includegraphics[width=\imgwidth]{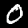} & 
        \includegraphics[width=\imgwidth]{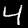} & 
        \includegraphics[width=\imgwidth]{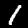} & 
        \includegraphics[width=\imgwidth]{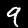} \\ \addlinespace[2pt]
        
        \textbf{M-OMP} & 
        \includegraphics[width=\imgwidth]{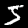} & 
        \includegraphics[width=\imgwidth]{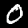} & 
        \includegraphics[width=\imgwidth]{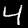} & 
        \includegraphics[width=\imgwidth]{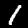} & 
        \includegraphics[width=\imgwidth]{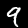} \\ \addlinespace[2pt]
        
        \textbf{M-SP} & 
        \includegraphics[width=\imgwidth]{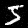} & 
        \includegraphics[width=\imgwidth]{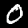} & 
        \includegraphics[width=\imgwidth]{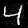} & 
        \includegraphics[width=\imgwidth]{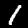} & 
        \includegraphics[width=\imgwidth]{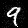} \\ \addlinespace[2pt]
        \textbf{M-FOCUSS} & 
        \includegraphics[width=\imgwidth]{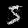} & 
        \includegraphics[width=\imgwidth]{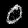} & 
        \includegraphics[width=\imgwidth]{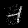} & 
        \includegraphics[width=\imgwidth]{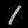} & 
        \includegraphics[width=\imgwidth]{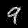} \\ 
        \addlinespace[2pt]
        \textbf{M-BSBL} & 
        \includegraphics[width=\imgwidth]{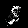} & 
        \includegraphics[width=\imgwidth]{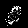} & 
        \includegraphics[width=\imgwidth]{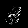} & 
        \includegraphics[width=\imgwidth]{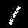} & 
        \includegraphics[width=\imgwidth]{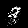} \\ \addlinespace[2pt]

        \textbf{AMP-MMV} & 
        \includegraphics[width=\imgwidth]{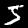} & 
        \includegraphics[width=\imgwidth]{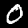} & 
        \includegraphics[width=\imgwidth]{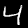} & 
        \includegraphics[width=\imgwidth]{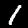} & 
        \includegraphics[width=\imgwidth]{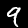} \\

        \bottomrule
    \end{tabular}
    \caption{Comparison of MNIST digit reconstruction using different methods. The first row shows the original images.}
    \label{fig:mnist_reconstruction}
\end{figure*}

As shown in Figure \ref{fig:mnist_reconstruction}, the proposed method successfully recovers the digit images, achieving a visual quality comparable to that of the other approaches without requiring any prior information about the signal sparsity.

\section{Code and Reproducibility}
\label{sec:IRMMV_Reproducability}
The source code, and instructions needed to reproduce all experimental results are provided in the supplementary material as a ZIP file. All experiments were conducted in a Python 3 environment with dependencies listed in `requirements.txt'. To reproduce all results, please follow the detailed instructions in the README.md file included in the ZIP file.

\end{document}